\documentclass[11pt]{article}

\usepackage[utf8]{inputenc}

\usepackage{color}
\usepackage{latexsym}
\usepackage{dsfont}
\usepackage{amssymb}
\usepackage{comment}
\usepackage{graphicx}
\usepackage{amsmath,amsfonts,amssymb,theorem,euscript,array,enumerate,amsfonts,mathrsfs}
\usepackage{hyperref}
\usepackage{appendix}
\usepackage[T1]{fontenc}
\usepackage{babel}
\numberwithin{equation}{section}
\usepackage{bbm}
\usepackage{subfigure}
\usepackage{color}

\usepackage{stmaryrd}

\usepackage[algo2e,ruled,vlined]{algorithm2e} 
 \usepackage{algorithm}
 \usepackage{algorithmicx}
\usepackage{algpseudocode}
\usepackage{ marvosym }
\usepackage{csquotes}
\usepackage{hyperref}

\usepackage{mathtools}
\mathtoolsset{showonlyrefs}

\usepackage{comment}

\usepackage{tikz}
\usetikzlibrary{fit,matrix,chains,positioning,decorations.pathreplacing,arrows}

\usepackage{mathtools}
\mathtoolsset{showonlyrefs}

\usepackage[a4paper,margin=1in,heightrounded]{geometry}

\usepackage{algpseudocode}
\usepackage{algorithm}

\usepackage[normalem]{ulem}

\def \trans{^{\scriptscriptstyle{\intercal}}}

\def \Frac{\displaystyle\frac}

\def \b1{\bf{1}}

\def \R{\mathbb{R}}

\def \E{\mathbb{E}}
\def \F{\mathbb{F}}

\def \P{\mathbb{P}}

\def \S{\mathbb{S}}
\def \U{\mathbb{U}}

\def\boM{{\boldsymbol M}}

\def \mrb{\mathrm{b}}

\def \d{\mathrm{d}}

\def\esssup_#1{\underset{#1}{\mathrm{ess\,sup\, }}}

\def\argmin_#1{\underset{#1}{\mathrm{argmin\, }}}
\def\argmax_#1{\underset{#1}{\mathrm{argmax\, }}}

\def \Ac{{\cal A}}
\def \Bc{{\cal B}}

\def \Dc{{\cal D}}
\def \Ec{{\cal E}}
\def \Fc{{\cal F}}
\def \Gc{{\cal G}}
\def \Hc{{\cal H}}

\def \Lc{{\cal L}}
\def \Pc{{\cal P}}

\def \Nc{{\cal N}}

\def \Sc{{\cal S}}
\def \Tc{{\cal T}}

\def \Wc{{\cal W}}

\def \ep{\hbox{ }\hfill$\Box$}

\def\Dt#1{\Frac{\partial #1}{\partial t}}

\def \mrm{\mathrm{m}}

\def \mrb{\mathrm{b}}

\def \d{\mathrm{d}} 
 
\def \mrp{\mathrm{p}}

\def \mrJ{\mathrm{J}} 
 
\def \mrG{\mathrm{G}} 
 
\def \mrV{\mathrm{V}}

\def\beqs{\begin{eqnarray*}}
\def\enqs{\end{eqnarray*}}
\def\beq{\begin{eqnarray}}
\def\enq{\end{eqnarray}}

\newcommand{\red}[1]{\textcolor{red}{#1}}

\def\red#1{{\color{red}#1}}

\newtheorem{Theorem}{Theorem}[section]

\newtheorem{Proposition}{Proposition}[section]
\newtheorem{Assumption}{Assumption}[section]
\newtheorem{Lemma}{Lemma}[section]

\newtheorem{Remark}{Remark}[section]

\numberwithin{equation}{section}

\newcommand{\bes}[1]{\begin{equation} \begin{split} #1\end{split} \end{equation}}

\title{Actor-Critic learning  for mean-field control in continuous time}

\author{Noufel FRIKHA\footnote{CES, UMR 8174,  Université Paris 1 Panthéon Sorbonne, \sf\href{mailto: Noufel.Frikha at univ-paris1.fr }{Noufel.Frikha at univ-paris1.fr} } 
\quad Maximilien GERMAIN\footnote{LPSM,  Universit\'e Paris Cité, \sf\href{mailto: maximilien.germain at gmail.com}{maximilien.germain at gmail.com} }  
\quad Mathieu LAURIERE\footnote{NYU Shanghai, \sf\href{mailto: mathieu(dot)lauriere at nyu.edu}{mathieu.lauriere at nyu.edu}}  \\
\quad Huy\^en PHAM\footnote{ LPSM,  Universit\'e Paris Cité,   \sf \href{mailto:pham at lpsm.paris}{pham at lpsm.paris}; This author  is supported by  the BNP-PAR Chair ``Futures of Quantitative Finance", 
and by FiME, Laboratoire de Finance des March\'es de l'Energie, and the ``Finance and Sustainable Development'' EDF - CACIB Chair}  
\quad Xuanye SONG\footnote{LPSM,  Universit\'e Paris Cité,   \sf \href{mailto:pham at lpsm.paris}{xsong at lpsm.paris};} }

\begin{document}

\maketitle

\begin{abstract}
We study policy gradient for mean-field control in continuous time in a  reinforcement learning setting. By considering randomised policies with entropy regularisation, we derive a gradient expectation representation of the value function, which is amenable to actor-critic type  algorithms, 
where the value functions and the policies are learnt alternately based on observation samples of the state  and model-free estimation of the population state distribution, either by offline or online learning.  
In the linear-quadratic mean-field framework, we obtain an exact parametrisation of the actor and critic functions defined on the Wasserstein space. Finally, we illustrate the results of our algorithms with some numerical experiments on  concrete examples. 
\end{abstract}

\vspace{3mm}






\section{Introduction}

{\it Mean-field control} (MFC in short),  also called {\it McKean-Vlasov}  (MKV in short) control problem  is concerned with the study of  large population models  of interacting agents who are cooperative and act for collective welfare according to a center of decision (or social planner).  
It has attracted a growing interest over the last years with the emergence of mean-field game, and there is now a large literature on the theory and its various applications in  economics/finance, population dynamics, social sciences and herd behavior. We refer to the seminal two-volume monograph \cite{cardel19}-\cite{cardel19b} for a detailed treatment of the topic.

Mean-field control problems lead to infinite dimensional problems in the Wasserstein space of probability measures, and analytical solutions are rarely available. It is then crucial to design efficient numerical schemes for solving such problems, and in the past few years, several works have proposed numerical methods in a model-based setting based 
either on forward-backward SDE characterisation of MKV from Pontryagin maximum principle, or Master Bellman equation from dynamic programming, and often relying on suitable class of neural networks, see e.g.  \cite{carlau21},  \cite{gerlauphawar21},  \cite{germikwar22}, \cite{hanhulong22}, \cite{reistozha21}, \cite{phawar22}. 

The question of learning solutions to MFC in a model-free setting, i.e. when the environment (model coefficients) is unknown, has recently attracted attention, see \cite{carlautan19,carmona2019linear}, \cite{guetal20},  \cite{angfoulau21}, and this is precisely the purpose of {\it Reinforcement learning} (RL):  learn optimal control  by trial and error, i.e., repeatedly try 
policy, observe the state, receive and evaluate the reward, and improve the policy.  There are two main approaches in RL: (i) {\it $Q$-learning} based on dynamic programming, and (ii) {\it Policy gradient}  based on parametrisation of policies, and a key feature in RL is the {\it exploration} of the unknown 
environment  to  broaden search space, which can be achieved  via randomised policies. RL is a very active branch of machine learning and we refer to the second edition of the monograph \cite{sutbar18} for an overview of this field. 

Most algorithms in RL are  limited to discrete-time frameworks for  Markov decision processes (MDP) or mean-field MDP, and the study of RL in continuous time has been recently initiated in \cite{wanzarzhou20}, \cite{jiazhou21a}, \cite{jiazhou21} for controlled diffusion processes.  
In line with these works, we provide in this paper a theoretical treatment of policy gradient methods for MFC  in continuous time and state/action space by relying on stochastic calculus that has been recently developed for MKV equations.  Our main theoretical result is to obtain a policy gradient representation for value function with 
randomised parametric policies and entropy regularisers for encouraging exploration. Based on this representation, we design model-free actor critic algorithms involving either the whole trajectories of the state (off-line learning), or the current and next state (online learning).  In the mean-field context, a key issue is to handle the population state distribution, which is an input of the policy (actor) and value function (critic), and instead of assuming that we have at disposal a simulator of the state distribution as in \cite{carlautan19}, we estimate it in a model-free manner as in \cite{angfoulau21}, which is more suitable for real-world applications. We next study the linear quadratic (LQ) case for which we derive explicit solutions, and this can be used for proposing an exact parametrisation of the critic and actor functions that is incorporated in  stochastic gradient  when updating the policies and value functions.  The explicit solutions in the LQ setting  are served as benchmarks for the numerical results of our algorithms in two examples. 

The rest of the paper is organized as follows. In Section \ref{sec:explorMF}, we formulate the mean-field control problem in continuous-time with randomised policies and entropy regularisers, and state the partial differential equation (PDE) characterisation of the value function in the Wasserstein space.  We develop in Section \ref{sec:PG} policy gradient methods by establishing a policy gradient representation, and its implication for actor-critic algorithms.  Section \ref{sec:LQ} is devoted to the linear-quadratic setting, and we present in Section \ref{sec:num} numerical results on two examples to illustrate the accuracy of our algorithms.  Finally, proofs of the policy gradient theorem are detailed in Appendix 
\ref{sec:appenA}, while the derivation of the explicit solution in the LQ case is shown in Appendix \ref{sec:LQappen}.


\vspace{2mm}

\noindent {\bf Notations.} 
The scalar product between two vectors $x$ and $y$ is denoted by $x \cdot y$, and $| \cdot |$ is  the Euclidian norm.  
Given two matrices $M$ $=$ $(M_{ij})$ and $N$ $=$ $(N_{ij})$, we denote by $M:N$ $=$ ${\rm Tr}(M\trans N)$ $=$ $\sum_{i,j} M_{ij} N_{ij}$ its inner product, and by $|M|$ the Frobenius norm of $M$. 
Here $\trans$ is the transpose matrice operator. 
Let $\boM$ $=$ $(\boM_{i_1i_2i_3})$ $\in$ $\R^{d_1\times d_2\times d_3}$ be  a tensor of order $3$. For $p$ $=$ $1,2,3$, the $p$-mode product of  $\boM$ with a vector $b$ $=$ $(b_i)$ $\in$ $\R^{d_p}$, is denoted by $\boM\bullet_p b$, and it is a tensor of order $2$, i.e. a matrix 
defined elementwise as 
\bes{
\big( \boM\bullet_1 b)_{i_2i_3} & = \; \sum_{i_1=1}^{d_1} M_{i_1i_2i_3} b_{i_1}, \;    \big( \boM\bullet_2 b)_{i_1i_3} \;  = \; \sum_{i_2=1}^{d_2}  M_{i_1i_2i_3} b_{i_2}, \;   \big( \boM\bullet_3 b)_{i_1i_2} \;  = \; \sum_{i_3=1}^{d_3} M_{i_1i_2i_3} b_{i_3}.  
} 
The $p$-mode product of a $3$-th order tensor $\boldsymbol{M}$ $\in$ $\R^{d_1\times d_2\times d_3}$ with a matrix $B$  $=$ $(B_{ij})$ $\in$ $\R^{d_p\times d}$, also denoted by $\boldsymbol{M}\bullet_p B$,  is a $3$-th order 
 tensor 
 defined elementwise as
\bes{
\big(\boldsymbol{M}\bullet_1 B \big)_{\ell i_2i_3} & \; = \sum_{i_1 =1}^{d_1} M_{i_1i_2i_3} B_{i_1\ell}, \quad \big(\boldsymbol{M}\bullet_2 B \big)_{i_1\ell i_3}  \; = \sum_{i_2 =1}^{d_2} M_{i_1i_2i_3} B_{i_2\ell} \\
\big( \boM\bullet_3 B)_{i_1i_2\ell} & = \; \sum_{i_3=1}^{d_3} M_{i_1i_2i_3} B_{i_3\ell}. 
}
Finally, the tensor contraction (or partial trace) of a $3$-th order tensor $\boldsymbol{M}$ $\in$ $\R^{d_1\times d_2\times d_3}$ whose dimensions $d_p$ and $d_q$ are equal is denoted as $\mathrm{Tr}_{p,q} \boldsymbol{M}$. This tensor contraction is a tensor of order 1, i.e. a vector, defined elementwise as 
\bes{
    \big(\mathrm{Tr}_{1,2} \boldsymbol{M} \big)_{i_3} = \; \sum_{\ell=1}^{d_1} M_{\ell \ell i_3}, \;     \big(\mathrm{Tr}_{1,3} \boldsymbol{M} \big)_{i_2} = \; \sum_{\ell=1}^{d_1} M_{\ell i_2 \ell}, \;     \big(\mathrm{Tr}_{2,3} \boldsymbol{M} \big)_{i_1} = \; \sum_{\ell=1}^{d_2} M_{i_1 \ell \ell}.
}


\section{Exploratory formulation of mean-field control} \label{sec:explorMF}

Let us  consider a mean-field control problem where the $\R^d$-valued controlled state  process $X$ $=$ $X^\alpha$  is governed by the dynamics
\begin{align} \label{SDEX}
\d X_s &= \;  b(X_s,\P_{X_s},\alpha_s) \d s  + \sigma(X_s,\P_{X_s},\alpha_s) \d W_s, \quad   s \geq 0,    
\end{align} 
with $W$ a standard $p$-dimensional Brownian motion on a  probability space $(\Omega,\Fc,\P)$ equipped with the filtration $\F$ $=$ $(\Fc_t)_{t\geq 0}$ generated by $W$, and augmented with a $\sigma$-algebra $\Gc$ rich enough to support a uniformly distributed random variable 
independent of $W$. The control $\alpha$ $=$ $(\alpha_t)_t$  is an $\F$-progressively measurable  process with $\alpha_t$ representing the action of the agent at time $t$, and valued in the action space $A$ $\subset$ $\R^q$.  Here, $\P_{X_t}$ denotes the marginal law of $X_t$, 
$\Pc_2(\R^d)$ is the Wasserstein space of probability measures $\mu$ with a finite second order moment, i.e., $M_2(\mu)$ $:=$ $\big(\int |x|^2 \mu(\d x)\big)^{1\over 2}$ $<$ $\infty$,  
equipped with the Wasserstein distance $\Wc_2$, and the coefficient $b$  (resp. $\sigma$) is a measurable  function from $\R^d\times\Pc_2(\R^d)\times A$ into $\R^d$ (resp. $\R^{d\times p}$).

Throughout the paper, we make the standard Lipschitz assumptions on the coefficients $b$ and $\sigma$ to ensure the existence and uniqueness of a strong solution  
to the stochastic differential equation (SDE in short) \eqref{SDEX} given any initial condition $\xi$ with law $\mu \in \Pc_2(\R^d)$. 

The objective of a mean-field  control problem on finite horizon $T$ $<$ $\infty$, is
to minimize over the control $\alpha$ an expected total cost of the form
\begin{align} \label{defJ}
\E \Big[ \int_0^T e^{-\beta s}  f(X_s,\P_{X_s},\alpha_s) \d s + e^{-\beta T} g(X_T,\P_{X_T}) \Big].  
\end{align} 
Here $f$ is a running cost function defined on $\R^d\times\Pc_2(\R^d)\times A$, while $g$ is a terminal cost function on $\R^d\times\Pc_2(\R^d)$, and $\beta$ $\in$ $\R_+$ is a given discount factor. In a model-based setting, i.e., when the coefficients $b$, $\sigma$, and the functions $f$, $g$ are known, the solution to MFC control problem can be characterised by a forward backward SDE arising from the maximum principle (see \cite{cardel15},  or by a Master  Bellman equation arsing from dynamic programming principle (see \cite{phawei17}). Moreover,  the optimisation over $\F$-progressively measurable process $\alpha$ (open-loop control), 
or feedback (also called closed-loop) controls $\alpha$, i.e., in the form $\alpha_t$ $=$ $\pi(t,X_t,\P_{X_t})$, $0\leq t\leq T$,  for some deterministic policy $\pi$, i.e., a  measurable function $\pi$ $:$  $[0,T]\times\R^d\times\Pc_2(\R^d)$ $\rightarrow$ $A$,  yields the same value function.

 In a model-free reinforcement learning (RL)  setting, when the coefficients are unknown, the agent can  only rely on observation samples of state and reward in order to learn the optimal strategy. This is achieved by {\it trial and error} where the agent tries a policy, receive and evaluate the reward and then improve performance 
 by repeating this procedure. A critical issue in reinforcement learning when the environment is unknown, is {\it exploration} in order to broaden search space, and a key and now common idea is to use {\it randomised} (or stochastic) policies: in a mean-field setting,  this is defined by a probability  transition kernel from 
 $[0,T]\times\R^d\times\Pc_2(\R^d)$ into $A$, i.e., a measurable function $\pi$ $:$ $(t,x,\mu)$ $\in$ $[0,T]\times\R^d\times\Pc_2(\R^d)$ $\mapsto$ $\pi(.|t,x,\mu)$ $\in$ $\Pc(A)$, the set of probability measures on $A$. 
 We then say that the process $\alpha$ $=$ $(\alpha_t)_t$  is a randomised feedback control generated from a stochastic  policy $\pi$, denoted by $\alpha$ $\sim$ $\pi$, if at each time $t$, the action $\alpha_t$ is sampled from the probability distribution $\pi(.|t,X_t,\P_{X_t})$. Note that the sampling is drawn at each time 
 from the $\sigma$-algebra $\Gc$ rich enough to support a uniformly distributed random variable independent of $W$. More precisely,  it is defined as follows: given a probability transition kernel  $\pi$, one can associate a measurable function $\phi_\pi$ $:$ $[0,T]\times\R^d\times\Pc_2(\R^d)\times [0,1]$ $\rightarrow$ $A$ such that 
 the law of $\phi_\pi(t,x,\mu,U)$ is $\pi(.|t,x,\mu)$ where $U$ is an uniform random variable on $[0,1]$.  We would then naturally define the control process by $\alpha_t$ $=$  $\phi_\pi(t,X_t,\P_{X_t},U_t)$, $0\leq t\leq T$,  
 for a collection of $\Gc$-measurable i.i.d. uniform random variables $(U_t)_t$,  but this raises some measurability issues as 
 $(t,\omega)$ $\mapsto$ $U_t(\omega)$ is not jointly measurable in the usual product space $([0,T]\times\Omega,\Bc_{[0,T]}\otimes\Gc,\d t \otimes\P)$.  To cope these issues, one can use the notion of Fubini extension, see  \cite{sun06}. We consider an atomless  probability space $([0,T],\Tc,\rho)$ extending the usual Lebesgue measure interval space 
 $([0,T],\Bc_{[0,T]},\d t)$, and a rich Fubini extension $([0,T]\times\Omega,\Tc\boxtimes\Gc,\rho\boxtimes\P)$ of the product space  $([0,T]\times\Omega,\Tc\otimes\Gc,\rho\otimes\P)$. Then, from Theorem 1 in \cite{sun06}, there exists a $\Tc\boxtimes\Gc$-measurable map $\U$ $:$ $[0,T]\times\Omega$ $\rightarrow$ $[0,1]$ such that the random variables 
 $U_t$ $=$ $\U(t,.)$ are essentially pairwise independent, and uniformly distributed on $[0,1]$. Denote by $\F$ the filtration generated by $(W,\U)$, and consider the controlled process governed by 
 \begin{align} \label{SDEXbis}
 \d X_s &= \;  b(X_s,\P_{X_s},\alpha_s) \d s + \sigma(X_s,\P_{X_s},\alpha_s) \d W_s,
 \end{align}
 where $\alpha_t$ $=$ $\phi_\pi(t,X_t,\P_{X_t},U_t)$ $\sim$ $\pi(.|t,X_t,\P_{X_t})$, $0\leq t\leq T$, is $\F$-progressively measurable.  Here, to alleviate notations, we write $\rho(\d t)$ $\equiv$ $\d t$.

Moreover, in order to encourage exploration of randomised policies, we shall substract entropy regularisers to the cost term, as adopted in the recent works by \cite{wanzarzhou20}, \cite{guoxuzar20}, by considering the Shannon differential entropy defined as
\beqs
\Ec(\pi(.|t,x,\mu)) & := & - \int_A \log p(t,x,\mu,a) \pi(\d a|t,x,\mu), 
\enqs  
by assuming that $\pi(.|t,x,\mu)$ admits a density $p(t,x,\mu,.)$ with respect to some measure $\nu$ on $A$. 
The goal of the social planner is now to minimise over randomised policies $\pi$ the cost 
\begin{align} \label{defJpi} 
J(\pi) &= \;  \E_{\alpha\sim\pi} \Big[ \int_0^T e^{-\beta s} \big[  f(X_s,\P_{X_s},\alpha_s) -  \lambda \Ec\big( \pi(.|s,X_s,\P_{X_s}) \big)  \big] \d s   +   e^{-\beta T} g(X_T,\P_{X_T}) \Big],
\end{align}
where $\lambda$ $\geq$ $0$ is a  temperature parameter on exploration.  Here, the notation in $\E_{\alpha\sim\pi}[.]$  means that the expectation operator is taken when the randomised feedback control $\alpha$ is generated from the stochastic policy $\pi$, and $X$ $=$ $X^\alpha$ is driven by the dynamics \eqref{SDEXbis}.

Let us now introduce the dynamic Markovian version of the above mean-field problem. 
Given a stochastic policy $\pi$, an initial time-state-distribution triple $(t,x,\mu)$ $\in$ $[0,T]\times\R^d\times\Pc_2(\R^d)$, and $\xi$ $\in$ $L^2(\Fc_t;\R^d)$ (the set of square-integrable $\Fc_t$-measurable random variables valued in $\R^d$) with distribution law $\mu$ ($\xi$ $\sim$ $\mu$),  we consider the decoupled  state processes $\{X_s^{t,\xi},t\leq s\leq T\}$ and $\{X_s^{t,x,\xi},t\leq s\leq T\}$ given by 
\begin{equation}
\label{McKean:Vlasov:SDE:and:decoupled:SDE}
\begin{aligned}
X_s^{t,\xi} &= \xi + \int_t^s b(X_r^{t,\xi},\P_{X_r^{t,\xi}},\alpha_r)  \d r + \int_t^s \sigma(X_r^{t,\xi},\P_{X_r^{t,\xi}},\alpha_r) \,  \d W_r,    \\
X_s^{t,x, \mu} &= x + \int_t^s b(X_r^{t,x, \mu},\P_{X_r^{t,\xi}},\alpha_r) \d r + \int_t^s \sigma(X_r^{t,x, \mu},\P_{X_r^{t,\xi}},\alpha_r) \, \d W_r, \quad t \leq s \leq T,  
\end{aligned}
\end{equation}
where $\alpha$ is a randomised feedback control generated from $\pi$, i.e.,  $\alpha_s$ is sampled at each time $s$ from $\pi(.|s,X_s^{t,x, \mu},\P_{X_s^{t,\xi}})$ (here, to alleviate notations, we omit the dependence  of $X^{t,\xi}$ and $X^{t,x, \mu}$ in $\alpha$ $\sim$ $\pi$). 
We make the standard Lipschitz regularity assumptions on the coefficients $b$ and $\sigma$ to ensure the existence and uniqueness of a strong solution   to  \eqref{McKean:Vlasov:SDE:and:decoupled:SDE} given any initial condition $t, \xi, x$.  
By weak uniqueness, it follows that the law of the process $(X_s^{t, \xi})_{s\in [t, T]}$ given by the unique solution to the first SDE in \eqref{McKean:Vlasov:SDE:and:decoupled:SDE} only depends upon $\xi$ through its law $\mu$. It thus makes sense to consider $(\P_{X_s^{t,\xi}})_{s\in [t, T]}$ as a function of $\mu$ without specifying the choice of the random variable $\xi$ that has $\mu$ as distribution. In particular, for any $0\leq t\leq s\leq T$, the random variable $X_s^{t,x,\mu}$ depends on 
$\xi$ only through its law $\mu$. As a consequence, we can define the cost value function of the stochastic policy $\pi$ as the function defined on $[0,T]\times\R^d\times\Pc_2(\R^d)$ by 
\begin{align} 
V^\pi(t,x,\mu) &= \;  \E_{\alpha\sim\pi} \Big[ \int_t^T e^{-\beta (s-t)}  \big[  f(X_s^{t,x, \mu}, \P_{X_s^{t,\xi}},\alpha_s) -  \lambda \Ec\big( \pi(.|s,X_s^{t,x,\mu},\P_{X_s^{t,\xi}}) \big) \big]  \d s  \nonumber \\
&  \hspace{3cm} + \;  e^{-\beta (T-t)} g(X_T^{t,x, \mu},\P_{X_T^{t,\xi}}) \Big].  \label{defVpi} 
\end{align} 
Since $X_s^{t, \xi, \xi} = X_s^{t, \xi}$ a.s., the initial cost  value in \eqref{defJpi} when starting from some initial random state $\xi$ $\in$ $L^2(\Gc;\R^d)$ with law $\mu$ is equal to  $J(\pi)$ $=$ $\E_{\xi\sim\mu}[V^\pi(0,\xi,\mu)]$.

 We complete this section by characterizing the cost value function $V^\pi$, for a given stochastic policy $\pi$, in terms of a linear parabolic partial differential equation (PDE) of mean-field type stated in the strip $[0,T] \times \mathbb{R}^d \times \mathcal{P}_2(\mathbb{R}^d)$. We first introduce the coefficients associated to the dynamics and the value function, given a stochastic policy $\pi$, namely
  \beqs
 b_\pi(t,x,\mu) \; = \; \int_A b(x,\mu, a)   \pi(\d a|t,x,\mu), & &  \Sigma_\pi(t,x,\mu) \; = \; \int_A (\sigma\sigma\trans)(x,\mu,a) \pi(\d a|t, x,\mu), \\
 f_\pi(t,x,\mu) \; = \; \int_A f(x, \mu, a) \pi(\d a|t,x,\mu), & &  E_\pi(t,x,\mu) \; = \;  - \int_A \log p(t,x,\mu,a) \pi(\d a|t,x,\mu),
 \enqs
 and let $\sigma_\pi := \Sigma_\pi^{1/2}$.

\vspace{1mm}

Before presenting the regularity assumptions, we introduce some notations regarding the Wasserstein derivative (also called L-derivative) of a real-valued smooth map $U$ defined on $\Pc_2(\R^d)$. We follow the common practice of denoting by $\partial_\mu U(\mu)(v) \in \R^d$ the Wasserstein derivative of $U$ with respect $\mu$ evaluated at $(\mu, v) \in \Pc_2(\R^d) \times \R^d$. Its $i$th coordinate is denoted by $\partial^{i}_\mu U(\mu)(v)$. We will also work with higher order derivatives. For a positive integer $n$, a multi-index $\lambda$ of $\left\{1, \cdots, d\right\}$, a $n$-tuple of multi-indices $\boldsymbol{\gamma}=(\gamma_1, \cdots, \gamma_n)$ of $\left\{1, \cdots, d\right\}$ and $\boldsymbol{v} = (v_1, \cdots, v_n)\in (\mathbb{R}^d)^{n}$, we denote by $\partial_{\mu}^{\lambda}U(\mu)(\boldsymbol{v})$ the derivative $\partial^{\lambda_n}_\mu[\cdots [\partial^{\lambda_1}_\mu U(\mu)](v_1)\cdots](v_n)$. If $\boldsymbol{v}\mapsto \partial_{\mu}^{\lambda}U(\mu)(\boldsymbol{v})$ is smooth, we write  $\partial^{\boldsymbol{\gamma}}_{\boldsymbol{v}}\partial_{\mu}^{\lambda}U(\mu)(\boldsymbol{v})$ for the derivative $\partial^{\gamma_n}_{v_n}\cdots \partial^{\gamma_1}_{v_1} \partial_{\mu}^{\lambda}U(\mu)(\boldsymbol{v})$.

We will often deal with maps that depend on additional time and space variables. In particular, we will work with the two spaces $\mathcal{C}^{2, 2}(\R^d \times \Pc_2(\R^d))$ and $\mathcal{C}^{1, 2, 2}([0,T] \times \R^d \times \Pc_2(\R^d))$ and refer the reader to \cite{cardel19} Chapter 5 for more details. 

Having these notations at hand, we make the following regularity assumptions on the coefficients $b_\pi , \sigma_\pi$, the cost functions $f_\pi, g$ and the Shannon differential entropy $E_\pi$. Below, $\pi:[0,T] \times \mathbb{R}^d \times \mathcal{P}_2(\mathbb{R}^d) \rightarrow \mathcal{P}(A)$ is a fixed stochastic policy.

 \begin{Assumption} \label{hypcoeff}
 \begin{enumerate}
\item[(i)] For any $h \in \{ b^i_\pi, \sigma_\pi^{i, j}, i=1, \cdots, d, j = 1, \cdots, p\}$, the following derivatives
$$
\partial_{x} h(t, x, \mu), \, \partial^2_x h(t, x, \mu), \, \partial_\mu h(t, x, \mu)(v), \, \partial_{v} [\partial_\mu h(t, x, \mu)](v),  
$$
\noindent exist for any $(t, x, v, \mu) \in [0,T] \times (\R^d)^2 \times \Pc_2(\R^d)$, are bounded and locally Lipschitz continuous with respect to $x, \mu, v$ uniformly in $t\in [0,T]$. Moreover, $h(t, .)$ is at most of linear growth, uniformly in $t\in [0,T]$, namely, there exists $C<\infty$ such that for all $t, x, \mu$
$$
|h(t, x, \mu)| \leq C (1+ |x| + M_2(\mu)).
$$

 \item[(ii)] For any $t\in [0,T]$, $f_\pi (t, .), \, E_\pi(t, .), \, g \in \mathcal{C}^{2,2}(\mathbb{R}^d \times \Pc_2(\mathbb{R}^d))$.
 \item[(iii)] There exists some constant $C< \infty$, such that for any $(t, x, v,\mu)$ $\in$ $[0,T]\times(\R^d)^2\times\Pc_2(\R^d)$, 
 $$
 |f_\pi (t, x, \mu)| + |E_\pi(t, x, \mu)| + |g(x, \mu)| \leq C (1 + |x|^2 + M_2(\mu)^q),
 $$ 
 $$
 | \partial_x f_\pi (t, x, \mu)| + |\partial_x E_\pi(t, x, \mu)| + |\partial_x g(x, \mu)| \leq C (1+ |x| + M_2(\mu)^q),
 $$
  $$
 | \partial_\mu f_\pi (t, x, \mu)(v) | + |\partial_\mu E_\pi(t, x, \mu)(v)| + |\partial_\mu g(x, \mu)(v)| \leq C (1+ |x| + |v| + M_2(\mu)^q),
 $$
 \begin{align*}
 | \partial_v [\partial_\mu f_\pi (t, x, \mu)](v) |+ &  | \partial^2_x f_\pi (t, x, \mu)| +  |\partial_v [\partial_\mu E_\pi(t, x, \mu)](v)| + |\partial^2_x E_\pi(t, x, \mu)| \\
 & + |\partial_v [\partial_\mu g(x, \mu)](v)|  + |\partial^2_x g(x, \mu)| \leq C (1+ M_2(\mu)^q),
 \end{align*}
 \noindent for some $q\geq 0$.
 \end{enumerate}
 \end{Assumption}

  \begin{Remark}
 It is readily seen from the integral form of $b_\pi$, $\Sigma_\pi$, $f_\pi$, $E_\pi$ that if for any $a \in A$, the functions $(x, \mu) \mapsto b(x, \mu, a), \sigma(x, \mu, a), \, f(x, \mu, a)$ and the density $(x, \mu) \mapsto p(t, x, \mu, a)$ of the probability measure $\pi(\d a|t,x,\mu)$ are smooth with derivatives satisfying some adequate estimates then Assumption 2.1 is satisfied. In particular, this will be the case when the coefficients $b$, $\sigma$ are linear functions and $f$ together with $g$ are quadratic functions of the variables of $x$, $\int_{\R^d} z \mu(\d z)$ and $a$ and if $p$ is a Gaussian density with a smooth mean and a time-dependent covariance-matrix as in the linear quadratic framework, see Section \ref{sec:LQ}. 
 \end{Remark}


We now have the following PDE characterisation of the cost value function $V^\pi$.

 \begin{Proposition} \label{proVpi} 
 Under Assumption \ref{hypcoeff}, the function $V^\pi$ defined by \eqref{defVpi} belongs to $C^{1,2, 2}([0,T]\times\R^d\times\Pc_2(\R^d))$ and satisfies the following linear parabolic PDE
 \begin{equation}\label{kolmogorov:PDE}
\Lc_\pi[V^\pi](t,x,\mu) + (f_\pi -  \lambda E_\pi)(t,x,\mu) = 0, \quad (t, x, \mu) \in [0,T)\times\R^d\times\Pc_2(\R^d),
 \end{equation}
\noindent with the terminal condition $V^\pi(T,x,\mu)$ $=$ $g(x,\mu)$, where 
 $\Lc_\pi$ is the operator defined by 
 \beqs
 \Lc_\pi[\varphi](t,x,\mu) 
 &=& - \beta \varphi(t, x, \mu) + \partial_t{\varphi}(t,x,\mu) 
 +  b_\pi(t,x,\mu) \cdot D_x\varphi(t,x,\mu) +  \frac{1}{2}  \Sigma_\pi(t,x,\mu) : D_x^2 \varphi(t,x,\mu)  \\
& & \; +  \; \E_{\xi\sim\mu} \Big[   b_\pi(t,\xi,\mu) \cdot \partial_\mu \varphi(t,x,\mu)(\xi) + \frac{1}{2} \Sigma_\pi(t,\xi,\mu) : \partial_\upsilon\partial_\mu \varphi(t,x,\mu)(\xi)    \Big].
 \enqs
 \end{Proposition}
 
 \begin{Remark}
 In particular, the above result indicates that provided the coefficients $b_\pi, \Sigma_\pi$, the functions $f_\pi$, $E_\pi$ and the terminal condition $g$ are smooth with derivatives satisfying some appropriate estimates, the solution $V^{\pi}$ to the Kolmogorov PDE \eqref{kolmogorov:PDE} is smooth. In this sense, it preserves the regularity of the terminal condition.
 
 However, one can weaken the regularity assumption on the terminal condition (and actually of the coefficients themselves) by benefiting from the smoothness of the underlying fundamental solution (or the transition density of the associated stochastic process) under some additional non-degeneracy assumption. We refer e.g. to \cite{crisan:murray}, \cite{CHAUDRUDERAYNAL20221}, \cite{CHAUDRUDERAYNAL20211} in the uniformly elliptic diffusion setting and to \cite{frikha:konakov:menozzi} in the case of non-degenerate stable driven SDE.
  \end{Remark}
 
\noindent {\bf Proof.}  See Appendix \ref{sec:proVpi} 
\ep

\section{Policy gradient method}  \label{sec:PG}

We now consider a parametric family of randomised policies $\pi_\theta$, with densities $p_\theta$, $\theta$ $\in$ $\Theta$, $\Theta$ being a non-empty open subset of $\R^D$, for some positive integer $D$, 
and denote by $\mrJ(\theta)$ $=$ $J(\pi_\theta)$ the associated cost function, viewed as a function of the parameters $\theta$, recalling that $J$ is defined by \eqref{defJpi}. The principle of  
policy gradient method is to minimize  over $\theta$ the function $\mrJ(\theta)$ by stochastic gradient descent  algorithm. In our RL setting, we aim to derive a probabilistic representation of the gradient function  $\nabla_\theta \mrJ(\theta)$ that does not involve model coefficients 
$b,\sigma$, but only observation samples of state $X_t$, state distribution $\P_{X_t}$, and rewards $f_t$ $:=$ $f(X_t,\P_{X_t},\alpha_t)$, $g_T$ $:=$ $g(X_T,\P_{X_T})$ when taking decision $\alpha$ $\sim$ $\pi_\theta$.

\subsection{Policy gradient representation}

We make the following assumptions on the parametric family of randomised policy and coefficients.

\begin{Assumption} \label{hyppitheta}
\begin{enumerate}
\item[(i)] For any $h \in \{ b^i_{\pi_\theta}, \sigma_{\pi_\theta}^{i, j}, f_{\pi_\theta}, E_{\pi_\theta}, g, i=1, \cdots, d, j = 1, \cdots, p\}$, any multi-indices $\alpha$, $\beta$, $\lambda$ of $\left\{1, \cdots, d\right\}$ such that $0\leq |\alpha| \leq 2$, $0\leq |\beta| \leq 1$, $\lambda$ being of length $n$, $0\leq n \leq 2$, any $n$-tuple of multi-indices $\boldsymbol{\gamma}=(\gamma_1, \cdots, \gamma_n)$ with $0\leq |\gamma_1| + \cdots + |\gamma_n|\leq 2$, denoting by $h_\theta(t, x, \mu)$ the value of $h$ at $(\theta, t, x, \mu)$, the following derivatives {\small 
\begin{align*}
& \partial^{\beta}_{\theta} \partial_x^{\alpha} \partial^{\boldsymbol{\gamma}}_{\boldsymbol{v}} \partial^{\lambda}_\mu h_{\theta}(t, x, \mu)(\boldsymbol{v}), \, \partial_x^{\alpha} \partial^{\beta}_{\theta}  \partial^{\boldsymbol{\gamma}}_{\boldsymbol{v}} \partial^{\lambda}_\mu h_{\theta}(t, x, \mu)(\boldsymbol{v}), \, \partial^{\alpha}_{x} \partial^{\boldsymbol{\gamma}}_{\boldsymbol{v}} \partial^{\beta}_{\theta}  \partial^{\lambda}_\mu h_{\theta}(t, x, \mu)(\boldsymbol{v}), \, \partial^{\alpha}_{x} \partial^{\boldsymbol{\gamma}}_{\boldsymbol{v}}  \partial^{\lambda}_\mu \partial^{\beta}_{\theta}  h_{\theta}(t, x, \mu)(\boldsymbol{v}), \\
& \partial^{\boldsymbol{\gamma}}_{\boldsymbol{v}} \partial^{\alpha}_{x} \partial^{\lambda}_\mu \partial^{\beta}_{\theta}  h_{\theta}(t, x, \mu)(\boldsymbol{v}), \partial^{\boldsymbol{\gamma}}_{\boldsymbol{v}}\partial^{\lambda}_\mu  \partial^{\alpha}_{x}  \partial^{\beta}_{\theta}  h_{\theta}(t, x, \mu)(\boldsymbol{v}),  \partial^{\boldsymbol{\gamma}}_{\boldsymbol{v}}\partial^{\lambda}_\mu   \partial^{\beta}_{\theta}  \partial^{\alpha}_{x} h_{\theta}(t, x, \mu)(\boldsymbol{v}), \partial^{\boldsymbol{\gamma}}_{\boldsymbol{v}}  \partial^{\beta}_{\theta}    \partial^{\lambda}_\mu 
\partial^{\alpha}_{x} h_{\theta}(t, x, \mu)(\boldsymbol{v}), \\ 
& \partial^{\beta}_{\theta}
\partial^{\boldsymbol{\gamma}}_{\boldsymbol{v}}      \partial^{\lambda}_\mu 
\partial^{\alpha}_{x} h_{\theta}(t, x, \mu)(\boldsymbol{v}), \partial^{\boldsymbol{\gamma}}_{\boldsymbol{v}}  \partial^{\beta}_{\theta}  \partial^{\alpha}_{x} \partial^{\lambda}_\mu   h_{\theta}(t, x, \mu)(\boldsymbol{v}),
 \partial^{\boldsymbol{\gamma}}_{\boldsymbol{v}}    \partial^{\alpha}_{x} \partial^{\beta}_{\theta} \partial^{\lambda}_\mu   h_{\theta}(t, x, \mu)(\boldsymbol{v}), \partial^{\beta}_{\theta} \partial^{\alpha}_{x} \partial^{\boldsymbol{\gamma}}_{\boldsymbol{v}}      \partial^{\lambda}_\mu   h_{\theta}(t, x, \mu)(\boldsymbol{v}),
\end{align*}
}

\noindent exist for any $(t,\theta, x, \boldsymbol{v}, \mu) \in  [0,T] \times \Theta  \times (\mathbb{R}^d)^{n+1} \times \Pc_2(\mathbb{R}^d)$ and are locally Lipschitz continuous with respect to $\theta, x, \mu, \boldsymbol{v}$ uniformly in $t\in [0,T]$\footnote{Hence, according to Clairaut's theorem, these partial derivatives are equal.}. Moreover, if $h= b^i_{\pi_\theta}$ or $\sigma_{\pi_\theta}^{i, j}$, the aforementioned derivatives of order greater or equal to one are bounded.
\item[(ii)]  The estimates of Assumption \ref{hypcoeff}(iii)  are satisfied for the family of policies $\left\{ \pi_\theta, \theta \in \Theta\right\}$, locally uniformly in $\theta$, i.e. for any $\theta \in \mathcal{K}$, $\mathcal{K}$ being any compact subset of $\Theta$. Additionally, there exists some constant $C< \infty$, such that for any $h\in \left\{f_{\pi_\theta}, E_{\pi_\theta}\right\}$, any $(t, \mu, x) \in [0,T] \times \Pc_2(\mathbb{R}^d) \times \mathbb{R}^d$, any $\boldsymbol{v}= (v_1, v_2) \in (\mathbb{R}^d)^2$, any $\theta \in \mathcal{K}$, $\mathcal{K}$ being any compact subset of $\Theta$, any multi-index $\lambda$, $|\lambda| =2 $, any multi-index $\lambda=(\lambda_1, \lambda_2)$ of $\left\{1, \cdots, d\right\}$, any couple of multi-indices $\boldsymbol{\gamma}=(\gamma_1, \gamma_2)$
 $$
 | \partial_{\theta} h_\theta (t, x, \mu)|  \leq C (1 + |x|^2 + M_2(\mu)^q),
 $$ 
 $$
 | \partial_\theta \partial_x h_\theta (t, x, \mu)|  + |\partial_x g(x, \mu)| \leq C (1+ |x| + M_2(\mu)^q),
 $$
  $$
 | \partial_\theta \partial_\mu h_\theta (t, x, \mu)(v_1) | + | \partial_\mu \partial_x h_\theta (t, x, \mu)(v_1) | + | \partial_\mu \partial_x g ( x, \mu)(v_1) |\leq C (1+ |x| + |v_1| + M_2(\mu)^q),
 $$
 \begin{align*}
&  | \partial_\theta \partial_{v_1} \partial_\mu h_\theta (t, x, \mu)(v_1) |+   | \partial_\theta \partial^2_x h_\theta (t, x, \mu)| +    | \partial_\mu \partial^2_x h_\theta (t, x, \mu)(v_1)| + | \partial_{\boldsymbol{v}}^{\boldsymbol{\gamma}}\partial^{\lambda}_\mu h_\theta(t, x, \mu)(\boldsymbol{v})|  \\
 & + |\partial_{v_1} \partial_\mu g(x, \mu)(v_1)|  + |\partial^2_x g(x, \mu)| + | \partial_{\boldsymbol{v}}^{\boldsymbol{\gamma}}\partial^{\lambda}_\mu g(t, x, \mu)(\boldsymbol{v})|\leq C (1+ M_2(\mu)^q),
 \end{align*}
 \noindent for some $q\geq 0$.
\end{enumerate}
\end{Assumption}

\vspace{3mm}

As shown in Appendix \ref{sec:diffV}, Assumption \ref{hyppitheta} guarantees that the derivatives $(t, \theta, x, \mu, v)$  $\mapsto$  $ \partial_\theta \partial_t \mrV_\theta(t, x, \mu)$, 
$\partial_\theta \mrV_\theta(t, x, \mu)$, $\partial_\theta \partial_{x} \mrV_\theta(t, x, \mu)$, $\partial_{\theta} \partial_\mu \mrV_\theta(t, x, \mu)(v)$,  $\partial_\theta \partial^2_{x} \mrV_\theta(t, x, \mu)$, $\partial_{\theta}\partial_{v}\partial_\mu \mrV_\theta(t, x, \mu)(v)$, where $\mrV_\theta(t, x, \mu)$ $:=$ $V^{\pi_\theta}(t, x, \mu)$ defined by \eqref{defVpi} with $\pi=\pi_\theta$, exist, are continuous and satisfy suitable growth conditions.

%

We then  let $\nabla_\theta\mrJ(\theta)$ $=$ $\E[\mrG_\theta(0, \xi, \mu)]$ where $\mrG_\theta(t, x, \mu) := \nabla_\theta \mrV_\theta(t, x, \mu)$.  The main result of this section provides a probabilistic representation of the gradient function $\mrG_\theta$. 

\begin{Theorem} \label{theogradient} 
Suppose that Assumption \ref{hyppitheta} holds. Assume moreover that for any $t, x, \mu, a$, the map $\Theta \ni \theta \mapsto p_\theta(t, x, \mu, a)$ is differentiable with a derivative satisfying the following estimates: for some constant $C<\infty$ and some $q\geq0$, for any $(t,x, \mu) \in [0,T] \times \R^d \times \Pc_
2(\R^d)$ and any compact subset $\mathcal{K} \subset \Theta$.
\begin{equation}
\label{diff:density:condition}
\begin{aligned}
\int_{A}\sup_{\theta \in \mathcal{K}} & \{ | \nabla_\theta p_\theta(t, x, \mu, a)| ( |b(x, \mu, a)| + |(\sigma \sigma)^{\trans}(x, \mu, a)| \\
&\quad  + |f(x, \mu, a)| + |\log(p_\theta(t, x, \mu, a))|) \} \nu(\d a) < \infty,
\end{aligned}
\end{equation}
\noindent and
\begin{equation}
\label{diff:density:condition:stochastic:integral}
\begin{aligned}
\int_{A} &  | \nabla_\theta \log(p_\theta(t, x, \mu, a))|^2 |\sigma(x, \mu, a)|^2 \, p_\theta(t, x, \mu, a) \, \nu(\d a) \leq C (1+ |x|^q + M_2(\mu)^q).
\end{aligned}
\end{equation}

Then, it holds 
\begin{align}\label{expressG} 
\mrG_\theta(t,x,\mu) & = \; \E_{\alpha\sim\pi_\theta}   \Big[ \int_t^T e^{-\beta(s-t)}  \nabla_\theta \log p_\theta(s,X_s^{t,x,\mu},\P_{X_s^{t,\xi}},\alpha_s) \Big\{ \d \mrV_\theta(s,X_s^{t,x,\mu},\P_{X_s^{t,\xi}}) \\
&  \hspace{1cm}  + \;  \big[  f(X_s^{t,x,\mu},\P_{X_s^{t,\xi}},\alpha_s) +  \lambda  \log p_\theta(s,X_s^{t,x,\mu},\P_{X_s^{t,\xi}},\alpha_s)  - \beta \mrV_\theta(s,X_s^{t,x,\mu},\P_{X_s^{t,\xi}})  \big] \d s  \Big\}   \\
& \hspace{3cm} + \; \int_t^T  e^{-\beta(s-t)} \Hc_\theta[\mrV_\theta](s,X_s^{t, x,\mu},\P_{X_s^{t,\xi}})   \d s \Big],  
\end{align} 
for any $(t,x,\mu, \theta)$ $\in$ $[0,T]\times\R^d\times\Pc_2(\R^d) \times \Theta$ and $\xi$ $\sim$ $\mu$, where $\Hc_\theta$ is the operator defined by 
\begin{align} \label{defH} 
\Hc_\theta[\varphi](t,x,\mu) &= \;   \E_{\xi\sim\mu} \big[ \nabla_\theta  b_\theta(t,\xi,\mu) \trans \partial_\mu \varphi(t,x,\mu)(\xi) \\
& \quad + \frac{1}{2} {\rm tr}_{1,2} \big( \nabla_\theta \Sigma_\theta(t,\xi,\mu) \bullet_1 \partial_\upsilon\partial_\mu \varphi(t,x,\mu)(\xi)\big)   \big], 
\end{align} 
\noindent and we set $b_\theta(t,x,\mu)$ $=$ $\int_A b(x,\mu,a) \, \pi_\theta(\d a|t,x,\mu)$, $\Sigma_\theta(t,x,\mu) \; = \; \int_A (\sigma\sigma\trans)(x,\mu,a) \, \pi_\theta(\d a|t,x,\mu)$. Here $\nabla_\theta \Sigma_\theta$ $=$ $(\frac{\partial \Sigma_\theta^{ij}}{\partial \theta_k})_{i,j,k}$ $\in$ $\R^{d\times d\times D}$ is a tensor of order $3$, and we used the product tensor notations $\bullet_1$  recalled in the introduction.  
\end{Theorem}
\begin{Remark}[On the martingale property of the policy gradient] 
The representation in Theorem \ref{theogradient} also means that the process 
\begin{align}
& \Big\{ e^{-\beta(s-t)}  \mrG_\theta(s,X_s^{t,x,\mu},\P_{X_s^{t,\xi}}) + \int_t^s  e^{-\beta(r-t)}  \nabla_\theta \log p_\theta(r,X_r^{t,x,\mu},\P_{X_r^{t,\xi}},\alpha_r) \Big\{ \d \mrV_\theta(r,X_r^{t,x,\mu},\P_{X_r^{t,\xi}})  \\
&   \hspace{1cm}  + \;  \big[  f(X_r^{t,x,\mu},\P_{X_r^{t,\xi}},\alpha_r) +  \lambda    \log p_\theta(r,X_r^{t,x,\mu},\P_{X_r^{t,\xi}},\alpha_r) - \beta \mrV_\theta(r,X_r^{t,x,\mu},\P_{X_r^{t,\xi}})   \big] \d r \Big\}  \\
& \hspace{2cm}  + \int_t^s e^{-\beta(r-t)}  \Hc_\theta[\mrV_\theta](r,X_r^{t,x,\mu},\P_{X_r^{t,\xi}})   \d r, \; t \leq s \leq T\Big\} 
\end{align} 
is a martingale, for any given $\alpha$ $\sim$ $\pi_\theta$. 
\end{Remark}
{\bf Proof.} See Appendix \ref{sec:theogradient}
\ep

\vspace{2mm}


In the next section, we show how the probabilistic representation formula of the gradient function $\mrG_\theta$ provided by Theorem \ref{theogradient} can be used to design two actor-critic algorithms for learning optimal cost function and randomised policy by relying on samples of the actions, states and state distributions.

\subsection{Actor-critic Algorithms}

Actor-critic (AC) methods combine policy gradient (PG) and performance evaluation (PE).  Compared to most existing works on RL for mean-field problems, mainly based on $Q$-learning (see e.g. \cite{carlautan19}, \cite{elieetal20}, \cite{guetal20})
we do not assume that the agent (the social planner) has at disposal a simulator for the state distribution, but instead will estimate the distribution of the population from the observation of the state of the representative player and by updating the distribution along repeated episodes. More precisely, for each episode $i$ 
$=$ $1,2,\ldots,N$,  from the observation of the state $X_{t_k}^i$ of a representative player $i$  at time $t_k$, we update the state distribution according to
\begin{align} \label{estimmu} 
\mu_{t_k}^i &= \;  (1-\rho^i_S) \mu_{t_k}^{i-1} + \rho^i_S \delta_{X_{t_k}^i},
\end{align} 
where $(\rho^i_S)_i$ is a sequence of learning parameters in $(0,1)$, e.g. $\rho^i_S$ $=$ $1/i$. It is expected from the propagation of chaos,  that when the number of episodes  $N$ goes to infinity, $\mu_{t_k}^N$ converge to the limiting distribution $\P_{X_{t_k}}$ of the population. 
Notice that a similar estimation procedure was  recently proposed in \cite{angfoulau21} in the context of a  MFC control problem in discrete time with finite state and action spaces over an infinite horizon.

In addition to the family of randomised policies $(t,x,\mu)$ $\mapsto$ $\pi_\theta(\d a|t,x,\mu)$ $=$ $p_\theta(t,x,\mu)\nu(\d a)$, with parameter $\theta$, we are given a family of  functions $(t,x,\mu)$ $\mapsto$  $\mrJ^\eta(t,x,\mu)$ on $[0,T]\times\R^d\times\Pc_2(\R^d)$, with parameter $\eta$, aiming to approximate the optimal cost value function. 
AC algorithm is then updating alternately the two parameters to find the optimal pair $(\theta^*,\eta^*)$, hence determining the approximate optimal randomised policy and the associated cost value function.  On the one hand, 
the loss function in the PE step  for learning $\mrJ^\eta$, for fixed policy $\pi_\theta$,  is based on the martingale formulation of the process
\beqs
\big\{ e^{-\beta t} \mrJ^\eta(t,X_t^{x,\mu},\P_{X_t^{\xi}}) + \int_0^t  e^{-\beta r} \big[  f(X_r^{x, \mu},\P_{X_r^{\xi}},\alpha_r)  +  \lambda \log p_\theta(r,X_r^{x,\mu},\P_{X_r^{\xi}},\alpha_r)  \big] \d r, \; 0 \leq t \leq T\big\}, 
\enqs
and on the other hand, the objective (here a cost) function in the PG step  for learning $\pi_\theta$, for fixed $\mrJ^\eta$, is based on the martingale formulation of the process
\begin{align} \label{marG} 
& \Big\{ e^{-\beta t} \mrG_\theta(t,X_t^{x,\mu},\P_{X_t^{\xi}}) + \int_0^t  e^{-\beta r}  \nabla_\theta \log p_\theta(r,X_r^{x,\mu},\P_{X_r^{\xi}},\alpha_r) \Big[ \d \mrJ^\eta(r,X_r^{x,\mu},\P_{X_r^{\xi}})  \\
&   \hspace{2mm}  + \;  \big(  f(X_r^{x,\mu},\P_{X_r^{\xi}},\alpha_r)  +  \lambda  \log p_\theta(r,X_r^{x,\mu},\P_{X_r^{\xi}},\alpha_r) - \beta \mrJ^\eta(r,X_r^{x,\mu},\P_{X_r^{\xi}}) \big) \d r \Big] \\
& \hspace{2cm} + \;  \Hc_\theta[\mrJ^\eta](r,X_r^{x,\mu},\P_{X_r^{\xi}})   \d r, \; 0 \leq t \leq T\Big\}.  
\end{align} 
Here, we denote $X^{x,\mu}$ $=$ $X^{0,x,\mu}$  (resp. $X^{\xi}$ $=$ $X^{0,\xi}$) when the initial time of the flow is $t$ $=$ $0$.  
We emphasise that these loss functions are minimised by training samples of the state trajectories $X_t^{x_0,\xi}$, actions $\alpha$ $\sim$ $\pi_\theta$,  estimation $\mu_t$ of $\P_{X_t^\xi}$ according to \eqref{estimmu}, and observation of the associated running and terminal costs. 

We  first develop  AC algorithms in the offline setting where all  state trajectories are sampled. In this case, given $\theta$, the proposed loss function for the PE step is
\beqs
L^{PE}(\eta) &=& \E_{\alpha\sim\pi_\theta} \Big[ \int_0^T \Big| e^{-\beta(T-t)} g(X_T,\P_{X_T}) \\
&&  + \int_t^T e^{-\beta(r-t)}  \big[ f(X_r,\P_{X_r},\alpha_r) +  \lambda \log p_\theta(r,X_r^{},\P_{X_r^{}},\alpha_r)\big] \d r - \mrJ^\eta(t,X_t,\P_{X_t}) \Big|^2 \d t \Big],
\enqs

\noindent which leads, after  time discretisation of $[0,T]$ on the grid $\{t_k = k \Delta t, k=0,\ldots,n\}$,  
and by applying stochastic gradient descent (SGD) with learning rate $\rho_E$,  to the following update rule: 
\beqs
\eta & \leftarrow & \eta + \rho_E \sum_{k=0}^{n-1} \Big( e^{-\beta(n-k)\Delta t} g_{t_n}   + \sum_{\ell=k}^{n-1}  e^{-\beta(\ell-k)\Delta t}  \big[ f_{t_\ell} +  \lambda \log p_\theta(t_\ell,X_{t_\ell},\mu_{t_\ell},\alpha_{t_\ell})\big] \Delta t \\
& & \hspace{3cm}  - \;  \mrJ^\eta(t_k,X_{t_k},\mu_{t_k}) \Big) \nabla_\eta \mrJ^\eta(t_k,X_{t_k},\mu_{t_k}) \Delta t, 
\enqs
where we set   $f_{t_\ell}$ $=$ $f(X_{t_\ell},\P_{X_{t_\ell}},\alpha_{t_\ell})$, as the output running cost at time $t_\ell$, for an input state $X_{t_l}$, action $\alpha_{t_\ell}$,   $\ell$ $=$ $0,\ldots,n-1$,   
and $g_T$ $=$ $g(X_T,\P_{X_T})$ the terminal cost for an input $X_T$.  
Given $\eta$, the learning in the PG step relies on the gradient representation \eqref{expressG}, and (after time discretisation) leads to the update rule
\beqs
\theta & \leftarrow & \theta -  \rho_G \hat G_\theta, \\
\mbox { with } \quad \hat G_\theta & = &  \sum_{k=0}^{n-1} e^{-\beta t_k}  \nabla_\theta \log p_\theta(t_k,X_{t_k},\mu_{t_k},\alpha_{t_k}) \Big[ \mrJ^\eta(t_{k+1},X_{t_{k+1}},\mu_{t_{k+1}}) - \mrJ^\eta(t_k,X_{t_k},\mu_{t_k}) \\
& & \hspace{2mm} + \;  \big( f_{t_k}    +  \lambda  \log p_\theta(t_k,X_{t_k},\mu_{t_k},\alpha_{t_k})    - \beta   \mrJ^\eta(t_k,X_{t_k},\mu_{t_k})  \big) \Delta t \Big]   +  \Hc_\theta[\mrJ^\eta](t_k,X_{t_k},\mu_{t_k})   \Delta t.  
\enqs
The pseudo-code is described in Algorithm \ref{offalgo}. 

\vspace{2mm}

\begin{algorithm2e}[H] 
\DontPrintSemicolon 
\SetAlgoLined 
\vspace{1mm}
{\bf Input data}: Number of episodes $N$, number of mesh time-grid $n$ ($\leftrightarrow$ time step $\Delta t$ $=$ $T/n$), learning rates  $\rho_S^i$, $\rho_E^i$, $\rho_G^i$ for the state distribution, PE and PG estimation, and function of the number of episodes $i$. 
Parameter $\lambda$ for entropy regularisation. 
Functional forms $\mrJ^\eta$ of cost value function,  $p_\theta$ of density policies. 

{\bf Initialisation}: $\mu_{t_k}$: state distribution on $\R^d$, for $k$ $=$ $0,\ldots,N$,  parameters $\eta$, $\theta$. 

\For{each episode $i$ $=$ $1,\ldots,N$}
{Initialise state $X_0$ $\sim$ $\mu_0$ \\
\For{$k$ $=$ $0,\ldots,n-1$} 
{Update state distribution: $\mu_{t_k}$ $\leftarrow$ $(1-\rho_S^i) \mu_{t_k} + \rho_S^i \delta_{X_{t_k}}$ \\
Generate action $\alpha_{t_k}$ $\sim$ $\pi_\theta(.|t_k,X_{t_k},\mu_{t_k})$ \\
Observe (e.g. by environment simulator) state $X_{t_{k+1}}$ and cost $f_{t_k}$ \\
If $k$ $=$ $n-1$, update terminal state distribution: $\mu_{t_{n}}$ $\leftarrow$ $(1-\rho_S) \mu_{t_{n}} + \rho_S \delta_{X_{t_{n}}}$, and observe terminal cost $g_{t_n}$ \\
$k$ $\leftarrow$ $k+1$
} 
{Compute
\beqs
\Delta_\eta &=& \sum_{k=0}^{n-1} \Big( e^{-\beta(n-k)\Delta t}  g_{t_n}   + \sum_{\ell=k}^{n-1} e^{-\beta(\ell-k)\Delta t}  \big[ f_{t_\ell}  +  \lambda  \log p_\theta(t_\ell,X_{t_\ell},\mu_{t_\ell},\alpha_{t_\ell})\big] \Delta t \\
& & \hspace{3cm}  - \;  \mrJ^\eta(t_k,X_{t_k},\mu_{t_k}) \Big) \nabla_\eta \mrJ^\eta(t_k,X_{t_k},\mu_{t_k}) \Delta t \\
\hat G_\theta & = &  \sum_{k=0}^{n-1} e^{-\beta t_k}  \nabla_\theta \log p_\theta(t_k,X_{t_k},\mu_{t_k},\alpha_{t_k}) \Big[ \mrJ^\eta(t_{k+1},X_{t_{k+1}},\mu_{t_{k+1}}) - \mrJ^\eta(t_k,X_{t_k},\mu_{t_k}) \\
& & \hspace{2mm} + \;   \big( f_{t_k}   +  \lambda    \log p_\theta(t_k,X_{t_k},\mu_{t_k},\alpha_{t_k}) - \beta \mrJ^\eta(t_k,X_{t_k},\mu_{t_k})  \big) \Delta t \Big]   +  \Hc_\theta[\mrJ^\eta](t_k,X_{t_k},\mu_{t_k})   \Delta t.  
\enqs
Critic Update: $\eta$ $\leftarrow$ $\eta + \rho_E^i \Delta _\eta$;  Actor Update: $\theta$ $\leftarrow$ $\theta  -   \rho_G^i \hat G_\theta$
}
}
{\bf Return}: $\mrJ^\eta$, $\pi_\theta$
\caption{Offline actor-critic mean-field algorithm  \label{offalgo} }
\end{algorithm2e}

\vspace{5mm}

We next develop AC algorithm for online setting where only past sample trajectory is available, and so the parameters $(\theta,\eta)$ are updated in real-time incrementally. In this case, given a policy $\pi_\theta$, we consider  at each time step $t_k$, $k$ $=$ $0,\ldots,n-1$, a loss function for  PE  
given by 
\beqs
L_{t_k}^{PE}(\eta) &=& \E_{\alpha\sim\pi_\theta} \Big[ \Big|  \mrJ^\eta(t_{k+1},X_{t_{k+1}},\P_{X_{t_{k+1}}})  - \mrJ^\eta(t_k,X_{t_k},\P_{X_{t_k}}) \\
& & \hspace{3mm} + \;  \big( f(X_{t_k},\P_{X_{t_k}},\alpha_{t_k})   +   \lambda   \log p_\theta(t_k,X_{t_k},\P_{X_{t_k}},\alpha_{t_k})  - \beta \mrJ^\eta(t_k,X_{t_k},\P_{X_{t_k}})\big) \Delta t \Big|^2 \Big].
\enqs
Concerning PG, we note that when $\theta$ is an optimal parameter, we should have $\mrG_\theta$ $=$ $0$.  Therefore, from the martingale condition in \eqref{marG}, this suggests to find $\theta$ such that at any time  $t_k$, $k$ $=$ $0,\ldots,n-1$
\beqs
 \E_{\alpha\sim\pi_\theta} \Big\{ \nabla_\theta \log p_\theta(t_k,X_{t_k},\P_{X_{t_k}},\alpha_{t_k})  \big[ \mrJ^\eta(t_{k+1},X_{t_{k+1}},\P_{X_{t_{k+1}}})  - \mrJ^\eta(t_k,X_{t_k},\P_{X_{t_k}})  &   & \\
  + \;  \big( f(X_{t_k},\P_{X_{t_k}},\alpha_{t_k})   +  \lambda   \log p_\theta(t_k,X_{t_k},\P_{X_{t_k}},\alpha_{t_k}) - \beta  \mrJ^\eta(t_k,X_{t_k},\P_{X_{t_k}})  \big) \Delta t   \big] + \Hc_\theta[\mrJ^\eta](t_k,X_{t_k},\mu_{t_k})   \Delta t \Big\}  &  & \\  
   =  0. &  &
\enqs
The pseudo-code is described in Algorithm \ref{onalgo}.

\vspace{2mm}

\begin{algorithm2e}[H] 
\DontPrintSemicolon 
\SetAlgoLined 
\vspace{1mm}
{\bf Input data}: Number of episodes $N$, number of mesh time-grid $n$ ($\leftrightarrow$ time step $\Delta t$ $=$ $T/n$), learning rates  $\rho_S^i$, $\rho_E^i$, $\rho_G^i$ for the state distribution, PE and PG estimation, and function of the number of episodes $i$.  Parameter $\lambda$ for entropy regularisation. 
Functional forms $\mrJ^\eta$ of cost value function,  $p_\theta$ of density policies. 

{\bf Initialisation}: $\mu_{t_k}$: state distribution on $\R^d$, for $k$ $=$ $0,\ldots,n$,  parameters $\eta$, $\theta$. 

\For{each episode $i$ $=$ $1,\ldots,N$}
{Initialise state $X_0$ $\sim$ $\mu_0$ \\
\For{$k$ $=$ $0,\ldots,n-1$} 
{Update state distribution: $\mu_{t_k}$ $\leftarrow$ $(1-\rho_S^i) \mu_{t_k} + \rho_S^i \delta_{X_{t_k}}$ \\
Generate action $\alpha_{t_k}$ $\sim$ $\pi_\theta(.|t_k,X_{t_k},\mu_{t_k})$ \\
Observe (e.g. by environment simulator) state $X_{t_{k+1}}$ and cost  $f_{t_k}$ \\
If $k$ $=$ $n-1$, update terminal state distribution: $\mu_{t_{k+1}}$ $\leftarrow$ $(1-\rho_S^i) \mu_{t_{k+1}} + \rho_S^i \delta_{X_{t_{k+1}}}$, and observe terminal cost $g_{t_{k+1}}$ \\

Compute
\beqs
\delta_\eta &=&  \mrJ^\eta(t_{k+1},X_{t_{k+1}},\mu_{t_{k+1}}) - \mrJ^\eta(t_k,X_{t_k},\mu_{{t_k}})   \\
& & \quad \quad +  \big( \; f_{t_k}  +  \lambda   \log p_\theta(t_k,X_{t_k},\mu_{{t_k}},\alpha_{t_k})  - \beta  \mrJ^\eta(t_k,X_{t_k},\mu_{{t_k}})  \big) \Delta t     \\
\Delta_\eta &=& \delta_\eta  \nabla_\eta \mrJ^\eta(t_k,X_{t_k},\mu_{{t_k}}) \\
\Delta_\theta &=& \delta_\eta \nabla_\theta  \log p_\theta(t_k,X_{t_k},\mu_{t_k},\alpha_{t_k}) +   \Hc_\theta[\mrJ^\eta](t_k,X_{t_k},\mu_{t_k})  \Delta t, 
\enqs
with the constraint that when $k$ $=$ $n-1$,  $\mrJ^\eta(t_{k+1},X_{t_{k+1}},\mu_{t_{k+1}})$ $=$ $g_{t_{k+1}}$.

Critic Update: $\eta$ $\leftarrow$ $\eta + \rho_E^i \Delta _\eta$;  Actor Update: $\theta$ $\leftarrow$ $\theta -  \rho_G^i  \Delta_\theta$ \\
$k$ $\leftarrow$ $k+1$
}
}
{\bf Return}: $\mrJ^\eta$, $\pi_\theta$
\caption{Online actor-critic mean-field algorithm  \label{onalgo} }
\end{algorithm2e}

\begin{Remark}[About the choice of actor and critic parametric functions]  \label{remchoice} 
In the Actor-critic algorithms, we have to specify a parametric family of  randomised policies $\pi_\theta$, and  a parametric family of critic functions $\mrJ^\eta$. In general, for critic functions, one can consider cylindrical neural network functions in the form
\begin{align} \label{cyl} 
\mrJ^\eta(t,x,\mu) &=\;  \Psi(t,x, <\varphi,\mu>), \quad (t,x,\mu) \in [0,T]\times\R^d\times\Pc_2(\R^d),
\end{align} 
where $\Psi$ is a feedforward neural network from $[0,T]\times\R^d\times\R^k$ into $\R$, and $\varphi$ is another feedforward neural network from $\R^d$ into $\R^k$ (called latent space), and we use the notation 
$<\phi,\mu>$ $:=$ $\int \phi(x)\mu(\d x)$.  The set of parameters $\eta$ is the union of the parameter sets for the two  neural networks $\Psi$ and $\varphi$. 
This choice is motivated by the density property of the set of cylindrical functions, i.e. functions in the form  \eqref{cyl} with continuous functions $\Psi$ and $\varphi$, with respect to continuous functions on $[0,T]\times\R^d\times\Pc_2(\R^d)$ as  shown  in \cite{guophawei21}, 
and the universal approximation property  of feedforward neural networks on finite-dimensional space, see \cite{hor89}. 

Concerning the policies, notice that when the temperature parameter  for exploration $\lambda$ is zero, the optimal policy is of pure (non randomised)  feedback form as a function of $(t,x,\mu)$. When 
$\lambda$ $>$ $0$, the optimal policy is in general truly randomised, and the larger is $\lambda$, the larger is the exploration in the sense that the variance of the randomised policy increases.  We can  then take for the parametric family of randomised policies,  for example Gaussian distributions: 
\beqs
\pi_\theta(.|t,x,\mu) & =& \Nc \big( \mrm(t,x,\mu); \vartheta(\lambda) \big), 
\enqs
where $\mrm$ is a cylindrical neural network function on $[0,T]\times\R^d\times\Pc_2(\R^d)$ valued in $A$ $\subset$ $\R^m$, and $\vartheta(.)$ is a given symmetric matrix-valued function, nondecreasing w.r.t. $\lambda$, with $\vartheta(\lambda)$ positive-definite for $\lambda$ $>$ $0$, and $\vartheta(0)$ $=$ $0$. 

In some particular mean-field models, we may know {\it a priori} the structural form of the optimal value function and optimal randomised policy, and this suggests alternately some specific form for the parametric family of actor and critic functions. This is typically the case of the linear quadratic model, as presented in the next section. 
\end{Remark}

\begin{Remark} 
The above actor-critic algorithms involve the computation of the term $\Hc_\theta[\mrJ^\eta]$ at each time $t_k$, and along the observed state $X_{t_k}$ and estimated state distribution $\mu_{t_k}$. This additional term, compared to the actor-critic algorithms designed in \cite{jiazhou21}  for standard stochastic control without mean-field interaction, involves the operator $\Hc_\theta$ defined in \eqref{defH}.  In the separable form case, namely when the coefficients of the mean-field process are in the form
\beqs
b(x,\mu,a) \; = \; \mrb(x,\mu) +  C(a), & & (\sigma\sigma\trans)(x,\mu,a) \; = \;  \Sigma(x,\mu) + F(a), 
\enqs
where $C$ and $F$ are known functions from $A$ into $\R^d$, resp. $\R^{d\times d}$, we notice that 
\beqs
\nabla_\theta b_\theta(t,x,\mu) & = &  \nabla_\theta C_\theta(t,x,\mu), \quad  \mbox{ with }  \quad  C_\theta(t,x,\mu) \; := \; \int C(a)   \pi_\theta(\d a|t,x,\mu), \\
\nabla_\theta \Sigma_\theta(t,x,\mu) & = & \nabla_\theta F_\theta(t,x,\mu),  \quad \mbox{ with }  \quad   F_\theta(t,x,\mu) \; := \; \int  F(a)   \pi_\theta(\d a|t,x,\mu), 
\enqs
are known functions,  and consequently  also the function $\Hc_\theta[\mrJ^\eta]$.  Another important case where the term $\Hc_\theta[\mrJ^\eta]$ is a known computable function is given in the linear quadratic framework as presented in the next section. 
\end{Remark}

\section{The linear quadratic case} \label{sec:LQ}

We focus on  the important class of MFC control problem with linear state dynamics and quadratic reward, namely 
\begin{equation} \label{LQmodel} 
\begin{cases}
b(x,\mu,a) \; = \;  B x + \bar B \bar\mu +  C a,  \quad  \sigma(x,\mu,a) \; = \;  \gamma + D x + \bar D \bar\mu + Fa,  \\
f(x,\mu,a) \; = \; x\trans Qx + \bar\mu\trans \bar Q\bar\mu  + a\trans N a  + 2a\trans Ix + 2a\trans \bar I \bar\mu +  2M.x + 2H.a, \\
g(x,\mu) \; = \; x\trans P x + \bar\mu\trans \bar P\bar\mu + 2L.x,
\end{cases} 
\end{equation} 
for $(x,\mu,a)$ $\in$ $\R^d\times\Pc_2(\R^d)\times\R^m$, where we denote by $\bar\mu$ $=$ $\int x \mu(\d x)$,  $B$, $\bar B$, $D$, $\bar D$ are constant matrices in $\R^{d\times d}$, $C$, $F$ are constant matrices in $\R^{d\times m}$, $\gamma$ is a constant in $\R^d$, 
$N$ is a  symmetric matrix in $\S_+^m$, $I$, $\bar I$ $\in$ $\R^{m\times d}$,  $Q$, $\bar Q$, $P$, $\bar P$ are  symmetric matrices in  $\S^d$,  with $Q$ $\geq$ $0$, $P$ $\geq$ $0$, 
$M$, $L$  $\in$ $\R^d$, $H$ $\in$ $\R^m$. 

In this case,  the optimal value function to this LQ MFC problem with entropy regularisation when minimizing over randomised controls a functional cost as in \eqref{defVpi},   is given by 
\begin{align}\label{vquadra} 
v(t,x,\mu) &=\;  (x - \bar\mu)\trans K(t) (x-\bar\mu) + \bar\mu\trans \Lambda(t)\bar\mu + 2 Y(t).x + R(t), 
\end{align} 
where $K$ (valued in $\S^d$), $\Lambda$ (valued in $\S^d$), $Y$ valued in $\R^d$, and $R$ valued in $\R$, are solutions to a system of ordinary differential equations on $[0,T]$ given in \eqref{ODEKdiscount}. 
Moreover,  the optimal randomised control is of feedback form with  Gaussian distribution:
\beqs
\pi^*(.|t,x,\mu) & =  &  \Nc \Big( - S(t)^{-1} \big(  U(t) x + (W(t) - U(t))\bar\mu + O(t) \big); \frac{\lambda}{2} S(t)^{-1} \Big), 
\enqs
where 
\beqs
S(t) \; = \; N + F\trans K(t) F, &  & O(t) \; = \;  H + C\trans Y(t)  + F\trans K(t) \gamma \\
U(t) \; = \; I + C\trans K(t) + F\trans K(t) D,  & & W(t) \; = \; I + \bar I + C \trans \Lambda(t) + F\trans K(t) (D + \bar D). 
\enqs
This is an extension of  the mean-field LQ control without entropy and control randomization, and the proof that adapts arguments in \cite{baspha19} is reported in Appendix \ref{sec:LQappen}.

 \vspace{1mm}

In a RL setting, the coefficients of the LQ model \eqref{LQmodel}  
are unknown, thus  $K$, $\Lambda$, $Y$, and $R$ cannot be solved from the system of ODEs, and $S$, $O$, $U$, and $W$ are also unknown.  We shall then employ our RL algorithms to solve the LQ problem in a model-free setting. 
In view of the above structure of the optimal value function and randomised policy, we parametrise the cost  value function by 
\begin{align} \label{paramJ} 
\mrJ^\eta(t,x,\mu) &=\;  (x - \bar\mu)\trans K^\eta(t) (x-\bar\mu) + \bar\mu\trans \Lambda^\eta(t)\bar\mu + 2 Y^\eta(t).x + R^\eta(t), 
\end{align} 
for some parametric functions $K^\eta$, $\Lambda^\eta$, $Y^\eta$, $R^\eta$ on $[0,T]$, with parameters $\eta$ $\in$ $\R^p$. On the other hand, we parametrise the randomised policies by 
\begin{align} \label{parampi} 
\pi_\theta(.|t,x,\mu) &= \;  \Nc\big(  \phi_1^\theta(t) x + \phi_2^\theta(t)\bar\mu + \phi_3^\theta(t); \Sigma^\theta(t) \big),
\end{align} 
for some parametric functions $\phi_1^\theta,\phi_2^\theta,\phi_3^\theta,\Sigma^\theta$ on $[0,T]$, with parameter $\theta$ $\in$ $\R^q$. 

The parametric functions  $K^\eta$, $\Lambda^\eta$, $Y^\eta$, $R^\eta$, and  $\phi_1^\theta,\phi_2^\theta,\phi_3^\theta,\Sigma^\theta$, could be  in general neural networks on $[0,T]$, but depending on the examples, we could take more specific forms, as discussed in the next section.

For parametrisation of the cost  value function and randomised policies as in \eqref{paramJ}, \eqref{parampi}, we  see that 
\beqs
\partial_\mu \mrJ^\eta(t,x,\mu)(x') &=& 
 -2 K^\eta(t)(x-\bar\mu) + 2 \Lambda\eta\bar\mu,  \;  \mbox{ and so } \partial_{x'} \partial_\mu  \mrJ^\eta(t,x,\mu)(x') = 0,  \\ 
\nabla_\theta b_\theta(t,x,\mu) & = &  C \nabla_\theta \phi_1^\theta(t) \bullet_2 x + C \nabla_\theta \phi_2^\theta(t) \bullet_2  \bar\mu + C \nabla_\theta \phi_3^\theta(t)
\enqs
and then  
\beqs
\Hc_\theta[\mrJ^\eta](t,x,\mu) &=& 2  \big[ (\nabla_\theta \phi_1^\theta(t) + \nabla_\theta \phi_2^\theta(t)) \bullet_2  \bar\mu + \nabla_\theta \phi_3^\theta(t) \big]\trans C\trans  \big( - K^\eta(t)(x-\bar\mu) +  \Lambda^\eta\bar\mu \big),
\enqs
which only involves, up to the knowledge of $C$, known functions of $(t,x,\mu)$. Notice also that when $\phi_1^\theta$ $=$ $-\phi_2^\theta$, and $\phi_3^\theta$ $\equiv$ $0$ (see below  the example of mean-field systemic risk), then $\Hc_\theta[\mrJ^\eta]$ $\equiv$ $0$.

\section{Numerical examples} \label{sec:num}

\subsection{Example 1: mean-field systemic risk}

 We consider a mean-field model of systemic risk introduced in \cite{carfousun15}. This fits into a LQ MFC with 
 \beqs \label{eq: example 1}
 \bar B = -  B > 0, \; C=1, \;  \gamma > 0, \; D = \bar D = F = 0 \\
I = - \bar I >0, \;  Q + \bar Q = 0, \;  N = \frac{1}{2}, \; M=H=L=0, \; P + \bar P = 0,
 \enqs
and $Q$ $\geq$ $2I^2$. We also take $X_0 \sim \Nc(0,1)$. In this case, the solution to the system of ODEs \eqref{ODEKdiscount}  yields the analytic expression: 
\beqs
K(t) &=& - \frac{1}{2} \Big[ \bar B + 2I - \sqrt{\Delta} \frac{ \sqrt{\Delta} \sinh(\sqrt{\Delta}(T-t))  + (\bar B + 2I  + 2P) \cosh (\sqrt{\Delta}(T-t))}{ \sqrt{\Delta} \cosh(\sqrt{\Delta}(T-t))  + 
(\bar B + 2I + 2P) \sinh (\sqrt{\Delta}(T-t))} \Big], \\
R(t) &=& \frac{\gamma^2}{2} \ln \Big[ \cosh (\sqrt{\Delta}(T-t)) + \frac{\bar B + 2I  + 2P}{\sqrt{\Delta}}  \sinh (\sqrt{\Delta}(T-t)) \Big] - \frac{\gamma^2}{2} (\bar B + 2I)(T-t) \\
& & -\frac{\lambda (T-t)}{2}\log(2\pi\lambda)
\enqs
with $\sqrt{\Delta}$ $=$ $\sqrt{(\bar B+2I)^2  + 2Q -  4I^2}$,  and $\Lambda$ $=$ $Y$ $=$ $0$, while the optimal randomised policy is given by
\beqs
\hat\pi(.|t,x,\mu) &=& \Nc \big( \phi(t)(x - \bar\mu) ; \lambda \big), \quad \mbox{ with } \;  \phi(t) \; = \;  -2(K(t) + I). 
\enqs 

In view of these expressions, we shall use  critic function as
\beqs
\mrJ^\eta(t,x,\mu) &=& K^\eta(t)  (x - \bar\mu)^2  + R^\eta(t), 
\enqs
for some parametric functions $K^\eta$ and $R^\eta$ on $[0,T]$ with parameters $\eta$, and actor functions as 
\beqs
\pi_\theta(.|t,x,\mu) &=& \Nc\big(  \phi^\theta(t) (x -\bar\mu); \lambda  \big),  \\
\mbox{ i.e. } \quad  \log p_\theta(t,x,\mu,a)  &=&  - \frac{1}{2}\log(2\pi\lambda) - 
\frac{\big|a - \phi^\theta(t)(x-\bar\mu)\big|^2}{2\lambda},
\enqs
for some parametric function $\phi^\theta$ on $[0,T]$ with parameter $\theta$.   As shown in Section \ref{sec:LQ}, we notice that $\Hc_\theta[\mrJ^\eta]$ $=$ $0$.

We shall test with two choices of parametric functions: 
\begin{enumerate}
\item {\it Exact parametrisation:}
\begin{equation} \label{exactsys} 
\begin{cases}
K^\eta(t) \; = \;   - \frac{1}{2} \Big[ \eta_3  - \eta_1 \frac{  \sinh(\eta_1(T-t))  + \eta_2  \cosh (\eta_1(T-t))}{  \cosh(\eta_1(T-t))  + 
\eta_2  \sinh (\eta_1(T-t))} \Big], \\
R^\eta(t) \; = \;  \eta_4 \ln \Big[ \cosh (\eta_1(T-t)) + \eta_2  \sinh (\eta_1(T-t)) \Big] -  \eta_3 \eta_4(T-t) -\frac{\lambda (T-t)}{2}\log(2\pi\lambda) \\
\phi^\theta(t)  \; = \;   \theta_3  - \theta_1 \frac{  \sinh(\theta_1(T-t))  + \theta_2  \cosh (\theta_1(T-t))}{  \cosh(\theta_1(T-t))  + 
\theta_2  \sinh (\theta_1(T-t))},
\end{cases}
\end{equation}
with parameters $\eta$ $=$ $(\eta_1,\eta_2,\eta_3,\eta_4)$ $\in$ $\R_+^4$, and $\theta$ $=$ $(\theta_1,\theta_2,\theta_3)$ $\in$ $\R_+^3$, so that the optimal solution in the model-based case corresponds to 
$\eta_1^*$ $=$ $\sqrt{\Delta}$, $\eta_2^*$ $=$ $\bar B + 2I  + 2P$, $\eta_3^*$ $=$ $\bar B + 2I$, $\eta_4^*$ $=$ $\gamma^2/2$, and $\theta_1^*$ $=$ $\sqrt{\Delta}$, $\theta_2^*$ $=$ $\bar B + 2I  + 2P$, $\theta_3^*$ $=$ $\bar B$. 
\item  {\it Neural networks:}  for $K^\eta$, $R^\eta$ and $\phi^\theta$, with time input. 
\end{enumerate}

We implement our actor-critic algorithms with a simulator of $X$ for coefficients equal to
\beqs
T = 1, \; \gamma = 1, \; \bar B = - B = 0.6, \;  I=0.4, \; P=Q=1,  
\enqs
 The simulator for  $X$ is  based on the real mean-field model:
\beqs
\d X_t &=& \big( \bar B(\E[X_t] -  X_t) + \alpha_t \big) \d t + \gamma \d W_t. 
\enqs
Since $\alpha$ $\sim$ $\pi^\theta$, we note that $\E[\alpha_t]$ $=$ $\phi^\theta(t)(\E[X_t] - \E[\bar\mu_t])$ $=$ $0$.  We deduce that under such $\alpha$, $\d \E[X_t]$ $=$ $0$, hence $\E[X_t]$ $=$ $\E[X_0]$.  From the above mean-field dynamics of $X$, we deduce that 
\beqs
X_{t_{k+1}} - \E[X_0] &=&  e^{-\bar B\Delta t} (X_{t_k} - \E[X_0]) +   \alpha_{t_k} \big(  \frac{ 1 - e^{-\bar B\Delta t}}{\bar B} \big) + \gamma \int_{t_k}^{t_{k+1}} e^{-\bar B(t_{k+1} - s)} \d W_s \\
& \simeq & e^{B\Delta t} (X_{t_k} - \E[X_0]) +   \alpha_{t_k} \big(  \frac{ 1 - e^{-\bar B\Delta t}}{\bar B} \big) + \gamma e^{-\bar B\Delta t} \Delta W_{t_k}. 
\enqs
The cost is simulated according to
\beqs
f_{t_k} &=& Q(X_{t_k} - \E[X_{0}])^2 + \frac{1}{2} \alpha_{t_k}^2 + 2 \alpha_{t_k} I (X_{t_k} - \E[X_{0}]), \quad g_T \; = \; P (X_T - \E[X_0])^2. 
\enqs

\vspace{1mm}

We first present the numerical results of our offline Algorithm \ref{offalgo} when using the exact parametrisation  \eqref{exactsys}.  The derivatives  w.r.t. to  $\eta$ of $K^\eta$, $R^\eta$, hence of $\mrJ^\eta$, as well as the derivative w.r.t. $\theta$ of $\log p_\theta$ have explicit analytic expressions that are implemented in 
the updating rule of the actor-critic algorithm. 

Here we used the following parameters: 
$\mu_{t_k}$ was initialized at $0$; the number of episodes was $N=2100$; the time horizon was $T=1$ and the time step $\Delta t= 0.02$. The values of the model parameters were as described above.
The learning rates $(\rho_S, \rho_E, \rho_G)$ and $\lambda$ were taken as $\rho_S=0.2$ constant, and at iteration $i$, 

$$
    \rho_E(i) = 
    \begin{cases}
            (0.01,0.1,0.01,0.2) \hbox { if } i \le 500
            \\
            (0.1,0.1,0.1,0.1) \hbox{ if } 500 < i \le 21000
    \end{cases}
    \quad 
    \rho_G(i) = 
    \begin{cases}
            (0.03,0.05,0.03) \hbox { if } i \le 7000
            \\
            (0.01,0.01,0.01) \hbox{ if } 7000 < i \le 10000
            \\
            (0.005,0.01,0.005)  \hbox{ if } 10000 < i \le 14000
            \\
            (0.002,0.002,0.002)   \hbox{ if } 17000 < i \le 21000
    \end{cases}
$$
and
$$
    \lambda(i) = 
    \begin{cases}
        0.1 \hbox{ if } i \le 8000, \\
        0.01 \hbox{ if } 8000 < i \le 14000, \\
        0.001 \hbox { if } 14000 < i   \le 21000
    \end{cases}
$$
Moreover, after $i=14000$ iterations, we also increase the size of the minibatch from $20$ to $40$. 
In Table \ref{tab:ex1exact}, we give the learnt parameters for the critic and actor functions, to be compared wih 
the exact value of the  parameters.  

\begin{table}[H]
\centering
\begin{tabular}{ |c|c|c|c|c|c|c|c|c| } 
 \hline
  & $\eta_1$ & $\eta_2$ & $\eta_3$ & $\eta_4$ & $\theta_1$ & $\theta_2$ & $\theta_3$ \\ 
  \hline
  exact & $1.8221$ & $1.8660$ & $1.4$ & $0.5$ & $1.8221$ & $1.8660$ & $0.6$  \\ 
  \hline
  learnt & $1.4197$ & $2.0536$ & $0.9997$ & $0.4824$ & $1.6204$ & $1.9167$  & $0.3660$  \\ 
 \hline
\end{tabular}
\caption{Learnt vs exact parameters of the critic and actor functions.} \label{tab:ex1exact}  
\end{table} 

\begin{figure}[H] 
\centering 
   \begin{minipage}[b]{0.45\linewidth}
  \centering
 \includegraphics[width=\textwidth, height=4.5cm]{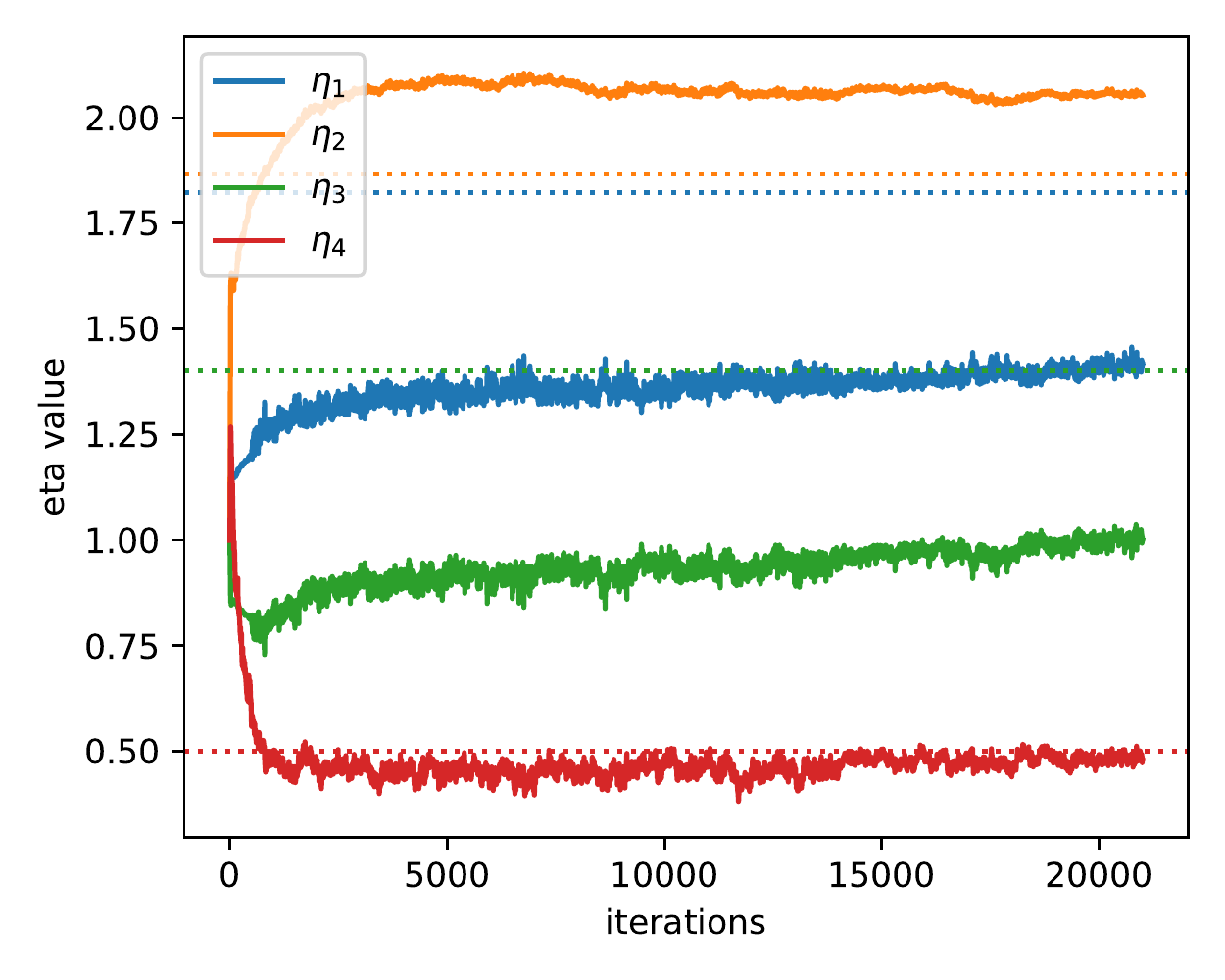}
 \end{minipage}
 \begin{minipage}[b]{0.45\linewidth}
  \centering
 \includegraphics[width=\textwidth, height=4.5cm]{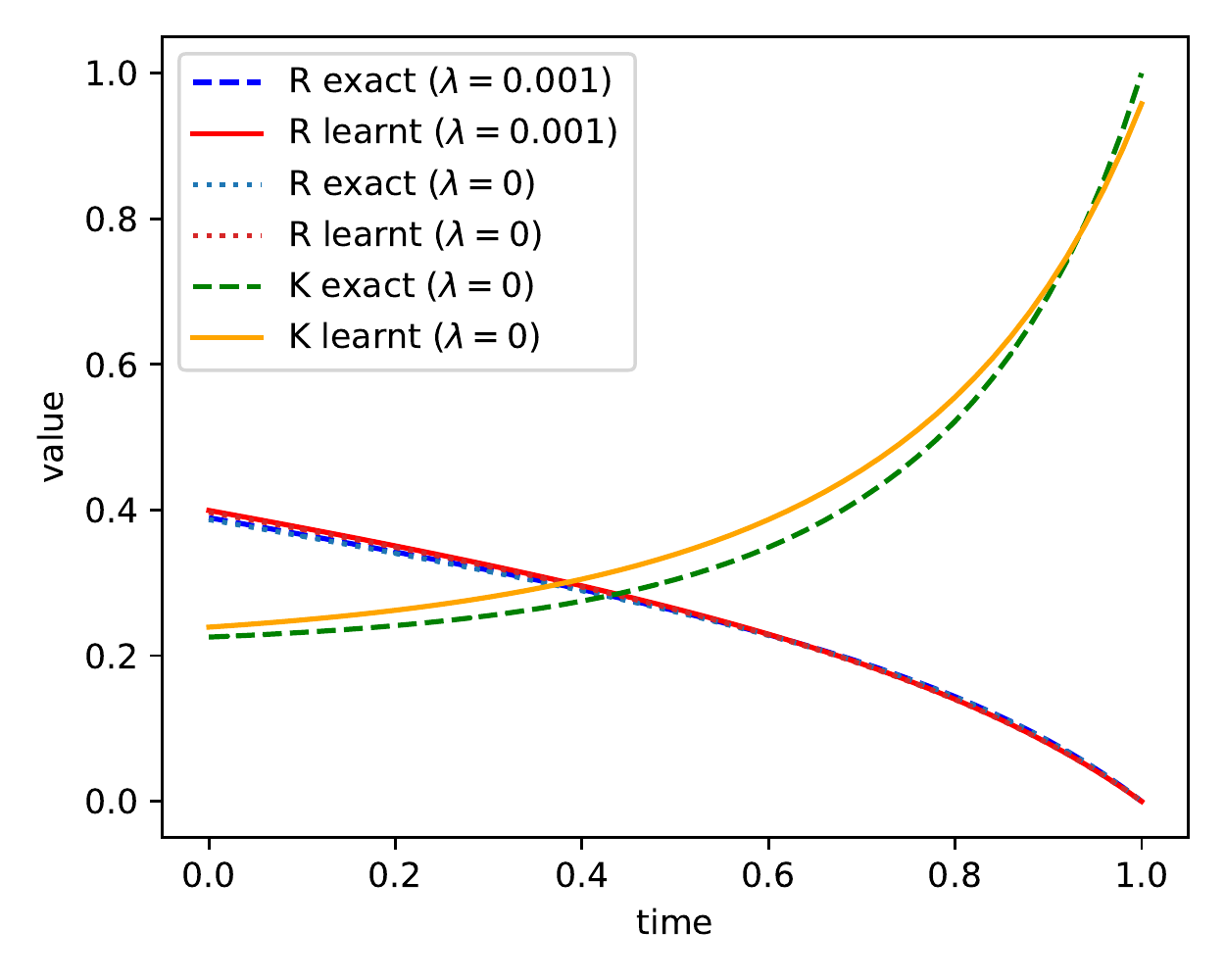}
 \end{minipage}
 
 \begin{minipage}[b]{0.45\linewidth}
  \centering
 \includegraphics[width=\textwidth, height=4.5cm]{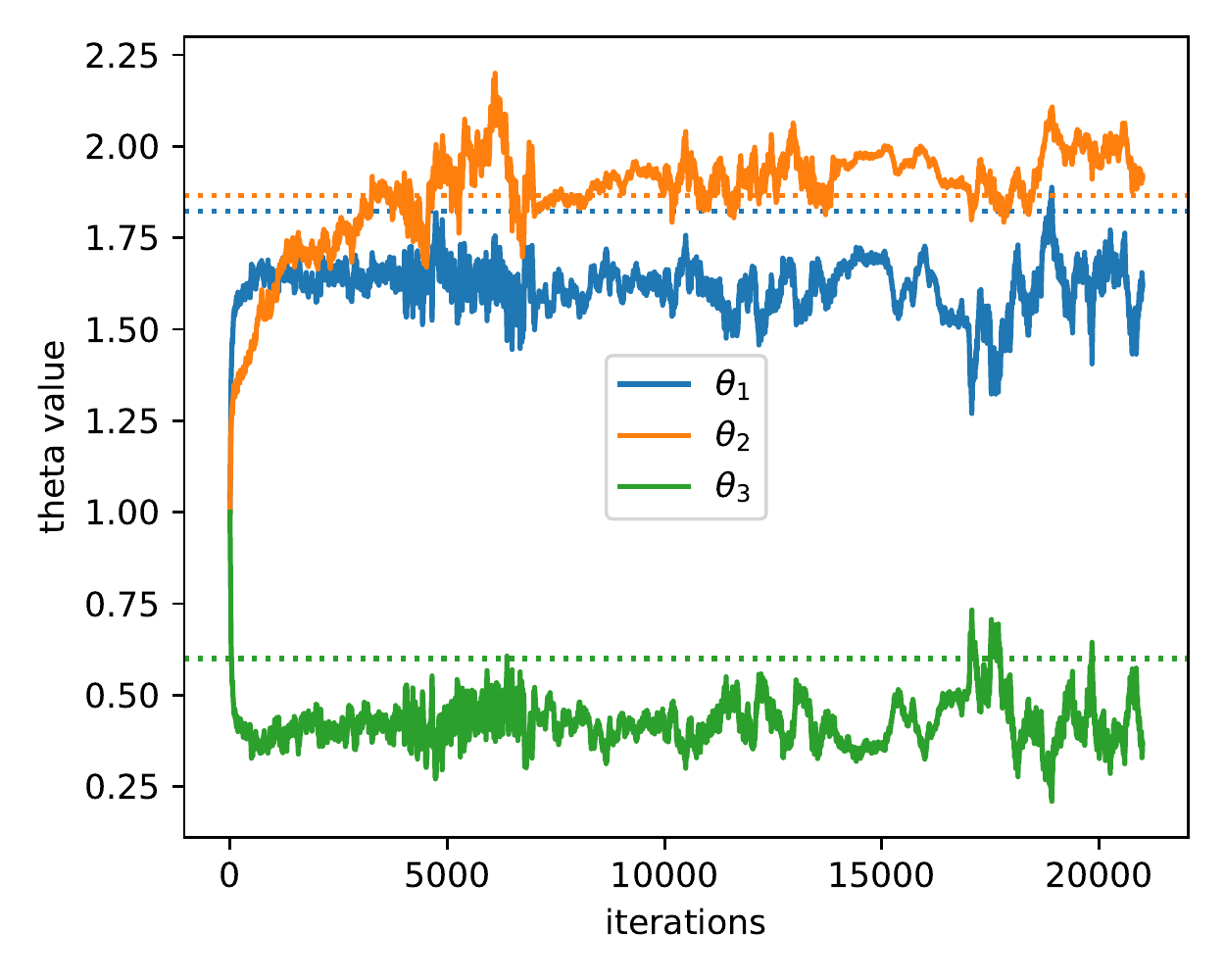}
 \end{minipage}
 \begin{minipage}[b]{0.45\linewidth}
  \centering
 \includegraphics[width=\textwidth,height=4.5cm]{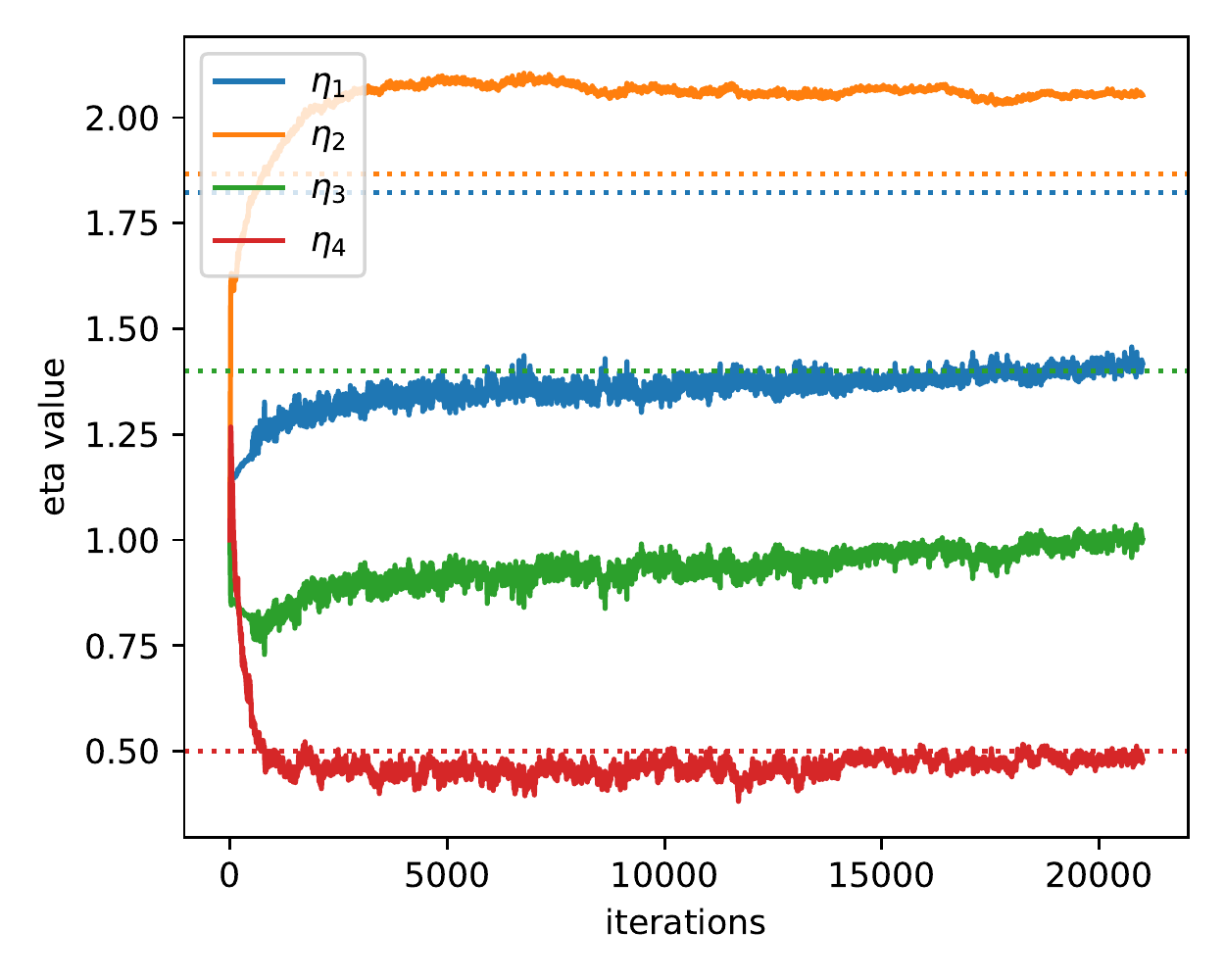}
 \end{minipage}
 

 \begin{minipage}[b]{0.45\linewidth}
  \centering 
 \includegraphics[width=\textwidth]{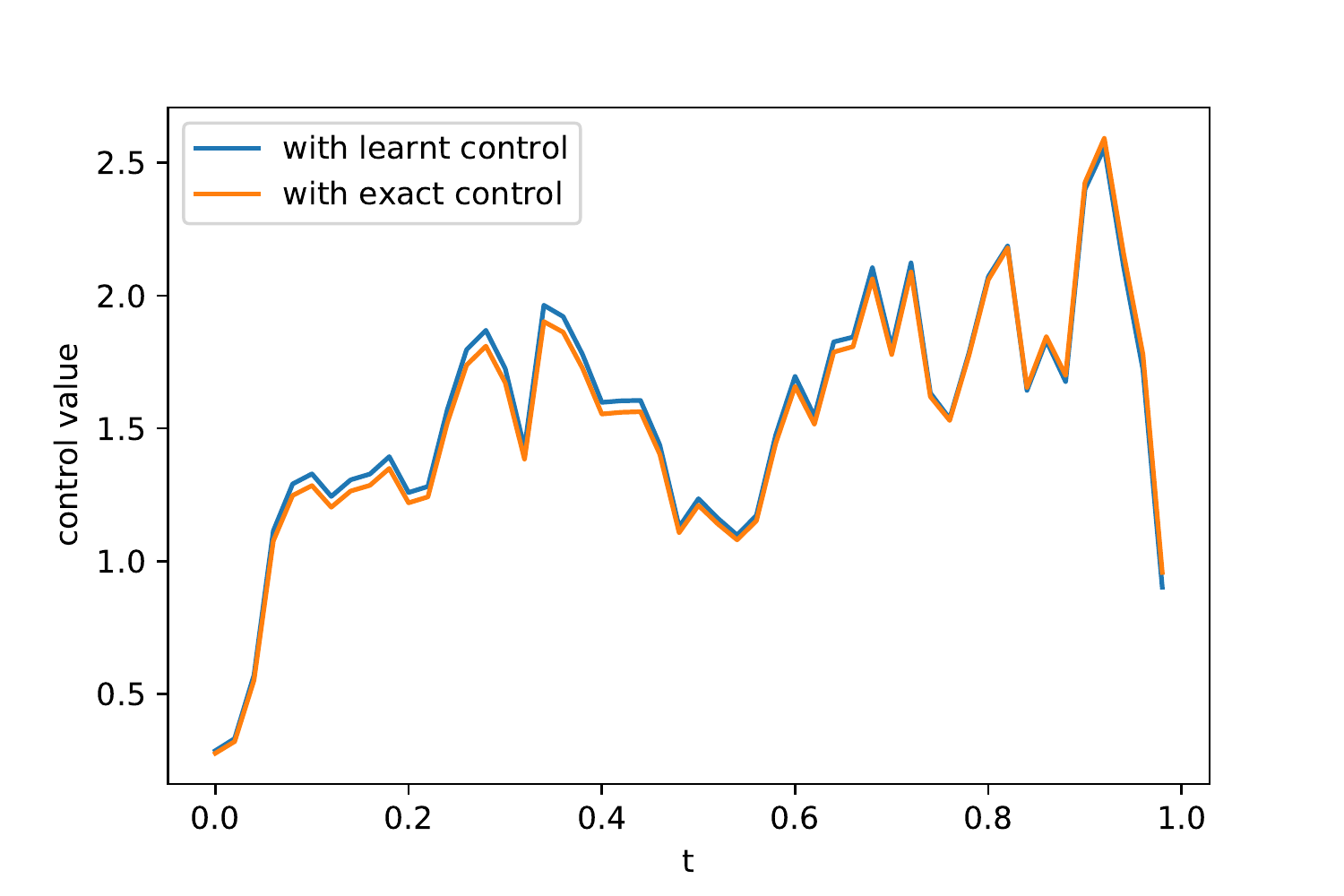}
 \end{minipage}
 \begin{minipage}[b]{0.45\linewidth}
  \centering 
 \includegraphics[width=\textwidth]{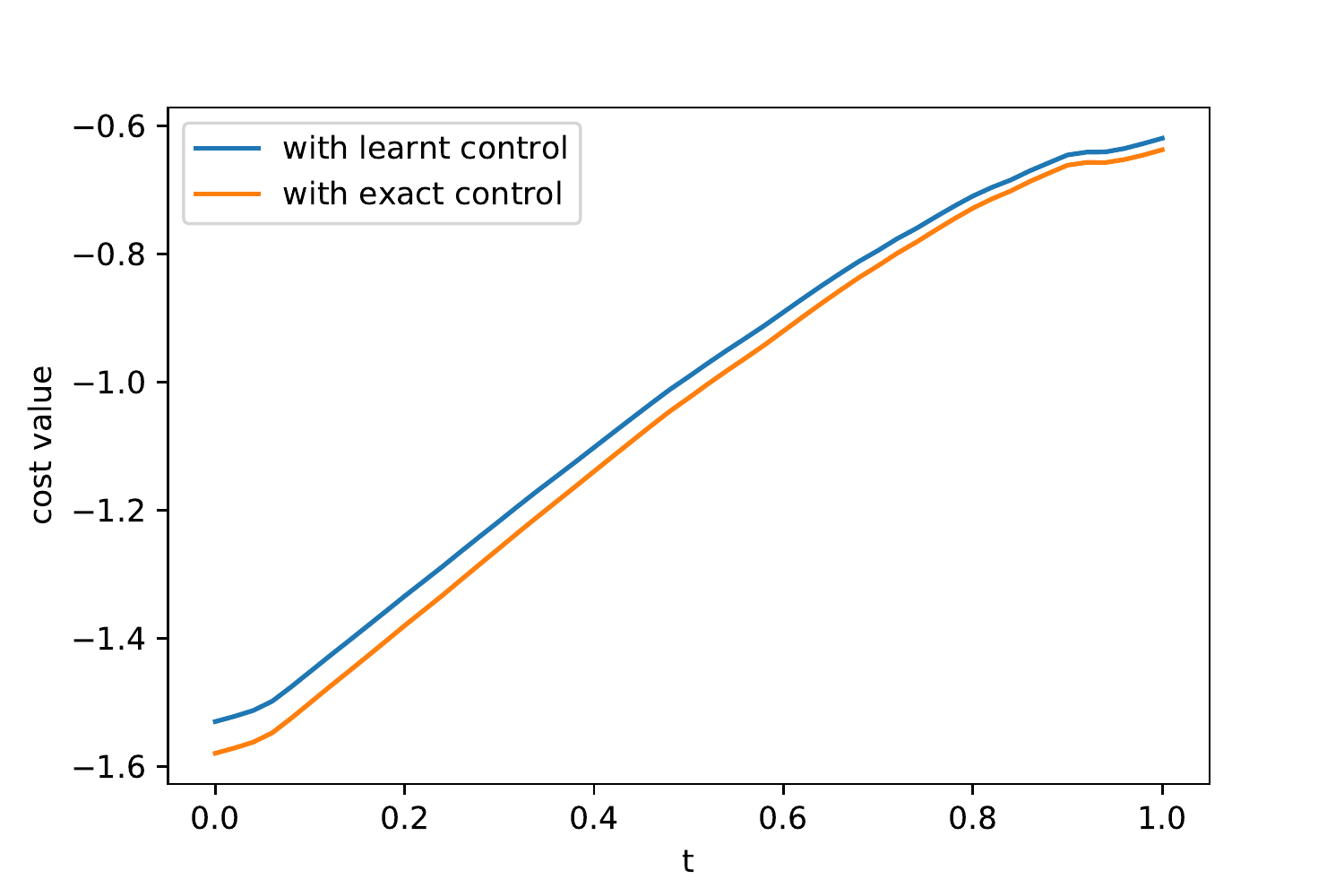}
 \end{minipage}
 
 \caption{\footnotesize{{\bf Convergence of the learnt value function and policy with exact parametrisation for the offline Algorithm \ref{offalgo}.}   {\it Top}: learned parameters of critic (left) and associated critic functions $K^\eta$, $R^\eta$ (right) vs optimal parameters and associated functions. {\it Middle: } learned parameters of actor (left) and associated actor function $\phi^\theta$ vs optimal parameters and associated function.  
 {\it Bottom:} one realization of the control (left) and one realization of the cost (right) vs the optimal ones, respectively along a state trajectory controlled by the learnt control and a state trajectory controlled optimally (both using the same realization of the Brownian motion). 
 }} \label{fig:algo1exactparam} 
\end{figure}

 In Figure~\ref{fig:algo1exactparam}, we see that, even though the parameters $\eta$ and $\theta$ (shown with full lines) are slightly different from the true optimal values (shown in dashed lines), the functions $K, R$ and $\phi$ are matched almost perfectly. 
 
We also display one realization of the control and of the cost. These are based on evaluating the control and the cumulative cost along one trajectory of the state. We first simulate $10^4$ realizations of a Brownian motion. Based on this, we generate trajectories for one $10^4$ population of agents using the learnt control and one population of $10^4$ agents using the optimal control. For the population that uses the learnt control, the control is given by the mean of the actor, namely, $\phi^\theta(t) (x -\bar\mu)$. In the dynamics, the cost and the control, the mean field term is replaced by the empirical mean of the corresponding population at the current time. We can see that the trajectories of control (resp. cost) are very similar. 

\begin{figure}[H]
   \begin{minipage}[b]{0.45\linewidth}
  \centering
 \includegraphics[width=\textwidth]{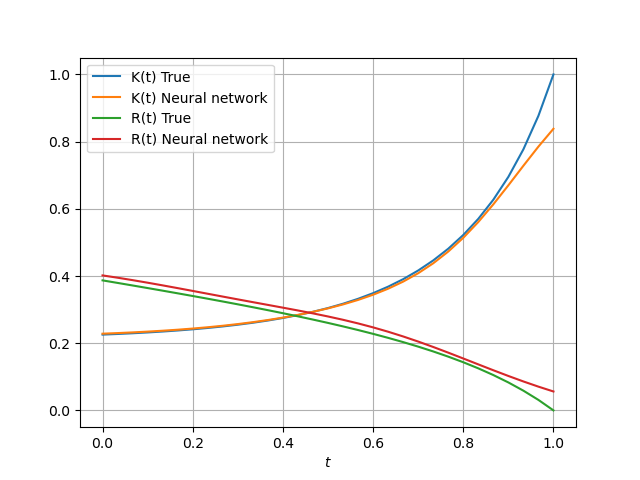}
 \end{minipage}
\begin{minipage}[b]{0.45\linewidth}
  \centering
 \includegraphics[width=\textwidth]{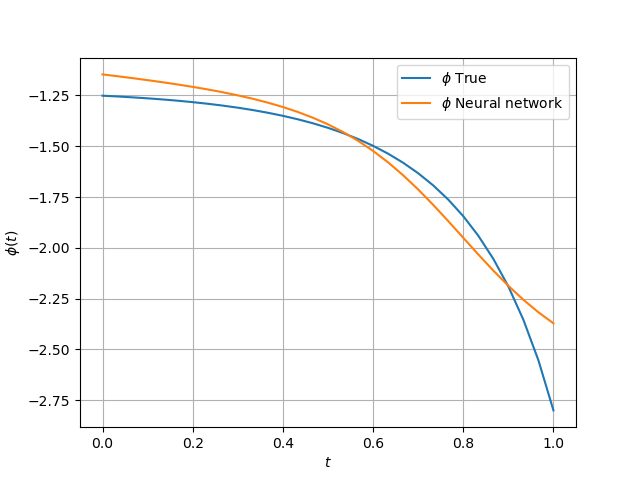}
 \end{minipage}
 \caption{\footnotesize{{\bf Learnt critic cost  function with neural networks for the online Algorithm \ref{onalgo}}. {\it Left panel}: Neural network functions $K^\eta,R^\eta$ vs optimal one. {\it Right panel}: Neural network function $\phi^\theta$ vs optimal one. }  } \label{fig:algo2NNcritic}
\end{figure}

\begin{figure}[H]
   \begin{minipage}[b]{0.45\linewidth}
  \centering
 \includegraphics[width=\textwidth]{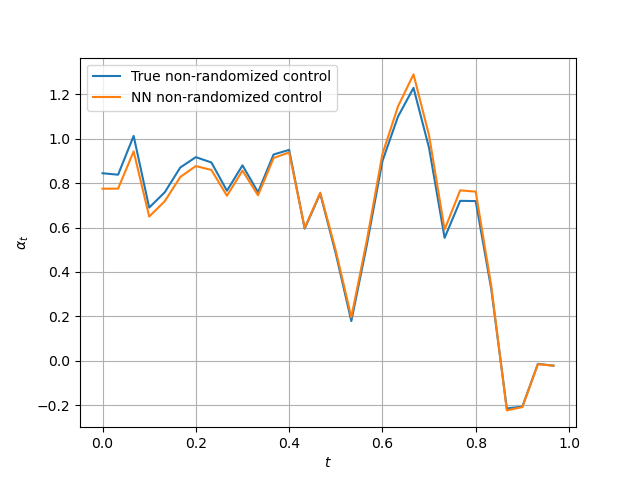}
 \end{minipage}
\begin{minipage}[b]{0.45\linewidth}
  \centering
 \includegraphics[width=\textwidth]{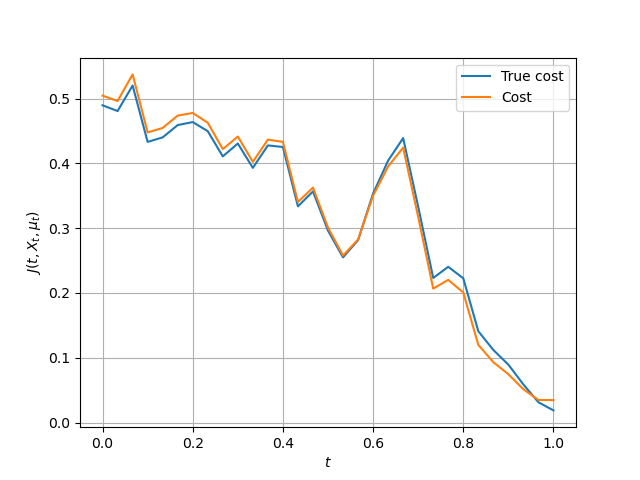}
 \end{minipage}
 \caption{\footnotesize{{\bf Learnt actor policy function with neural networks for the online Algorithm \ref{onalgo}}. {\it Left panel}:  One realization of the control vs the optimal non-randomised one with $\lambda=0$
 {\it Right panel}: } Plot of one realization of  $t$ $\mapsto$ $\mrJ^\eta(t,X_t,\P_{X_t})$, 
 respectively along a state trajectory controlled by the learnt policy  and a state trajectory controlled optimally (both using the same realization of the Brownian motion).
 } 
 \label{fig:algo2NNactor}
\end{figure}

 Next, we present in Figure \ref{fig:algo2NNcritic} and  Figure \ref{fig:algo2NNactor}  the numerical results of our online Algorithm \ref{onalgo} when using neural networks. In this case, the derivatives  w.r.t. to  $\eta$ of $K^\eta$, $R^\eta$, hence of $\mrJ^\eta$, as well as the derivative w.r.t. $\theta$ of $\log p_\theta$ 
 are computed by automatic differentiation. 
 We use neural networks with 3 hidden layers, 10 neurons per layer and tanh activation functions. We take $n=30$, $N=15000$ iterations, batch size 500 (10000 for the law estimation in the simulator), constant learning rates $10^{-3}$, except $\omega_S = 1$.  
  We change $\lambda$ along episodes: $\lambda$ $=$ $0.1$ for the first 3334 ones, then $0.01$ for the next 3333 ones, then $0.001$ until the end.


Finally, we test  in Table \ref{table:cost} the learnt policies 
from the exact and NN parametrisation by computing the associated initial expected social costs.  
We simulate $10$ populations, each consisting of $10^4$ agents. All the agents use the control function with the parameters learnt by the algorithm. For the dynamics, the cost and the control, the mean field term is replaced by the empirical mean of the corresponding population at the present time step. For each population, we compute the social cost. We then average over the $10$ populations in order to get a Monte Carlo estimate of the social cost. We report in the table the value of this average social cost, the standard deviation over the $10$ populations, and the relative error between the average social cost and the optimal cost computed by the formula $K^{\eta^*}_0 {\rm Var}(X_0) +R^{\eta^*}_0$ with the optimal parameter $\eta^*$.

\begin{table}[H]
\centering
\begin{tabular}{|c|c|c|}
\hline
& Initial cost (Std dev.) & Rel. error \\
\hline
Learnt with exact parameterization &  $0.625$ ($0.006$) &  $2.00\%$   \\
\hline
Learnt with NN & $0.632$  ($0.006$) &  $3.10\%$ \\
\hline
Exact value &  0.613  & \\
\hline 
\end{tabular}
\caption{Initial costs when following learnt policies vs optimal ones
} \label{table:cost} 
\end{table}

\subsection{Example 2: optimal trading}

We consider an optimal trading problem where the inventory is governed by 
\beqs
\d X_t &=& \alpha_t \d t + \gamma \d W_t, 
\enqs
and we aim to minimize over randomised trading rate $\alpha$ $\sim$ $\pi$ the cost functional
\beqs
\E \Big[ \int_0^T \alpha_t^2 + 2 H \alpha_t  - \lambda \Ec(\pi_t) \d t + P {\rm Var}(X_T) \Big]. 
\enqs
where $\gamma$ $>$ $0$, $H$ $>$ $0$ is the transaction price per trading,  $P$ $>$ $0$ is a risk aversion parameter, and $\lambda$ $>$ $0$ is the temperature parameter.  
This model fits into the LQ framework, and the solution to the system of ODEs \eqref{ODEKdiscount} is given by 
\beqs
K(t)   \; = \;  \frac{P}{1 + P(T-t)},  & &   R(t) \; = \;  \gamma^2 \log (1 + P(T-t)) - \big(H^2 + \frac{\lambda}{2} \log( \pi\lambda) \big) (T-t),
\enqs
$\Lambda$ $=$ $Y$ $\equiv$ $0$, 
while the optimal randomised policy is given by 
\beqs
\hat\pi(•|t,x,\mu) & \sim & \Nc \Big(  - K(t) (x - \bar\mu)  - H; \frac{\lambda}{2} \Big). 
\enqs

In a RL setting, the coefficients $\sigma$, $H$ and $P$ are unknown, and we use critic function as
\beqs
\mrJ^\eta(t,x,\mu) &=& K^\eta(t)  (x - \bar\mu)^2  + R^\eta(t), 
\enqs
for some parametric functions $K^\eta$ and $R^\eta$ on $[0,T]$ with parameters $\eta$, and actor functions as 
\beqs
\pi_\theta(.|t,x,\mu) &=& \Nc\big(  \phi^\theta(t) (x -\bar\mu) + \phi_3^\theta(t); \frac{\lambda}{2}     \big),  \\
\mbox{ i.e. } \quad  \log p_\theta(t,x,\mu,a)  &=&  - \frac{1}{2}\log(\pi\lambda) - 
\frac{\big|a - \phi^\theta(t)(x-\bar\mu)- \phi_3^\theta(t) \big|^2}{\lambda},
\enqs
for some parametric functions $\phi^\theta$, $\phi_3^\theta$  on $[0,T]$ with parameter $\theta$.  Given such  family of parametric actor/critic functions, we have 
 \beqs
 \Hc_\theta[\mrJ^\eta](t,x,\mu)   &=& 
-  2  C K^\eta(t)(x-\bar\mu)  \nabla_\theta \phi_3^\theta(t). 
 \enqs
We shall test with two choices of parametric functions: 
\begin{enumerate}
\item {\it Exact parametrisation:}
\begin{equation} \label{exactex3} 
\begin{cases}
K^\eta(t)  \; = \;   \frac{\eta_1}{1 + \eta_1(T-t)} \\
R^\eta(t) \; = \;   \eta_2  \log (1 + \eta_1(T-t)) - \big( \eta_3  + \frac{\lambda}{2} \log( \pi\lambda) \big) (T-t)\\
\phi^\theta(t) \; = \;    - \frac{\theta_1}{1 + \theta_1(T-t)}, \quad \phi_3^\theta(t) \; = \;  -  \theta_2, 
\end{cases}
\end{equation} 
with parameters $\eta$ $=$ $(\eta_1,\eta_2,\eta_3)$  $\in$ $(0,\infty)^3$, $\theta$ $=$ $(\theta_1,\theta_2)$ $\in$ $(0,\infty)^2$, so that the optimal solution in the model-based case corresponds to 
$(\eta_1^*,\eta_2^*,\eta_3^*)$ $=$ $(P,\gamma^2,H^2)$,  and $(\theta_1^*,\theta_2^*)$  $=$  $(P,H)$.    
\item  {\it Neural networks:}  for $K^\eta$, $R^\eta$,  $\phi^\theta$, and $\phi_3^\theta$  with time input.  Actually, we take for $\phi_3^\theta$ a constant function. 
\end{enumerate}


We first  present the numerical results of our offline Algorithm \ref{offalgo} when using the exact parametrisation \eqref{exactex3}. 
The derivatives  w.r.t. to  $\eta$ of $K^\eta$, $R^\eta$, hence of $\mrJ^\eta$, as well as the derivative w.r.t. $\theta$ of $\log p_\theta$, and $\Hc_\theta[\mrJ^\eta]$  have explicit analytic expressions that are implemented in 
the updating rule of the actor-critic algorithm. 
Here we used the following parameters: the learning rates $(\rho_S, \rho_E, \rho_G)$ and $\lambda$ were taken as $\rho_S=0.2$ constant,
 and at iteration $i$,
 $$
    \rho_E(i) = 
    \begin{cases}
            (0.05,0.05,0.05) \hbox { if } i \le 8000
            \\
            (0.05,0.05,0.01) \hbox{ if } 8000 < i \le 20000
    \end{cases}
    \quad 
    \rho_G(i) = 
    \begin{cases}
            (0.005,0.005) \hbox { if } i \le 8000
            \\
            (0.001,0.001) \hbox{ if } 8000 < i \le 13000
            \\
            (0.0005,0.0005) \hbox{ if } 8000 < i \le 13000
    \end{cases}
$$
and
$$
    \lambda(i) = 
    \begin{cases}
        0.1 \hbox{ if } i \le 8000, \\
        0.01 \hbox { if } 8000 < i \le 13000 \\
        0.001 \hbox { if } 13000 < i \le 20000
    \end{cases}
$$
$\mu_{t_k}$ was initialized at $0$; the number of episodes was $N=20000$; the time horizon was $T=1$ and the time step $\Delta t= 0.02$. The values of the model parameters are: $P=3$, $H=2$, $\gamma=1$, and $X_0 \sim \Nc(1,1)$.

In Table \ref{tab:ex2exact}, we give the learnt parameters for the critic and actor function to be compared with the exact values, when using the learnt policy with learnt empirical distribution from the algorithm.

\begin{table}[H]
\centering
\begin{tabular}{ |c|c|c|c|c|c| } 
 \hline
  & $\eta_1$ & $\eta_2$ & $\eta_3$ & $\theta_1$ & $\theta_2$  \\ 
  \hline
  exact & $3$ & $1$ & $4$ & $3$ & $2$  \\ 
  \hline
  learnt & $2.9864$ & $0.9637$ & $3.9154$ & $3.0161$ & $2.0016$     \\ 
 \hline
\end{tabular}
\caption{Learnt vs exact parameters of the critic and actor functions. } \label{tab:ex2exact} 
\end{table}

\begin{figure}[H] 
\centering 
   \begin{minipage}[b]{0.45\linewidth}
  \centering 
 \includegraphics[width=\textwidth]{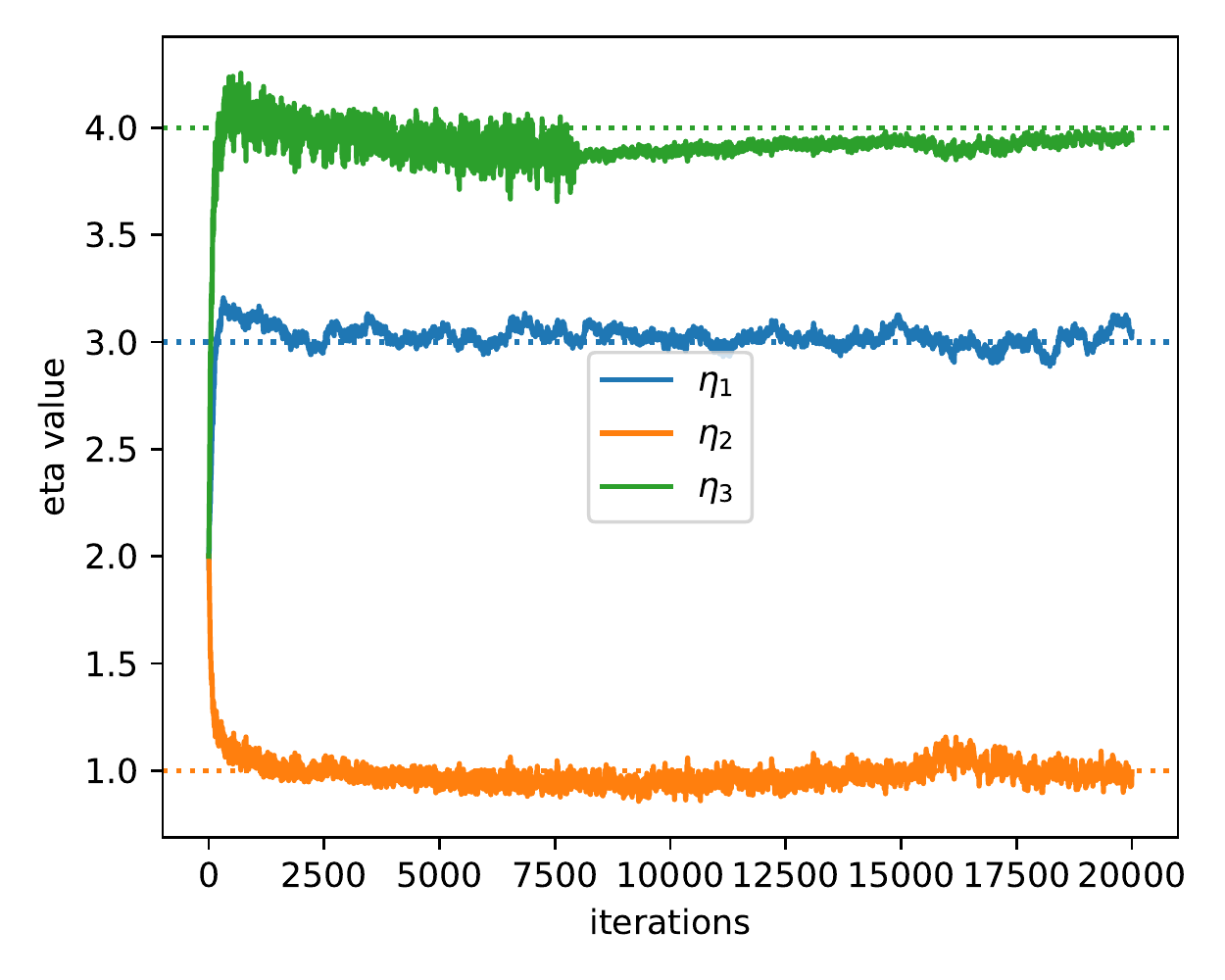}
 \end{minipage}
 \begin{minipage}[b]{0.45\linewidth}
  \centering 
 \includegraphics[width=\textwidth]{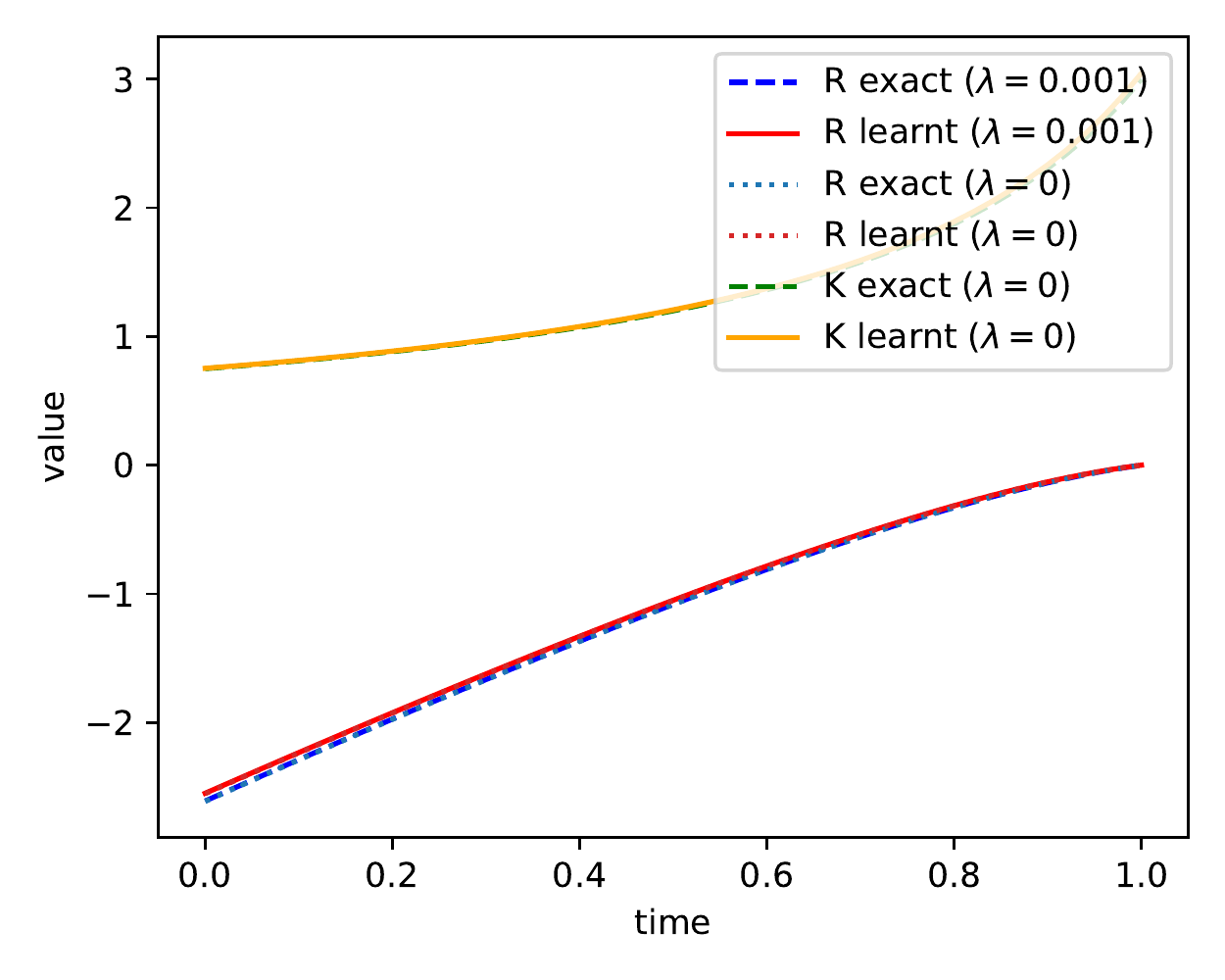}
 \end{minipage}
 
 \begin{minipage}[b]{0.45\linewidth}
  \centering 
 \includegraphics[width=\textwidth]{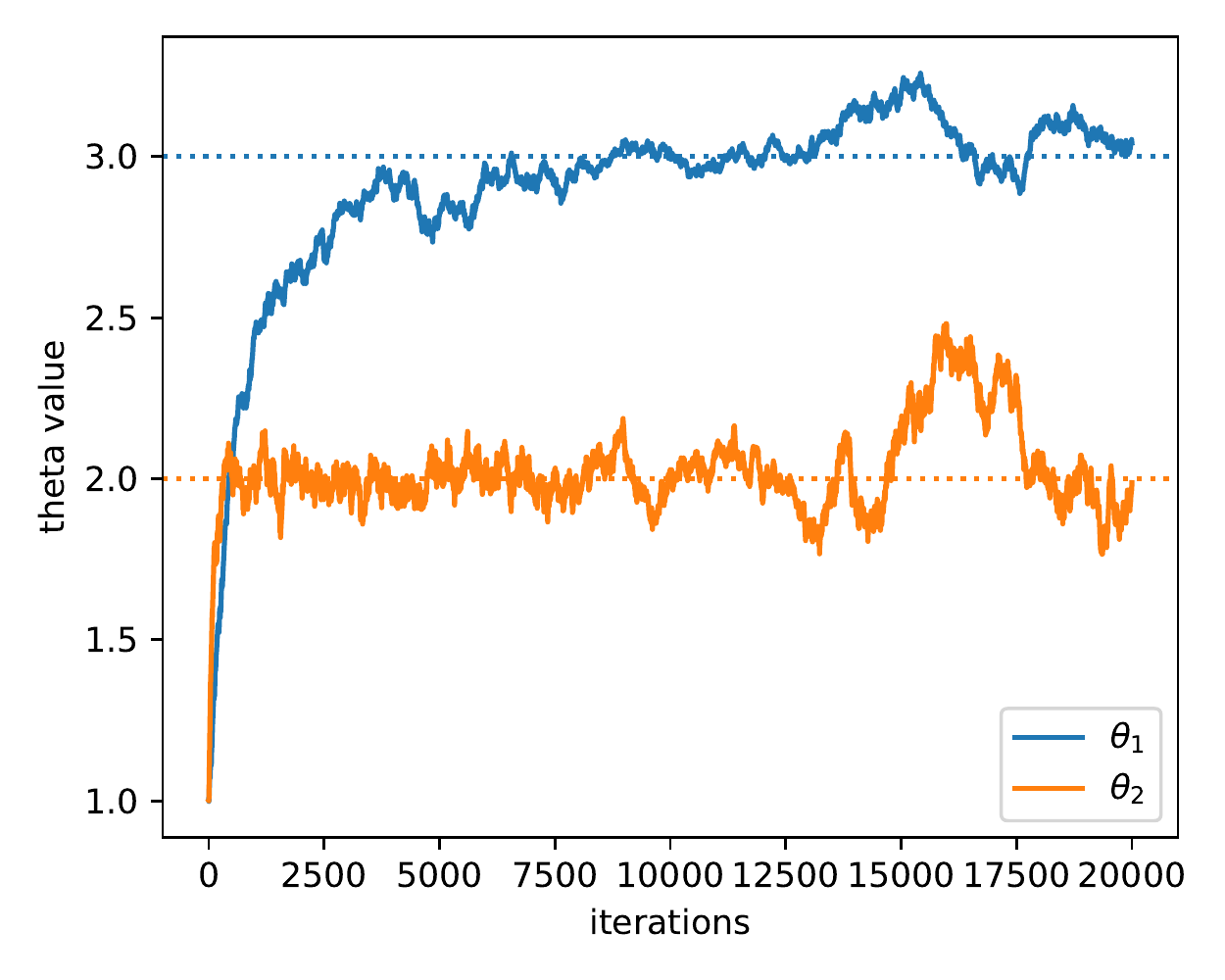}
 \end{minipage}
 \begin{minipage}[b]{0.45\linewidth}
  \centering 
 \includegraphics[width=\textwidth]{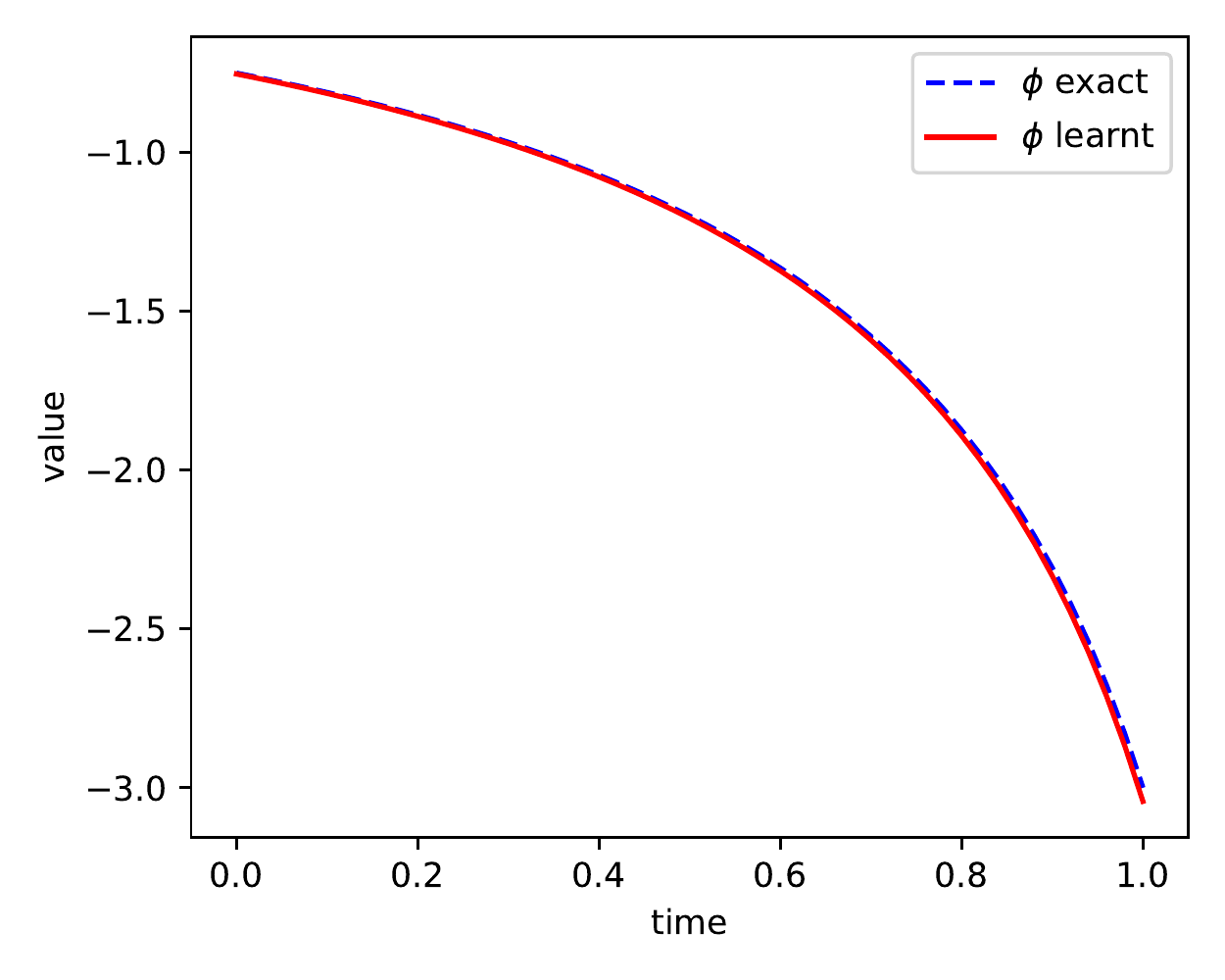}
 \end{minipage}
 
 \begin{minipage}[b]{0.45\linewidth}
  \centering 
 \includegraphics[width=\textwidth]{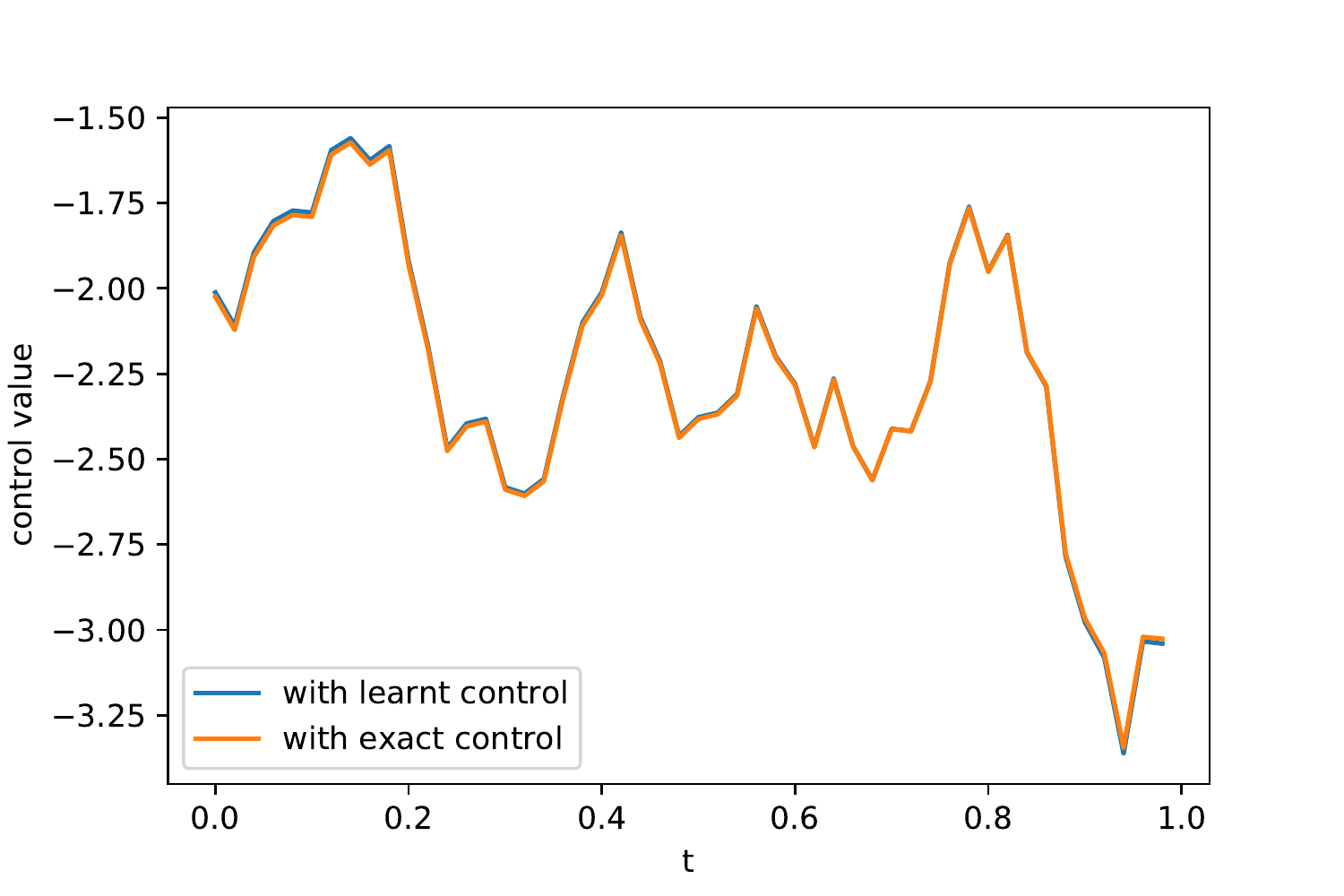}
 \end{minipage}
 \begin{minipage}[b]{0.45\linewidth}
  \centering 
 \includegraphics[width=\textwidth]{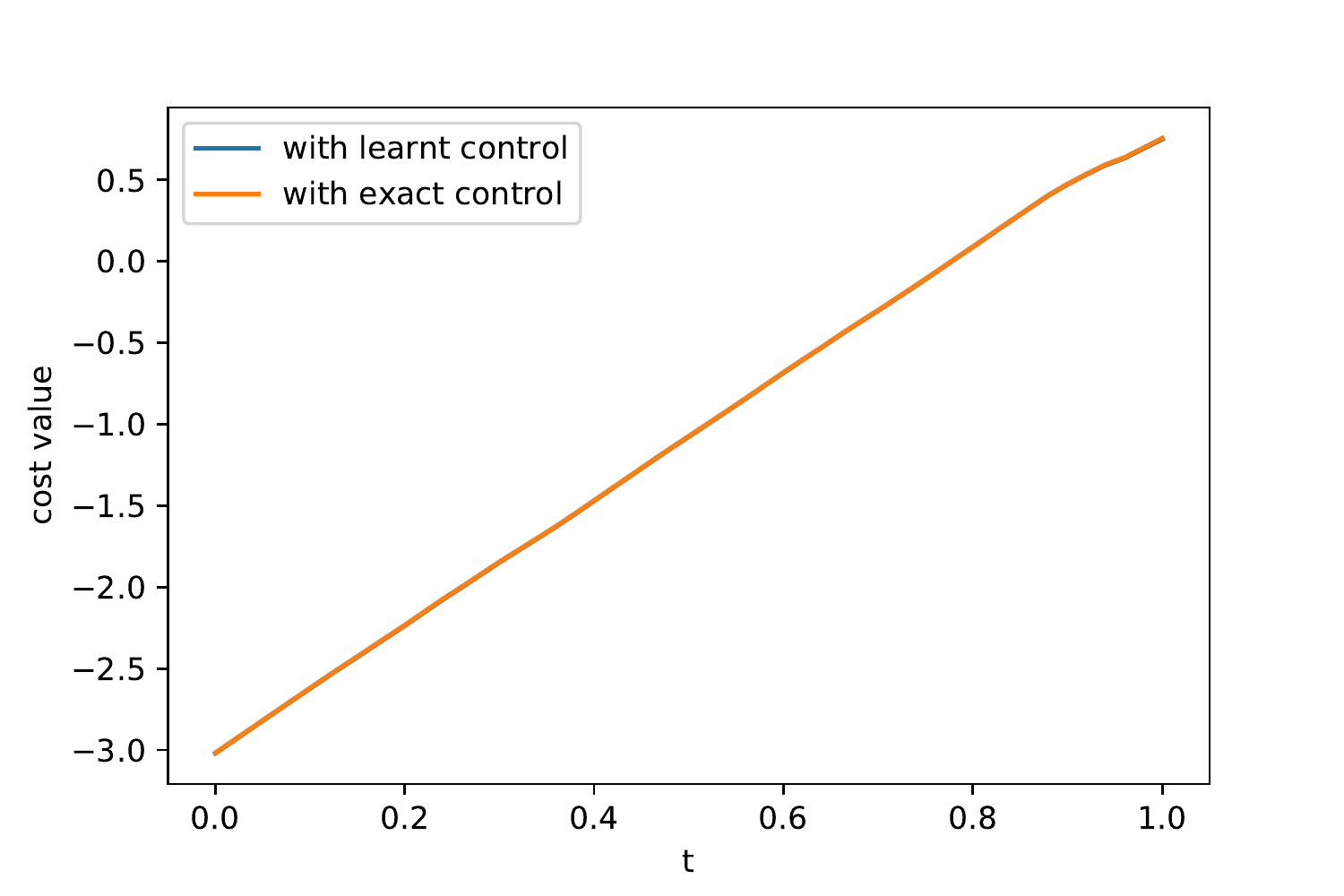}
 \end{minipage}
 \caption{\footnotesize{{\bf Convergence of the learnt value function and policy with exact parametrisation for the offline Algorithm \ref{offalgo}.}  
  {\it Top}: learned parameters of critic (left) and associated critic functions $K^\eta$, $R^\eta$ (right) vs optimal parameters and associated functions. {\it Middle: } learned parameters of actor (left) and associated actor function $\phi^\theta$ vs optimal parameters and associated function.  
 {\it Bottom:} one realization of the control (left) and one realization of the cost (right) vs the optimal ones, respectively along a state trajectory controlled by the learned control and a state trajectory controlled optimally (both using the same realization of the Brownian motion).
 }} \label{fig:algo1ex3exactparam} 
\end{figure}

In Figure~\ref{fig:algo1ex3exactparam}, we see that the parameters and, hence, the functions $K, R$ and $\phi$ are matched almost perfectly. We also display one realization of the control and of the cost. These are based on evaluating the control and the cumulative cost along one trajectory of the state. We first simulate $10^4$ realizations of a Brownian motion. Based on this, we generate trajectories for one $10^4$ population of agents using the learnt control and one population of $10^4$ agents using the optimal control. For the population that uses the learnt control, the control is given by the mean of the actor, namely, $\phi^\theta(t) (x -\bar\mu) + \phi_3^\theta(t)$. In the dynamics, the cost and the control, the mean field term is replaced by the empirical mean of the the corresponding population at the current time. We can see that the trajectories of control (resp. cost) are very similar. 

\vspace{2mm}

\begin{figure}[H]
  \centering
 \includegraphics[width=6cm,height=3.7cm]{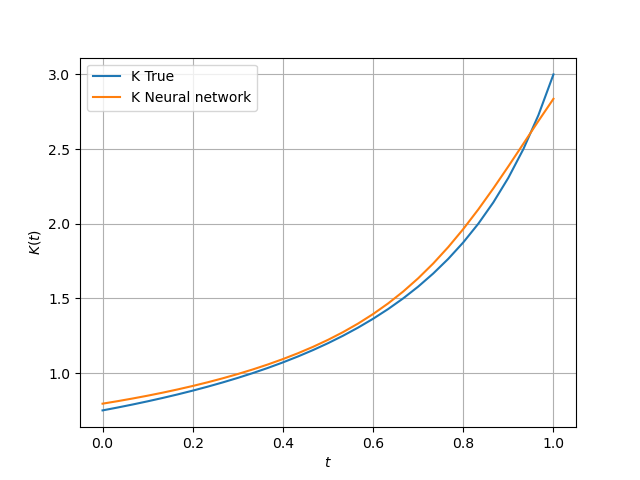}
 \includegraphics[width=6cm,height=3.7cm]{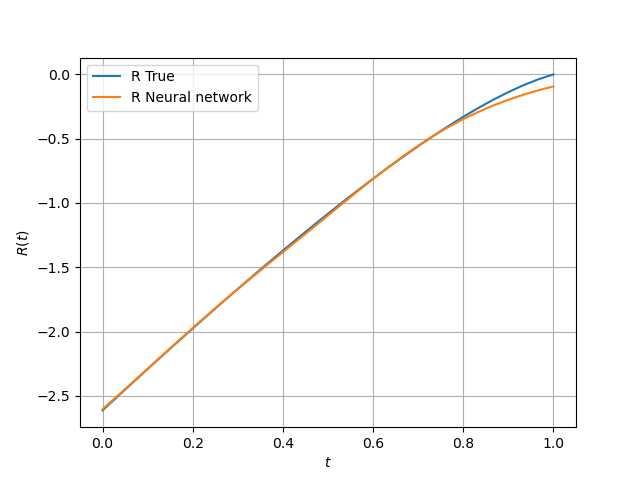}
 \caption{\footnotesize{{\bf Learnt critic cost  function with neural networks for the online Algorithm \ref{onalgo}}. {\it Left panel}: Neural network function $K^\eta$ vs optimal one. {\it Right panel}: Neural network function $R^\eta$ vs optimal one.} }  \label{fig:algo2ex3NNcritic}
\end{figure}

\begin{figure}[H]
  \centering
 \includegraphics[width=6cm,height=4cm]{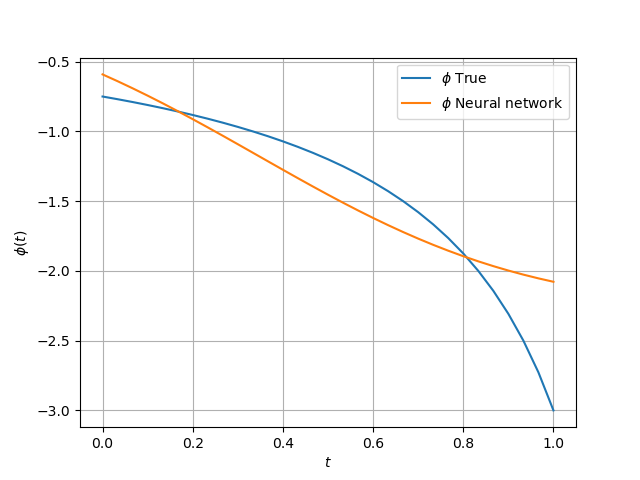}
  \centering
 \includegraphics[width=6cm,height=4cm]{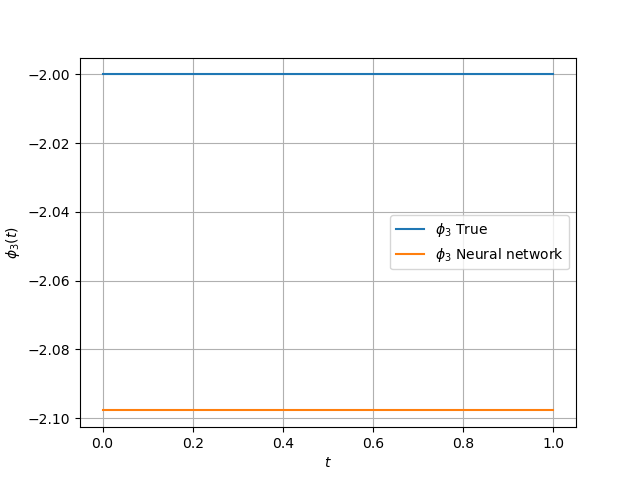}
  \centering
 \includegraphics[width=6cm,height=4cm]{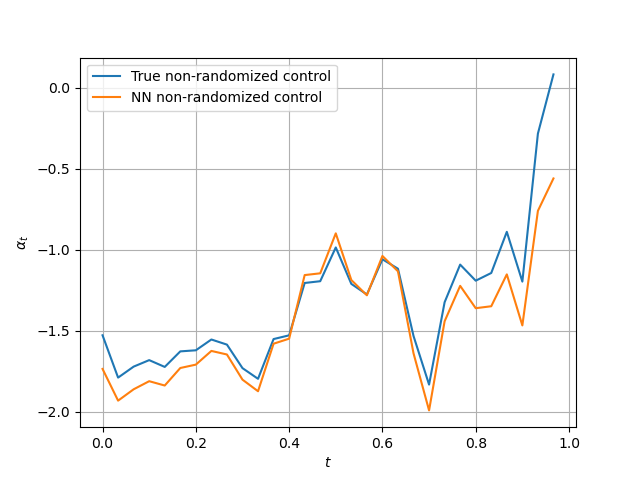}
  \centering
 \includegraphics[width=6cm,height=4cm]{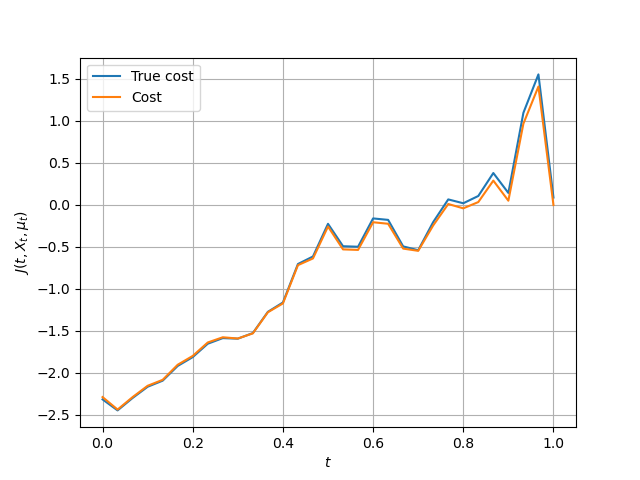}
 \caption{\footnotesize{{\bf Learnt actor policy function with neural networks (NN)  for the online Algorithm \ref{onalgo}}. {\it Left up panel}: NN $\phi^\theta$ vs optimal one.  {\it Right  up panel}: NN  $\phi_3^\theta$ vs optimal one.
 {\it  Bottom panel}: ({\it left}) One realization of the control vs the optimal non-randomised one with $\lambda=0$, and ({\it right}) one realization of the cost  $t$ $\mapsto$ $\mrJ^\eta(t,X_t,\P_{X_t})$ , 
 respectively along a state trajectory controlled by the learnt policy and a state trajectory controlled optimally (both using the same realization of the Brownian motion).  
 } 
  }  \label{fig:algo2ex3NNactor}
\end{figure}

Next, we present in Figure \ref{fig:algo2ex3NNcritic} and Figure \ref{fig:algo2ex3NNactor} the numerical results of our online Algorithm \ref{onalgo} when using neural networks. In this case, the derivatives  w.r.t. to  $\eta$ of $K^\eta$, $R^\eta$, hence of $\mrJ^\eta$, as well as the derivative w.r.t. $\theta$ of $\log p_\theta$ 
 are computed by automatic differentiation. 
We use neural networks with 3 hidden layers,  10 neurons per layer and tanh activation functions. We take $n=30$, $N=15000$ iterations, batch size 300 (10000 for the law estimation in the simulator), constant learning rates $10^{-3}$, except $\omega_S = 1$. 
Again, we change $\lambda$ along episodes: $\lambda$ $=$ $0.1$ for the second 3334 ones, then $0.01$ for the next 3333 ones, then $0.001$ until the end.

Finally, we test in Table \ref{table:costex2} the learnt policies from the exact and NN parametrisation by computing the associated initial expected social costs. 
We simulate $10$ populations, each consisting of $10^4$ agents. All the agents use the control function with the parameters learnt by the algorithm. For the dynamics, the cost and the control, the mean field term is replaced by the empirical mean of the corresponding population at the present time step. For each population, we compute the social cost. We then average over the $10$ populations in order to get a Monte Carlo estimate of the social cost. We report in the table the value of this average social cost, the standard deviation over the $10$ populations, and the relative error between the average social cost and the optimal cost computed by the formula $K^{\eta^*}_0 {\rm Var}(X_0) +R^{\eta^*}_0$ with the optimal parameter $\eta^*$.

\begin{table}[H]
\centering
\begin{tabular}{|c|c|c|}
\hline
& Social cost (Std dev.) & Rel. error \\
\hline
Learnt with exact parametrisation &  $-1.861$ ($0.025$) &  $0.11\%$   \\
\hline
Learnt with NN & $-1.787$ ($0.035$) & $4.08\%$ \\
\hline
Exact value &  $-1.863$ & \\
\hline
\end{tabular}
\caption{Initial social  costs when following learnt policies vs  optimal one.
} \label{table:costex2} 
\end{table}

\appendix

\section{Proofs of some representation results}  \label{sec:appenA}

\subsection{Proof of Proposition \ref{proVpi}} \label{sec:proVpi}

\noindent \emph{Step 1:} For a fixed policy $\pi$, we introduce the non-linear McKean-Vlasov SDE with dynamics
\begin{equation}
\label{SDE:McKean:fixed:policy}
\tilde X^{t, \xi}_s = \xi + \int_t^s  b_\pi(r,\tilde X^{t, \xi}_r,\P_{\tilde X^{t, \xi}_r}) \, \d r + \int_t^s \sigma_\pi(r,\tilde X^{t, \xi}_r,\P_{\tilde X^{t, \xi}_r})\, \d W_r, 
\end{equation}

\noindent recalling that $\sigma_\pi$ $=$ $\Sigma_\pi^{1/2}$, as well as its associated decoupled SDE with dynamics
\begin{equation}
\label{SDE:decoupled:McKean:fixed:policy}
\tilde X^{t, x, \mu}_s = x + \int_t^s  b_\pi(r,\tilde X^{t, x, \mu}_r,\P_{\tilde X^{t, \xi}_r}) \, \d r + \int_t^s \sigma_\pi(r,\tilde X^{t, x, \mu}_r,\P_{\tilde X^{t, \xi}_r})\, \d W_r. 
\end{equation}
Under Assumption \ref{hypcoeff}(i), the coefficients $b_\pi$ and $\sigma_\pi$ are Lipschitz-continuous and with at most linear growth with respect to the variable $x$ and $\mu$ locally uniformly in time. Hence, the SDEs \eqref{SDE:McKean:fixed:policy}-\eqref{SDE:decoupled:McKean:fixed:policy} admit a unique strong solution.

Denoting by $\P$ the probability measure on $\mathcal{C}([0, \infty), \R^d)$ (the space of continuous functions defined on $[0,\infty)$ taking values in $\R^d$) induced by the unique solution to the SDE \eqref{SDE:McKean:fixed:policy} and by $\P(t)$ its marginal at time $t$, its infinitesimal generator is given by
\begin{align*}
\tilde{\mathcal{L}}^{\pi}_t\varphi(x) & = \sum_{i=1}^d \int_{A} b_{i}( x, \P(t), a) \, \pi(\d a| t, x, \P(t)) \partial_{x_i} \varphi(x) \\
& \quad + \;  \frac12 \sum_{i,j=1}^d \int_{A} (\sigma\sigma\trans)_{i, j}(x, \P(t), a)  \, \pi(\d a| t, x, \P(t)) \partial^2_{x_i, x_j} \varphi(x).
\end{align*}
Now, coming back to the dynamics of the McKean-Vlasov SDE \eqref{McKean:Vlasov:SDE:and:decoupled:SDE}, we importantly point out that since at each time $s$, the action $\alpha_s$ is sampled from the probability distribution $\pi(.|s, X^{t, \xi}_s, \P_{X_s^{t, \xi}})$ independently of $W$, the infinitesimal generator at time $t$ of \eqref{McKean:Vlasov:SDE:and:decoupled:SDE} is exactly given by $\tilde{\mathcal{L}}^{\pi}_t$. Hence, it follows from the uniqueness of the martingale problem associated to $\tilde{\mathcal{L}}^{\pi}$ that $X^{t, \xi}$ and $\tilde X^{t, \xi}$ have the same law \footnote{This was formally shown by law of large numbers in \cite{wanzarzhou20} in the standard diffusion case.}.

We thus conclude that $V^\pi$ can be written as 
\begin{equation}
\label{new:expression:vpi}
\begin{aligned} 
V^\pi(t, x,\mu) &= \;  \E_{} \Big[ \int_t^T e^{-\beta (s-t)}  (f_\pi-\lambda E_\pi)(s,\tilde X_s^{t,x,\mu},\P_{\tilde X_s^{t,\xi}})  \, \d s 
+ \;  e^{-\beta (T-t)} g(\tilde X_T^{t, x, \mu},\P_{\tilde X_T^{t,\xi}}) \Big].
\end{aligned}
\end{equation}

\vspace{1mm}

\noindent \emph{Step 2:} We know, see e.g. \cite{crisan:murray} or \cite{chacridel15}, that Assumption \ref{hypcoeff}(i) guarantees the existence of  a modification of $\tilde X^{t, x, \mu}$ such that:
 \begin{itemize}
 \item The map $x\mapsto \tilde X_s^{t, x, \xi}$ is $\mathbb{P}$-a.s. twice continuously differentiable, 
 \item  for any $x$ $\in$ $\mathbb{R}^d$,  $0\leq t\leq s$, and any $p\geq1$, the map  $\mathcal{P}_2(\mathbb{R}^d) \ni \mu \mapsto \tilde X_s^{t, x, \mu} \in L^{p}(\mathbb{P})$ is differentiable and the map $\mathbb{R}^d \ni v \mapsto \partial_\mu \tilde X_s^{t, x, \mu}(v) \in L^{p}(\mathbb{P})$ is differentiable,
 \item for any $p\geq1$, the derivatives $(t, x, \mu, v) \mapsto \partial_x \tilde X_s^{t, x, \mu}$, $\partial_x^2 \tilde X_s^{t, x, \mu}$, $\partial_\mu \tilde X_s^{t, x, \mu}(v)$, $\partial_v [\partial_\mu \tilde X_s^{t, x, \mu}](v) \in L^{p}(\mathbb{P})$ are continuous.
 \end{itemize}
Moreover, the following estimates hold for $n=1,2$ and any $p\geq1$
 $$
\sup_{0\leq t \leq s \leq T, (x,v) \in (\mathbb{R}^d)^2, \mu \in \Pc_2(\mathbb{R}^d)} \left\{ \|\partial_x^n \tilde X_s^{t, x, \mu}\|_{L^{p}(\mathbb{P})} + \| \partial_\mu \tilde X_s^{t, x, \mu}(v) \|_{L^{p}(\mathbb{P})} +  \| \partial_v \partial_\mu \tilde X_s^{t, x, \mu}(v) \|_{L^{p}(\mathbb{P})} \right\} < \infty. 
 $$

We thus deduce that the functions $x\mapsto f_\pi(s,\tilde X_s^{t,x,\mu},\P_{\tilde X_s^{t,\xi}}), \,  E_\pi(s,\tilde X_s^{t, x,\mu},\P_{\tilde X_s^{t,\xi}}), \, g(\tilde X_T^{t, x, \mu},\P_{\tilde X_T^{t,\xi}})$ are $\mathbb{P}$-a.s. twice continuously differentiable with derivatives that belong to $L^{p}(\mathbb{P})$, for any $p\geq1$, uniformly in $x$, $\mu$ and $t\in [0,s]$. The dominated convergence theorem eventually guarantees that $x\mapsto V^\pi (t, x, \mu)$ is twice continuously differentiable with
\begin{align}\label{first:space:derivative:value:function}
\partial_{x_i} V^{\pi}(t, x, \mu) &= \;  \E_{} \Big[ \int_t^T e^{-\beta (s-t)} \sum_{k=1}^d  \partial_{x_k} (f_\pi-\lambda E_\pi)(s,\tilde X_s^{t, x,\mu},\P_{\tilde X_s^{t,\xi}})   \partial_{x_i} (\tilde X_s^{t, x, \mu})^k  \d s \nonumber \\
&  \hspace{3cm} + \;  e^{-\beta (T-t)} \sum_{k=1}^d \partial_{x_k} g(\tilde X_T^{t, x, \mu},\P_{\tilde X_T^{t,\xi}})  \partial_{x_i} (\tilde X_T^{t, x, \mu})^k \Big],
\end{align}
\noindent and
\begin{align*}
\partial^2_{x_i, x_j} V^{\pi}(t, x, \mu) &= \;  \E_{} \Big[ \int_t^T e^{-\beta (s-t)} \sum_{k, \ell=1}^d \partial^2_{x_k, x_\ell} (f_\pi-\lambda E_\pi)(s,\tilde X_s^{t, x,\mu},\P_{\tilde X_s^{t,\xi}})   \partial_{x_i} (\tilde X_s^{t, x, \mu})^k  \partial_{x_j} (\tilde X_s^{t, x, \mu})^\ell  \d s \nonumber \\
& \quad+ \;  e^{-\beta (T-t)} \sum_{k, \ell=1}^d \partial^2_{x_k, x_\ell} g(\tilde X_T^{t, x, \mu},\P_{\tilde X_T^{t,\xi}})  \partial_{x_i} (\tilde X_T^{t, x, \mu})^k  \partial_{x_j} (\tilde X_T^{t, x, \mu})^\ell \\
& + \int_t^T e^{-\beta (s-t)} \sum_{k=1}^d  \partial_{x_k} (f_\pi-\lambda E_\pi)(s,\tilde X_s^{t, x,\mu},\P_{\tilde X_s^{t,\xi}})   \partial^2_{x_i, x_j} (\tilde X_s^{t, x, \mu})^k   \d s \nonumber \\
& \quad + \;  e^{-\beta (T-t)} \sum_{k=1}^d \partial_{x_k} g(\tilde X_T^{t, x, \mu},\P_{\tilde X_T^{t,\xi}})  \partial^2_{x_i, x_j} (\tilde X_T^{t, x, \mu})^k  \Big].
\end{align*}
It follows from the above expression and again the dominated convergence theorem that $(t, x, \mu) \mapsto \partial_{x_i} V^{\pi}(t, x, \mu), \partial^2_{x_i, x_j} V^{\pi}(t, x, \mu)$ are continuous.

Similarly, note that under the current assumption, the functions $\mu \mapsto h(t,\tilde X_s^{t, x, \mu},\P_{\tilde X_s^{t,\xi}})$, $g(\tilde X_T^{t, x, \mu},\P_{\tilde X_T^{t,\xi}})$, where $h \in \left\{   f_\pi, E_\pi \right\}$, are L-differentiable with derivatives satisfying
\begin{equation}\label{expression:L:derivatives}
\begin{aligned}
\partial^i_\mu [h(s,\tilde X_s^{t, x,\mu},\P_{\tilde X_s^{t,\xi}})](v) & = \sum_{k=1}^d \partial_{x_k} h(s,\tilde X_s^{t, x,\mu},\P_{\tilde X_s^{t,\xi}}) \partial^{i}_\mu [(\tilde X_s^{t, x,\mu})^k](v) \\
& \quad +\widehat{\mathbb{E}}\Big[ \sum_{k=1}^d \partial^k_\mu h(s, \tilde X_s^{t, x,\mu}, \P_{\tilde X_s^{t,\xi}})(\widehat{X}_s^{t, v, \mu}) \partial_{x_i} (\widehat{X}_s^{t, v, \mu})^k\Big]  \\
& \quad + \int_{\mathbb{R}^d}\widehat{\mathbb{E}}\Big[\sum_{k=1}^d \partial^k_\mu h(s,\tilde X_s^{t, x,\mu}, \P_{\tilde X_s^{t,\xi}})(\widehat{X}_s^{t, x', \mu}) \partial^i_\mu [(\widehat{X}_s^{t, x', \mu})^k](v) \Big] \,  \mu(\d x'), \\
\partial^i_\mu [g(\tilde X_T^{t, x, \mu},\P_{\tilde X_T^{t,\xi}})](v) & =  \sum_{k=1}^d \partial_{x_k} g(\tilde X_T^{t, x,\mu},\P_{\tilde X_T^{t,\xi}}) \partial^{i}_\mu [(\tilde X_T^{t, x,\mu})^k](v)\\
& \quad + \widehat{\mathbb{E}}[ \sum_{k=1}^d \partial^k_\mu g(\tilde X_T^{t, x,\mu}, \P_{\tilde X_T^{t,\xi}})(\widehat{X}_T^{t, v, \mu}) \partial_{x_i} (\widehat{X}_T^{t, v, \mu})^k ]  \\
& \quad + \int_{\mathbb{R}^d}\widehat{\mathbb{E}}[\sum_{k=1}^d \partial^k_\mu g(\tilde X_T^{t, x,\mu}, \P_{\tilde X_T^{t,\xi}})(\widehat{X}_T^{t, x', \mu}) \partial^i_\mu [(\widehat{X}_T^{t, x', \mu})^k] (v) ] \,  \mu(\d x'),
\end{aligned} 
\end{equation}

\noindent where $(\widehat{X}_s^{t, x, \mu})_{s\in [t, T]}$ stands for a copy of $(\tilde{X}^{t, x, \mu}_s)_{s \in [t, T]}$ defined on a copy $(\widehat{\Omega}, \widehat{\mathcal{F}}, \widehat{\mathbb{P}})$ of the original probability space $(\Omega, \mathcal{F}, \mathbb{P})$. Under Assumption \ref{hypcoeff}, it follows from the above identities that $(t, x, \mu, v) \mapsto \partial_\mu [h(s,\tilde X_s^{t, x,\mu},\P_{\tilde X_s^{t,\xi}})](v) , \, \partial_\mu [g(\tilde X_T^{t, x, \mu},\P_{\tilde X_T^{t,\xi}})](v) \in L^{p}(\mathbb{P})$, $p\geq1$, are continuous and satisfy
\begin{align*}
|\partial_\mu [h(s,\tilde X_s^{t, x, \mu}, \P_{\tilde X_s^{t,\xi}})](v) |&  \leq K (1+ |\tilde X_s^{t, x, \mu}| + |v| + M_2( \P_{\tilde X_s^{t,\xi}})^q) (1+ |\partial_\mu \tilde{X}^{t, x, \mu}_s(v)| ) \\
& \leq K (1+ |\tilde X_s^{t, x,\mu}| + |v| + M_2(\mu)^q) (1+ |\partial_\mu \tilde{X}^{t, x, \mu}_s(v)| ),
\end{align*}
\noindent and
\begin{align*}
| \partial_\mu [g(\tilde X_T^{t, x, \mu},\P_{\tilde X_T^{t,\xi}})](v) | & \leq K (1 + |\tilde X_T^{t, x,\mu}| + |v| + M_2( \P_{\tilde X_T^{t,\xi}})^q) (1+ |\partial_\mu \tilde{X}^{t, x, \mu}_T(v)| )\\
& \leq K (1+ |\tilde X_T^{t, x,\mu}| + |v| + M_2(\mu)^q) (1+ |\partial_\mu \tilde{X}^{t, x, \mu}_T(v)| ),
\end{align*}

\noindent where we used the fact that $M_2( \P_{\tilde X_s^{t,\xi}}) \leq K(1+M_2(\mu))$, for any $s\in [t, T]$, for the last inequality. Similarly, it follows from \eqref{expression:L:derivatives} and the dominated convergence theorem that $v\mapsto \partial_\mu [h(t,\tilde X_s^{t, x, \mu},\P_{\tilde X_s^{t,\xi}})](v)$, $ \partial_\mu [g(\tilde X_T^{t, x, \mu},\P_{\tilde X_T^{t,\xi}})](v)$ are continuously differentiable with derivatives being continuous with respect to their entries and satisfying
\begin{align*}
| \partial_v \partial_\mu [h(s,\tilde X_s^{t, x,\mu}, \P_{\tilde X_s^{t,\xi}})](v) |&  \leq  K (1+ |\tilde X_s^{t, x, \mu}| + |v| + M_2(\mu)^q),\\
| \partial_v \partial_\mu [g(\tilde X_T^{t, x, \mu},\P_{\tilde X_T^{t,\xi}})](v) | & \leq K (1+ |\tilde X_T^{t, x, \mu}| + |v| + M_2(\mu)^q).
\end{align*}

Coming back to \eqref{new:expression:vpi} and using the above estimates together with the dominated convergence theorem allows to conclude that $\mu \mapsto V^{\pi}(t, x, \mu)$ is L-differentiable and that $v\mapsto \partial_\mu V^{\pi}(t, x, \mu)(v)$ is differentiable. Moreover, both derivatives $\partial_\mu V^{\pi}(t, x, \mu)(v)$, $\partial_v \partial_\mu V^{\pi}(t, x, \mu)(v)$ are continuous with respect to their entries and satisfy
\begin{equation}\label{estimate:first:crossed:L:derivative:vpi}
\sup_{t \in [0,T]}\left\{ | \partial_\mu V^{\pi}(t, x, \mu)(v)| + |\partial_v \partial_\mu V^{\pi}(t, x, \mu)(v)| \right\} \leq K (1+ |x| +|v| + M_2(\mu)^q).
\end{equation}
We thus conclude that $V^{\pi} \in \mathcal{C}^{0, 2, 2}([0,T] \times \mathbb{R}^d \times \Pc_2(\mathbb{R}^d))$.\\

\vspace{1mm}

\noindent \emph{Step 3:}  Let us now prove that $(t, x, \mu) \mapsto V^{\pi}(t, x, \mu) \in  \mathcal{C}^{1, 2, 2}([0,T] \times \mathbb{R}^d \times \Pc_2(\mathbb{R}^d))$. From the Markov property satisfied by the SDE \eqref{SDE:McKean:fixed:policy}, stemming from its strong well-posedness, for any $0\leq h \leq t$, the following relation is satisfied
\begin{equation}
\begin{aligned}
V^{\pi}(t-h, x, \mu) & =  e^{-\beta h} \mathbb{E}\Big[ \int_{t-h}^{t} e^{-\beta (s-t)}   (f_\pi-\lambda E_\pi)(s,\tilde X_s^{t-h, x, \mu},\P_{\tilde X_s^{t-h, \xi}})   \d s\Big] \\
\quad & + e^{-\beta h} \mathbb{E}\Big[ V^{\pi}(t, \tilde X_t^{t-h, x, \mu}, \mathbb{P}_{\tilde X_t^{t-h, \xi}})\Big].
\end{aligned}
\end{equation}
Now, combining the fact that $V^{\pi}(t, .) \in\mathcal{C}^{2, 2}(\mathbb{R}^d \times \Pc_2(\mathbb{R}^d))$ with \eqref{estimate:first:crossed:L:derivative:vpi} guarantees that one may apply It\^o's rule, see e.g. Proposition 5.102 \cite{cardel19}. We thus obtain
\begin{equation}\label{ratio:time:value:function}
\begin{aligned}
h^{-1} (V^{\pi}(t-h, x, \mu)  & -V^{\pi}(t, x, \mu)) \\
& =  e^{-\beta h} h^{-1} \int_{t-h}^{t} e^{-\beta (s-t)}   \mathbb{E}\Big[ (f_\pi-\lambda E_\pi)(s,\tilde X_s^{t-h, x,\mu},\P_{\tilde X_s^{t-h, \xi}})  \Big] \d s \\
 & \quad + e^{-\beta h} h^{-1} \mathbb{E}\Big[ V^{\pi}(t, \tilde X_t^{t-h, x, \mu}, \mathbb{P}_{\tilde X_t^{t-h, \xi}}) - V^{\pi}(t, x, \mu)\Big]\\
& \quad + h^{-1} (e^{-\beta h} - 1) V^{\pi}(t, x, \mu)\\
 & = e^{-\beta h} h^{-1} \int_{t-h}^{t} e^{-\beta (s-t)}   \mathbb{E}\Big[ (f_\pi-\lambda E_\pi)(s,\tilde X_s^{t-h, x,\mu},\P_{\tilde X_s^{t-h, \xi}})    \Big] \d s \\
 & \quad + e^{-\beta h} h^{-1}  \int_{t-h}^{t} \mathbb{E}\Big[\tilde \Lc[V^{\pi}](t, \tilde X_s^{t-h, x, \mu}, \mathbb{P}_{\tilde X_s^{t-h, \xi}})\Big] \d s\\
& \quad + h^{-1} (e^{-\beta h} - 1) V^{\pi}(t, x, \mu),
\end{aligned}
\end{equation}
where
 \beqs
 \tilde \Lc_\pi[\varphi](t, x,\mu) 
 &=&   b_\pi(t, x, \mu) \cdot D_x\varphi(t, x, \mu) +  \frac{1}{2}  \Sigma_\pi(t, x, \mu) : D_x^2 \varphi(t, x, \mu)  \\
& & \; +  \; \E_{\xi\sim\mu} \Big[   b_\pi(t,\xi,\mu) \cdot \partial_\mu \varphi(t, x, \mu)(\xi) + \frac{1}{2} \Sigma_\pi(t, \xi, \mu) : \partial_\upsilon\partial_\mu \varphi(t, x, \mu)(\xi)    \Big].
 \enqs
Letting $h\downarrow 0$ in \eqref{ratio:time:value:function}, from the continuity and quadratic growth of $f_\pi$, $E_\pi$ as well as the continuity of $\tilde \Lc[V^{\pi}](t,.)$, we deduce that $t\mapsto V^{\pi}(t, x, \mu)$ is left-differentiable on $(0,T)$. Still from the continuity of  $f_\pi$, $E_\pi$ and $\tilde \Lc[V^{\pi}]$, we eventually conclude that it is differentiable on $[0,T)$ with a derivative satisfying
$$
\partial_t V^{\pi}(t, x, \mu) - \beta V^{\pi}(t, x, \mu) +  \tilde \Lc_\pi[V^{\pi}](t, x,\mu) +  (f_\pi-\lambda E_\pi)(t, x, \mu) = 0. 
$$
The proof is now complete.

\subsection{Differentiability of the parametric critic function} \label{sec:diffV}

Under the standard assumption that the coefficients $b_{\pi_\theta}(t, .), \, \sigma_{\pi_\theta}(t, .)$ are Lipschitz-continuous on $\mathbb{R}^d \times \Pc_2(\mathbb{R}^d)$ uniformly in $t\in [0,T]$ and $\theta \in \Theta$, the system of SDEs \eqref{McKean:Vlasov:SDE:and:decoupled:SDE} admits a unique strong solution when $\alpha$ $\sim$ $\pi_\theta$. We will denote by $(X_s^{t, \xi}(\theta), X_s^{t, x, \xi}(\theta))$ the solution taken at time $s$. We will also use the more compact notation
 \begin{equation}
\label{McKean:Vlasov:SDE:and:decoupled:SDE:compact:notation}
\begin{aligned}
X_s^{t,\xi}(\theta) &= \xi + \int_t^s \sum_{j=0}^{p} g^{j}_\theta(r, X_r^{t, \xi}(\theta), \P_{X_r^{t, \xi}(\theta)}) \, \d W^{j}_r    \\
X_s^{t, x,\mu}(\theta) &= x + \int_t^s \sum_{j=0}^p g^{j}_\theta(r, X_r^{t, x, \mu}(\theta), \P_{X_r^{t, \xi}(\theta)}) \, \d W^{j}_r, \quad t \leq s \leq T,  
\end{aligned}
\end{equation}

\noindent with $g^{0}_\theta(t, x, \mu) = b_{\pi_\theta}(t, x, \mu)$, $g^{j}_\theta(t, x, \mu) = \sigma^{., j}_{\pi_\theta}(t, x, \mu)$, $\d W_r = ( \d W_r^{0}, \cdots, \d W_r^{p})$ with $\d W^{0}_r = \d r$.

\begin{Lemma}\label{lem:reg:flow}
Under Assumption \ref{hyppitheta}, the derivatives $(t, \theta, x, \mu, v) \mapsto \partial_{\theta} \partial_{x} \tilde X_s^{t, x, \mu}(\theta)$, $\partial_{x} \partial_{\theta} \tilde X_s^{t, x, \mu}(\theta)$, $\partial_{\theta} \partial^2_{x} \tilde X_s^{t, x, \mu}(\theta)$, $\partial^2_{x} \partial_{\theta} \tilde X_s^{t, x, \mu}(\theta)$, $\partial_{\theta} [\partial_\mu \tilde X_s^{t, x, \mu}(\theta)](v)$, $ \partial_\mu \partial_{\theta} \tilde X_s^{t, x, \mu}(\theta)(v)$, $\partial_{\theta}\partial_{v}[\partial_\mu \tilde X_s^{t, x, \mu}(\theta)](v)$, 
$\partial_{v} [\partial_\mu \partial_{\theta} \tilde X_s^{t, x, \mu}(\theta)](v)$ $\in$ $L^{p}(\mathbb{P})$ exist and are locally Lipschitz continuous for all $p\geq1$. 
\end{Lemma}
{\bf Proof.} The proof of the existence and continuity of the derivatives of the flow $X_s^{t, x, \xi}(\theta)$ with respect to the parameters $x$, $\mu$, $v$ and $\theta$ is rather standard but quite mechanical and actually follows similar lines of reasonings as those employed for the proof of Theorem 3.2 in \cite{crisan:murray}. We thus omit it.  
\ep

With the same notations as Lemma \ref{lem:reg:flow}, under Assumption \ref{hyppitheta}, taking $h_\theta = f_{\pi_\theta}$, $E_{\pi_\theta}$ or $g(x, \mu)$, we deduce from the above result that the derivatives $(t, \theta, x, \mu, \boldsymbol{v}) \mapsto \partial_{\theta} \partial_{x} [h_\theta(s, \tilde X_s^{t, x, \mu}(\theta), \P_{\tilde X_s^{t, \xi}(\theta)})]$, $\partial_{x} \partial_{\theta} [h_\theta(s, \tilde X_s^{t, x, \mu}(\theta), \P_{\tilde X_s^{t, \xi}(\theta)})]$, $\partial_{\theta} \partial_\mu  [h_\theta(s, \tilde X_s^{t, x, \mu}(\theta), \P_{\tilde X_s^{t, \xi}(\theta)})](v)$, $ \partial_\mu \partial_{\theta}  [h_\theta(s, \tilde X_s^{t, x, \mu}(\theta), \P_{\tilde X_s^{t, \xi}(\theta)})](v)$, $\partial_{\theta}\partial_{v}\partial_\mu [h_\theta(s, \tilde X_s^{t, x, \mu}(\theta), \P_{\tilde X_s^{t, \xi}(\theta)})](v)$,  $\partial_{v} \partial_\mu \partial_{\theta} [h_\theta(s, \tilde X_s^{t, x, \mu}(\theta), \P_{\tilde X_s^{t, \xi}(\theta)})](v)$ $\in L^{p}(\mathbb{P})$, for any $p \geq1$ and any $0\leq t\leq s \leq T$, exist and are continuous. For instance, standard computations give {\small
 \begin{align*}
\partial_{\theta_l} \partial_{x_i} & [h_\theta(s, \tilde X_s^{t, x, \mu}(\theta), \P_{\tilde X_s^{t, \xi}(\theta)})] \\
& = \sum_{j=1}^d \partial_{\theta_l} \partial_{x_j} h_\theta(s, \tilde X_s^{t, x, \mu}(\theta), \P_{\tilde X_s^{t, \xi}(\theta)}) \partial_{x_i} (\tilde X_s^{t, x, \mu}(\theta))^{j} \\
& \quad + \sum_{j,k=1}^d \partial^2_{x_j, x_k} h_\theta(s, \tilde X_s^{t, x, \mu}(\theta), \P_{\tilde X_s^{t, \xi}(\theta)}) \partial_{x_i} (\tilde X_s^{t, x, \xi}(\theta))^j  \partial_{\theta_l} (\tilde X_s^{t, x, \mu}(\theta))^k \\
& \quad +  \sum_{j, k=1}^d \widehat{\mathbb{E}}\Big[ [\partial_\mu \partial_{x_j} h_\theta(s, \tilde X_s^{t, x, \mu}(\theta), \P_{\tilde X_s^{t, \xi}(\theta)})]_{k}(\widehat X_s^{t, \mu}(\theta)) \partial_{\theta_l} (\widehat X_s^{t, \xi}(\theta))^k \Big] \partial_{x_i} (\tilde X_s^{t, x, \mu}(\theta))^{j}\\
& \quad + \sum_{j=1}^d \partial_{x_j} h_\theta(s, \tilde X_s^{t, x, \mu}(\theta), \P_{\tilde X_s^{t, \xi}(\theta)}) \partial_{\theta_l}\partial_{x_i} (\tilde X_s^{t, x, \mu}(\theta))^{j}
 \end{align*}
 \noindent and 
  \begin{align*}
\partial_{\theta_l} \partial^{i}_{\mu} & [h_\theta(s, \tilde X_s^{t, x, \mu}(\theta), \P_{\tilde X_s^{t, \xi}(\theta)})](v) \\
& = \sum_{j = 1}^d \partial_{\theta_l} \partial_{x_j} h_\theta(s, \tilde X_s^{t, x, \mu}(\theta), \P_{\tilde X_s^{t, \xi}(\theta)}) \partial^{i}_{\mu} (\tilde X_s^{t, x, \mu}(\theta))^{j} \\
& \quad + \sum_{j,k =1}^d \partial^2_{x_j, x_k} h_\theta(s, \tilde X_s^{t, x, \mu}(\theta), \P_{\tilde X_s^{t, \xi}(\theta)}) \partial^{i}_{\mu} (\tilde X_s^{t, x, \mu}(\theta))^j  \partial_{\theta_l} (\tilde X_s^{t, x, \mu}(\theta))^k \\
& \quad + \sum_{j,k =1}^d \widehat{\mathbb{E}}\Big[ [\partial_\mu \partial_{x_j} h_\theta(s, \tilde X_s^{t, x, \mu}(\theta), \P_{\tilde X_s^{t, \xi}(\theta)})]_{k}(\widehat X_s^{t, \xi}(\theta)) \partial_{\theta_l} (\widehat X_s^{t, \xi}(\theta))^k \Big] \partial^{i}_{\mu} (\tilde X_s^{t, x, \mu}(\theta))^{j}\\
& \quad  + \sum_{j=1}^d \partial_{x_j} h_\theta(s, \tilde X_s^{t, x, \mu}(\theta), \P_{\tilde X_s^{t, \xi}(\theta)}) \partial_{\theta_l}\partial^{i}_{\mu} (\tilde X_s^{t, x, \mu}(\theta))^{j} \\
& \quad + \sum_{j=1}^d \widehat{\mathbb{E}}[ \partial_{\theta_{l}} \partial^j_\mu h_\theta(s, \tilde X_s^{t, x,\mu}(\theta), \P_{\tilde X_s^{t,\xi}(\theta)})(\widehat{X}_s^{t, v, \mu}(\theta)) \partial_{v_i} (\widehat{X}_s^{t, v, \mu}(\theta))^{j} ]  \\
& \quad + \sum_{j, k=1}^d \widehat{\mathbb{E}}[ \partial_{x_{k}} \partial^j_\mu h_\theta(s, \tilde X_s^{t, x,\mu}(\theta), \P_{\tilde X_s^{t,\xi}(\theta)})(\widehat{X}_s^{t, v, \mu}(\theta)) \partial_{v_i} (\widehat{X}_s^{t, v, \mu}(\theta))^{j} ] \partial_{\theta_{l}} (\tilde X_s^{t, x, \mu})^k \\
& \quad + \sum_{j, k=1}^d   \widehat{\mathbb{E}}\check{\mathbb{E}}[ \partial^{k}_\mu \partial^{j}_\mu h_\theta(s,\tilde X_s^{t, x,\mu}(\theta), \P_{\tilde X_s^{t,\xi}(\theta)})(\widehat{X}_s^{t, v, \mu}(\theta), \check{X}_s^{t, \xi}(\theta)) \partial_{v_i} (\widehat{X}_s^{t, v, \mu}(\theta))^{j} \partial_{\theta_l} (\check{X}_s^{t, \xi}(\theta))^{k} ] \,  \\
& \quad + \sum_{j, k =1}^d   \widehat{\mathbb{E}}[ \partial_{v_{k}} \partial^j_\mu h_\theta(s, \tilde X_s^{t, x,\mu}(\theta), \P_{\tilde X_s^{t,\xi}(\theta)})(\widehat{X}_s^{t, v, \mu}(\theta)) \partial_{v_i} (\widehat{X}_s^{t, v, \mu}(\theta))^{j} \partial_{\theta_l} (\widehat{X}_s^{t, v, \mu}(\theta))^{k} ]  \\
& \quad + \sum_{j=1}^d \widehat{\mathbb{E}}[  \partial^j_\mu h_\theta(s, \tilde X_s^{t, x,\mu}(\theta), \P_{\tilde X_s^{t,\xi}(\theta)})(\widehat{X}_s^{t, v, \mu}(\theta)) \partial_{\theta_l} \partial_{v_i} (\widehat{X}_s^{t, v, \mu}(\theta))^{j} ]\\
& \quad + \sum_{j=1}^d \int_{\mathbb{R}^d }\widehat{\mathbb{E}}[\partial_{\theta_l}  \partial^j_\mu h_\theta(s, \tilde X_s^{t, x, \mu}(\theta), \P_{\tilde X_s^{t, \xi}(\theta)})(\widehat{X}_s^{t, x', \mu}(\theta)) \partial^{i}_\mu (\widehat X_s^{t, x', \mu}(\theta))^j(v) ] \, \mu(\d x') \\
& \quad +  \sum_{j, k=1}^d \int_{\mathbb{R}^d }\widehat{\mathbb{E}}[\partial_{x_k} \partial^j_\mu h_\theta(s, \tilde X_s^{t, x, \mu}(\theta), \P_{\tilde X_s^{t, \xi}(\theta)})(\widehat{X}_s^{t, x', \mu}(\theta)) \partial_{\theta_l} (\tilde{X}_s^{t, x, \mu}(\theta))^{k} \partial^{i}_\mu (\widehat X_s^{t, x', \mu}(\theta))^j(v) ] \, \mu(\d x')\\
& \quad + \sum_{j, k=1}^d \int_{\mathbb{R}^d }\widehat{\mathbb{E}}\check{\mathbb{E}}[ \partial^{k}_{\mu} \partial^j_\mu h_\theta(s, \tilde X_s^{t, x, \mu}(\theta), \P_{\tilde X_s^{t, \xi}(\theta)})(\widehat{X}_s^{t, x', \mu}(\theta), \check{X}_s^{t, \xi}(\theta)) \partial^{i}_{\mu} (\widehat{X}_s^{t, x', \mu}(\theta))^j(v) \partial_{\theta_l} (\check{X}_s^{t, \xi}(\theta))^k] \, \mu(\d x')\\
& \quad + \sum_{j, k=1}^d \int_{\mathbb{R}^d }\widehat{\mathbb{E}}[ \partial_{v_k} \partial^j_\mu h_\theta(s, \tilde X_s^{t, x, \mu}(\theta), \P_{\tilde X_s^{t, \xi}(\theta)})(\widehat{X}_s^{t, x', \mu}(\theta)) \partial_{\theta_l} (\widehat{X}_s^{t, x', \mu}(\theta))^k \partial^{i}_\mu (\widehat X_s^{t, x', \mu}(\theta))^j(v) ] \, \mu(\d x')\\
& \quad + \sum_{j=1}^d \int_{\mathbb{R}^d }\widehat{\mathbb{E}}[  \partial^j_\mu h_\theta(s, \tilde X_s^{t, x, \mu}(\theta), \P_{\tilde X_s^{t, \xi}(\theta)})(\widehat{X}_s^{t, x', \mu}(\theta)) \partial_{\theta_l} \partial^{i}_\mu (\widehat X_s^{t, x', \mu}(\theta))^j(v)] \, \mu(\d x').
 \end{align*}
 }

In the above identity, $\check{X}^{t,  \xi}_s(\theta)$ stands for a random variable independent of $(\tilde X^{t, x, \mu}_s, \widehat{X}_s^{t, v, \mu}, \widehat{X}_s^{t, x', \mu})$ with the same law as $\tilde X_s^{t, \xi}$.
 
Then, starting from the expression of $\mrV_\theta$ in \eqref{new:expression:vpi} (with $\pi=\pi_\theta$), the dominated convergence theorem guarantees that the derivatives $(t, \theta, x, \mu, v) \mapsto \partial_{\theta}\partial_{x} \mrV_\theta(t, x, \mu)$, $\partial_{x} \partial_{\theta} \mrV_\theta(t, x, \mu)$, $  \partial_{\theta}\partial^2_{x} \mrV_\theta(t, x, \mu)$, $\partial^2_{x} \partial_{\theta} \mrV_\theta(t, x, \mu)$, $\partial_{\theta} \partial_\mu \mrV_\theta(t, x, \mu)(v)$, $ \partial_\mu \partial_{\theta}\mrV_\theta(t, x, \mu)(v)$, $\partial_{\theta}\partial_{v}\partial_\mu \mrV_\theta(t, x, \mu)(v)$, $\partial_{v} \partial_\mu \partial_{\theta}\mrV_\theta(t, x, \mu)(v)$ exist and are locally Lipschitz continuous. Hence, from Clairaut's theorem, we deduce that $\partial_{\theta}\partial_{x} \mrV_\theta(t, x, \mu) = \partial_{x} \partial_{\theta} \mrV_\theta(t, x, \mu)$, $\partial_{\theta}\partial^2_{x} \mrV_\theta(t, x, \mu) = \partial^2_{x} \partial_{\theta} \mrV_\theta(t, x, \mu)$, $\partial_{\theta} \partial_\mu \mrV_\theta(t, x, \mu)(v) = \partial_\mu \partial_{\theta}\mrV_\theta(t, x, \mu)(v)$ and $\partial_{\theta}\partial_{v}\partial_\mu \mrV_\theta(t, x, \mu)(v) = \partial_{v} \partial_\mu \partial_{\theta}\mrV_\theta(t, x, \mu)(v)$ for all $t, x, \mu, \theta, v$.

Moreover, from Assumption \ref{hyppitheta} and Lemma \ref{lem:reg:flow}, there exist $q$ and $C$ such that for any $t, x, \mu, v$ and any $\theta \in \mathcal{K}$, $\mathcal{K}$ being a compact subset of $\Theta$ 
\begin{equation}\label{bound:deriv:theta:Vtheta}
|\partial_\theta \mrV_\theta(t, x, \mu)| \leq C(1 + |x|^2 + M_2(\mu)^q),
\end{equation}
\begin{equation}\label{bound:deriv:theta:deriv:x:mu:Vtheta}
|\partial_\theta \partial_{x} \mrV_\theta(t, x, \mu)| + |\partial_{\theta} \partial_\mu \mrV_\theta(t, x, \mu)(v)|  \leq C ( 1+ |x| + |v|+ M_2(\mu)^q),
\end{equation}
\noindent and
\begin{equation}\label{bound:deriv:theta:deriv:v:deriv:mu:Vtheta}
 |\partial_\theta \partial^2_{x} \mrV_\theta(t, x, \mu)| + |\partial_{\theta}\partial_{v}\partial_\mu \mrV_\theta(t, x, \mu)(v)|  \leq C ( 1+ |v| + M_2(\mu)^q).
\end{equation}

Now, differentiating with respect to $\theta$ both sides of \eqref{kolmogorov:PDE}, we deduce that $\theta\mapsto \partial_t \mrV_\theta(t, x, \mu)$ is differentiable with a derivative $ \partial_\theta \partial_t \mrV_\theta(t, x, \mu)$ being continuous with respect to $t, x, \mu, \theta$. Also, taking $\pi=\pi_\theta$ and differentiating with respect to $\theta$ both sides of the identity of \eqref{ratio:time:value:function} (using Lemma \ref{lem:reg:flow} together with the estimates \eqref{bound:deriv:theta:Vtheta}, \eqref{bound:deriv:theta:deriv:x:mu:Vtheta}, \eqref{bound:deriv:theta:deriv:v:deriv:mu:Vtheta} and the dominated convergence theorem to differentiate the right-hand side therein) and then passing to the limit as $h\downarrow 0$, we get that $t\mapsto \partial_\theta \mrV_\theta(t, x, \mu)$ is differentiable with a derivative $\partial_t \partial_\theta \mrV_\theta(t, x, \mu)$ being continuous with respect to $t, x, \mu, \theta$.  We thus conclude that the two derivatives $ \partial_\theta \partial_t \mrV_\theta(t, x, \mu)$ and $\partial_t \partial_\theta  \mrV_\theta(t, x, \mu)$ coincide for all $t, x, \mu, \theta$.

\subsection{Proof of Theorem \ref{theogradient}} \label{sec:theogradient}

\noindent \emph{Step 1:} We start from the PDE characterisation of $\mrV_\theta$ in Proposition \ref{proVpi} that we write as 
\begin{align} \label{PDEVtheta} 
\int_A  \big\{ \Lc_\theta^a[\mrV_\theta](t,x,\mu) + f(x,\mu,a) +  \lambda \log p_\theta(t,x,\mu,a) \big\} \, \pi_\theta(\d a|t,x,\mu) &= \;  0,
\end{align} 
where 
\beqs
 \Lc_\theta^a[\varphi](t,x,\mu)  &=& - \beta \varphi(t,x,\mu)  + \partial_t{\varphi}(t,x,\mu)  +  b(x,\mu,a) \cdot D_x\varphi(t,x,\mu) +  \frac{1}{2}  \sigma\sigma\trans(x,\mu,a) : D_x^2 \varphi(t,x,\mu)  \\
 & & \; +  \; \E_{\xi\sim\mu} \Big[   b_\theta(t,\xi,\mu) \cdot \partial_\mu \varphi(t,x,\mu)(\xi) + \frac{1}{2} \Sigma_\theta(t,\xi,\mu) : \partial_\upsilon\partial_\mu \varphi(t,x,\mu)(\xi)    \Big],
 \enqs

\noindent recalling that $b_\theta(t,x,\mu)$ $=$ $\int_A b(x,\mu,a) \, \pi_\theta(\d a|t,x,\mu)$, $\Sigma_\theta(t,x,\mu) \; = \; \int_A (\sigma\sigma\trans)(x,\mu,a) \,\pi_\theta(\d a|t,x,\mu)$.  

For any fixed $t, x, \mu$, we now differentiate w.r.t. $\theta$ $\in$ $\Theta$ both sides of \eqref{PDEVtheta} to get a new system of linear PDEs satisfied by $\mrG_\theta$.
In particular, using the identity
$$
\nabla_\theta \Big[\Lc_\theta^a[\mrV_\theta](t,x,\mu)\Big] = \Lc_\theta^a[\mrG_\theta](t,x,\mu) +   \Hc_\theta[\mrV_\theta](t,x,\mu),
$$
\noindent together with \eqref{diff:density:condition} and the dominated convergence theorem, we get
\begin{equation}
\label{first:form:PDE:gradient:theta:Vtheta}
\begin{aligned}
\int_A \Big\{ \Lc_\theta^a[\mrG_\theta](t,x,\mu)  +   \Hc_\theta[\mrV_\theta](t,x,\mu) \\
 +  \;  \big[ \Lc_\theta^a[\mrV_\theta](t,x,\mu) + f(x,\mu,a) +  \lambda \log p_\theta(t,x,\mu,a) \big]  \nabla_\theta \log p_\theta(t,x,\mu,a)  \Big\} \pi_\theta(\d a|t,x,\mu) &=\;  0, 
\end{aligned}
\end{equation}

\noindent with terminal condition $\mrG_\theta(T,x,\mu)$ $=$ $0$.
Note that we have used the fact that 
\beqs
\int_A \nabla_\theta \log p_\theta(t,x,\mu,a)  \pi_\theta(\d a|t,x,\mu)  &=& \nabla_\theta \int_A \pi_\theta(\d a|t,x,\mu) \;  = \; 0,  
\enqs
and  the above PDE is a system of $D$ equations, where  $\Lc_\theta^a[\mrG_\theta]$ denotes the operator applied to each component of the $\R^D$-valued function $\mrG_\theta$. \\ 

\noindent \emph{Step 2:} Denote by 
\begin{equation}
\begin{aligned}
\tilde F_\theta(t,x,\mu,a) &=  \big\{  \Lc_\theta^a[\mrV_\theta](t,x, \mu) + f(x,\mu,a) + \lambda   \log p_\theta(t,x,\mu,a) \big\}  \nabla_\theta \log p_\theta(t,x,\mu,a) \\
&  \quad \quad + \;  \Hc_\theta[\mrV_\theta](t,x,\mu),
\end{aligned}
\end{equation}
and
\beqs
\tilde f_{\pi_\theta}(t,x,\mu) &=&  \int_{A} \tilde F_\theta(t,x,\mu,a) \, \pi_\theta(\d a|t,x,\mu),
\enqs

\noindent so that the linear PDE \eqref{first:form:PDE:gradient:theta:Vtheta} satisfied by $\mrG_\theta$ now writes 
\beqs
  \Lc_{\pi_\theta}[\mrG_\theta](t,x,\mu)  +  \tilde f_{\pi_\theta}(t,x,\mu) &= \;  0,
\enqs
 with terminal condition $\mrG_\theta(T,x,\mu)$ $=$ $0$. Observe that the above PDE is similar to \eqref{kolmogorov:PDE}. In order to obtain the announced probabilistic representation formula, we first apply the chain rule formula on the strip $[t, T] \times \R^d\times\Pc_2(\R^d)$, see e.g. Proposition 5.102 in \cite{cardel19}, to $(e^{-\beta s} \mrG_\theta(s, \widetilde{X}_s^{t,x, \mu},\P_{\widetilde{X}_s^{t, \xi}}))_{s \in [t, T]}$ using the estimates \eqref{bound:deriv:theta:Vtheta} and \eqref{bound:deriv:theta:deriv:v:deriv:mu:Vtheta}. We thus obtain
\begin{align*}
d(e^{-\beta s} \mrG_\theta(s, \widetilde{X}_s^{t,x, \mu},\P_{\widetilde{X}_s^{t, \xi}})) & = - e^{-\beta s} \tilde f_{\pi_\theta}(s, \widetilde{X}_s^{t,x,\mu},\P_{\widetilde{X}_s^{t,\xi}}) \, \d s \\
& + e^{-\beta s} \partial_x \mrG_\theta(s,  \widetilde{X}_s^{t, x, \mu},\P_{\widetilde{X}_s^{t, \xi}})\trans \sigma_{\pi_\theta}(\widetilde{X}_s^{t, x, \mu}, \P_{ \widetilde{X}_s^{t,\xi}})\,  \d W_s.
\end{align*}

Observe that \eqref{bound:deriv:theta:deriv:x:mu:Vtheta} together with the fact that for any $\theta \in \R^D$, $|\sigma_{\pi_\theta}(x, \mu)| \leq C(1+|x|+M_2(\mu))$, for some constant $C$, directly yields that the stochastic integral is a square integrable martingale. Hence, integrating from $t$ to $T$ both sides of the above and using the facts that $\mrG_\theta(T,x,\mu)$ $=$ $0$ and $\mathbb{P}_{\widetilde{X}_s^{t, \xi}} = \mathbb{P}_{X_s^{t, \xi}}$, $\mathbb{P}_{\widetilde{X}_s^{t, x, \mu}} = \mathbb{P}_{X_s^{t, x, \mu}}$ , we eventually deduce
\begin{align} \label{Ginter} 
\mrG_\theta(t,x,\mu) &= \;   \E_{\alpha\sim\pi_\theta}   \Big[ \int_t^T  e^{-\beta(s-t)}  \tilde F_{\theta}(s, X_s^{t,x,\mu},\P_{ X_s^{t,\xi}}, \alpha_s)  \,  \d s  \Big]. 
\end{align} 

\noindent \emph{Step 3:} On the other hand,  applying again the chain rule formula to $\mrV_\theta(s,X_s^{t,x,\mu},\P_{X_s^{t,\xi}})$, when  $\alpha$ $\sim$ $\pi_\theta$, see e.g. Proposition 5.102 in \cite{cardel19}, we have 
\beqs
\d \mrV_\theta(s, X_s^{t,x,\mu}, \P_{ X_s^{t,\xi}}) 
&=& 
\big( \Lc^{\alpha_s}_{\theta}[\mrV_\theta](s, X_s^{t, x, \mu}, \P_{ X^{t, \xi}_s})  + \beta \mrV_\theta(s, X_s^{t,x,\mu},\P_{ X_s^{t,\xi}}) \big) \,  \d s  \\
& & \quad \; + \;  
D_x  \mrV_\theta(s, X_s^{t,x,\mu},\P_{ X_s^{t,\xi}})\trans\sigma( X_s^{t,x,\mu},\P_{ X_s^{t,\xi}}, \alpha_s) \, \d W_s, \quad t \leq s \leq T,  
\enqs
and thus by definition of $\tilde F_\theta$
\begin{align*}
&\int_t^T  e^{-\beta(s-t)}   \tilde F_{\theta}(s, X_s^{t,x,\mu},\P_{ X_s^{t,\xi}}, \alpha_s)  \,  \d s\\
& =  \int_t^T e^{-\beta(s-t)} \nabla_\theta \log(p_\theta(s,  X_s^{t, x, \mu}, \P_{ X_s^{t, \xi}}, \alpha_s)) \Big(\d \mrV_{\theta}(s,  X_s^{t, x, \mu}, \P_{X_s^{t, \xi}}) - \beta \mrV_{\theta}(s,  X_s^{t, x, \mu}, \P_{ X_s^{t, \xi}}) \\
& \quad + \Hc_\theta[\mrV_\theta](s,X_s^{t, x, \mu}, \P_{X_s^{t, \xi}}) \Big) \, \d s \\
& + \int_t^T  e^{-\beta(s-t)}  \nabla_\theta \log(p_\theta(s,  X_s^{t, x, \mu}, \P_{ X_s^{t, \xi}}, \alpha_s)) \Big( f(X_s^{t,x,\mu},\P_{ X_s^{t,\xi}}, \alpha_s) + \lambda \log(p_\theta (s,  X_s^{t, x, \mu}, \P_{ X_s^{t, \xi}}, \alpha_s) )   \Big) \,  \d s \\
& - \int_t^T e^{-\beta(s-t)}\nabla_\theta \log(p_\theta(s,  X_s^{t, x, \mu}, \P_{ X_s^{t, \xi}}, \alpha_s)) \,  D_x  \mrV_\theta(s, X_s^{t,x,\mu},\P_{ X_s^{t,\xi}})\trans\sigma( X_s^{t,x,\mu},\P_{ X_s^{t,\xi}}, \alpha_s)   \, \d W_s.
\end{align*}

Note that \eqref{diff:density:condition:stochastic:integral} as well as the bound $|D_x V_\theta(s, x, \mu)|\leq C(1+|x|+|\mu|^q)$, for some $q\geq0$, directly deduced from the identity \eqref{first:space:derivative:value:function} and Assumption \ref{hypcoeff}, guarantees that the stochastic integral appearing in the right-hand side of the above identity is a square integrable martingale. Hence, taking expectation in both sides of the above identity eventually yields
\begin{align} 
G_\theta(t,x,\mu) & := \;   \E_{\alpha\sim\pi_\theta}   \Big[ \int_t^T  e^{-\beta(s-t)}  \nabla_\theta \log p_\theta(s,X_s^{t,x,\mu},\P_{X_s^{t,\xi}},\alpha_s) \Big\{ \d \mrV_\theta(s,X_s^{t,x,\mu},\P_{X_s^{t,\xi}}) \\
&  \hspace{1.1cm}  + \;  \big[  f(X_s^{t,x,\mu},\P_{X_s^{t,\xi}},\alpha_s) +  \lambda  \log p_\theta(s,X_s^{t,x,\mu},\P_{X_s^{t,\xi}},\alpha_s)  - \beta \mrV_\theta(s,X_s^{t,x,\mu},\P_{X_s^{t,\xi}}) \big] \d s  \Big\}   \\
& \hspace{3cm} + \; \int_t^T e^{-\beta(s-t)}  \Hc_\theta[\mrV_\theta](s,X_s^{t,x,\mu},\P_{X_s^{t,\xi}})   \d s \Big].
\end{align} 
This proves the announced probabilistic representation formula for $G_\theta$. 

\section{Linear quadratic mean-field control with randomised controls and entropy regularisation} \label{sec:LQappen}

A  stochastic policy is  a probability  transition kernel from $[0,T]\times\R^d\times\Pc_2(\R^d)$ into $A$ $=$ $\R^m$, i.e., a  measurable function $\pi$ $:$ $(t,x,\mu)$ $\in$ $[0,T]\times\R^d\times\Pc_2(\R^d)$ $\mapsto$ $\pi(.|t,x,\mu)$ $\in$ $\Pc(\R^m)$.  We denote by $\Pi$ the set of stochastic policies 
$\pi$ with densities $p$  with respect to the Lebesgue measure on $\R^m$: $\pi( \d a |t,x,\mu)$ $=$ $p(t,x,\mu,a)\d a$.  
We say that the process $\alpha$ $=$ $(\alpha_t)_t$  is a randomised feedback control generated from a stochastic  policy $\pi$ $\in$ $\Pi$, denoted by $\alpha$ $\sim$ $\pi$, if at each time $t$, the action $\alpha_t$ is sampled (according to the $\sigma$-algebra $\Gc$) 
from the probability distribution $\pi(.|t,X_t,\P_{X_t})$. The dynamics $X$ $=$ $X^\alpha$ follows a linear mean-field dynamics with coefficients $b(x,\mu,a)$ $=$ $\bar b(x,\bar \mu,a)$,   $\sigma(x,\mu,a)$ $=$ $\bar \sigma(x,\bar \mu,a)$ in the form
\beqs
\bar b(x,\bar x,a) \; = \;  B x + \bar B \bar x +  C a,  \quad  \bar\sigma(x,\bar x,a) \; = \;  \gamma + D x + \bar D \bar x + Fa, 
\enqs
for $(x,\mu,\bar x,a)$ $\in$ $\R^d\times\Pc_2(\R^d)\times\R^d\times\R^m$, where we denote by $\bar\mu$ $=$ $\int x \mu(\d x)$,  $B$, $\bar B$, $D$, $\bar D$ are constant matrices in $\R^{d\times d}$, $C$, $F$ are constant matrices in $\R^{d\times m}$, $\gamma$ is a constant in $\R^d$.  

Given a stochastic policy $\pi$ $\in$ $\Pi$, we consider the functional cost $V^\pi$ with entropy regulariser defined in \eqref{defVpi} with quadratic functions $f(x,\mu,a)$ $=$ $\bar f(x,\bar \mu,a)$ and $g(x,\mu)$ $=$ $\bar g(x,\bar \mu)$:
\beqs
\bar f(x,\bar x,a) \; = \; x\trans Qx + \bar x\trans \bar Q\bar x  + a\trans N a  + 2a\trans Ix + 2a\trans \bar I \bar x +  2M.x + 2H.a, \\
\bar g(x,\bar x) \; = \; x\trans P x + \bar x\trans \bar P\bar x + 2L.x,
\enqs
where $N$ is a  symmetric matrix in $\S_+^m$, $I$, $\bar I$ $\in$ $\R^{m\times d}$,  $Q$, $\bar Q$, $P$, $\bar P$ are  symmetric matrices in  $\S^d$,  
$M$, $L$  $\in$ $\R^d$, $H$ $\in$ $\R^m$, assumed to satisfy the conditions: 

\vspace{1mm}

\noindent {\bf (H1)} (i) There exists $\delta$ $>$ $0$ s.t. 
\beqs
N \; \geq \; \delta I_m, \quad P   \; \geq \; 0, \quad Q  - I\trans N^{-1}I \; \geq \; 0.  
\enqs
or (ii)  $n$ $=$ $m$ $=$ $1$, $I$ $=$ $0$, $F$ $\neq$ $0$, $Q$ $\geq$ $0$,  $P$ $>$ $0$. 

\vspace{1mm} 

\noindent {\bf (H2)} (i) There exists $\delta$ $>$ $0$ s.t. 
\beqs
N \; \geq \; \delta I_m, \quad P + \bar P  \; \geq \; 0, \quad (Q + \bar Q)  -  (I + \bar I)\trans N^{-1} (I + \bar I) \; \geq \; 0.  
\enqs
or (ii)  $I+\bar I$ $=$ $0$, $F$ $\neq$ $0$,  $Q + \bar Q$ $\geq$ $0$,  $P+\bar P$ $\geq$ $0$, $P$ $>$ $0$.

\vspace{3mm}

The solution to the LQ mean-field control problem with entropy regulariser is then given by the following theorem:

\begin{Theorem} \label{theoLQentropy}
Let Assumptions {\bf (H1)}-{\bf (H2)} hold. Then, the value function is equal to 
\beqs
v(t,x,\mu) \; := \; \inf_{\pi \in \Pi} V^\pi(t,x,\mu) &=&  (x-\bar{\mu})\trans K(t)(x-\bar{\mu})+\bar{\mu}\trans \Lambda(t)\bar{\mu}+2Y(t)\trans x+R(t),
\enqs
for $(t,x,\mu)$ $\in$ $[0,T]\times\R^d\times\Pc_2(\R^d)$, where  the quadruple $(K,\Lambda,Y,R)$ valued in $(\S_+^d,\S_+^d,\R^d,\R)$ is solution on $[0,T]$ to the system of Riccati equations:
\begin{equation} \label{ODEKdiscount} 
\begin{cases} 
        &      \dot{K}(t) - \beta K(t) +Q+K(t)B+B\trans K(t)+D\trans K(t) D \\
        &        -(I+C\trans K(t)+F\trans K(t) D)\trans (N+F\trans K(t) F)^{-1}(I+C\trans K(t)+F\trans K(t) D)   \;  = \; 0,  \\
        & \dot{\Lambda}(t) -\beta\Lambda(t) + \hat Q +\Lambda(t)\hat B +\hat B\trans \Lambda(t)+\hat D\trans K(t)\hat D  \\
        & - \big(\hat I +C\trans\Lambda(t)+F\trans K(t)\hat D  \big)\trans(N+F\trans K(t) F)^{-1}\big(\hat I +C\trans\Lambda(t)+F\trans K(t)\hat D \big) \; = \; 0  \\
        &  \dot{Y}(t) -\beta Y(t) +M+\hat B \trans Y(t)+ \hat D \trans K(t)\gamma\\
        & -(\hat I +C\trans\Lambda(t)+F\trans K(t) \hat D)\trans(N+F\trans K(t) F)^{-1}(H+C\trans Y(t)+F\trans K(t)\gamma)  \; = \; 0\\
        & \dot{R}(t)  -\beta R(t)  +\gamma\trans K(t)\gamma  +  \frac{\lambda m}{2}\log(2\pi)-\frac{\lambda}{2}\log\lvert\frac{\lambda}{2 \text{det}(N+F\trans K(t) F)} \rvert   \\
        & -(H+C\trans Y(t)+F\trans K(t)\gamma)\trans(N+F\trans K(t) F)^{-1}(H+C\trans Y(t)+F\trans K(t)\gamma)  \; = \; 0 
\end{cases}        
\end{equation}
with the terminal condition $(K(T),\Lambda(T),Y(T),R(T))$ $=$ $(P,\hat P,L,0)$, where we set $\hat I$ $:=$ $I+\bar I$, $\hat B$ $:=$ $B+\bar B$, $\hat D$ $:=$ $D+\bar D$,  
$\hat Q$ $:=$ $Q + \bar Q$, $\hat P$ $:=$ $P+\bar P$.  

Moreover, the optimal stochastic policy follows a Gaussian distribution: 
\begin{align} \label{optpolicy} 
\pi^*(.|t,x,\mu) & =  \;   \Nc \Big( - S(t)^{-1} \big(  U(t) x + (\hat U(t) - U(t))\bar\mu + O(t) \big); \frac{\lambda}{2} S(t)^{-1} \Big), 
\end{align} 
where we set 
\beqs
S(t) \; := \; N + F\trans K(t) F, &  & O(t) \; := \;  H + C\trans Y(t)  + F\trans K(t) \gamma \\
U(t) \; := \; I + C\trans K(t) + F\trans K(t) D,  & & \hat U(t) \; := \; \hat I  + C \trans \Lambda(t) + F\trans K(t) \hat D. 
\enqs
\end{Theorem}

\begin{Remark}
Conditions {\bf (H1)} and {\bf (H2)} ensure the existence and uniqueness of a solution $(K,\Lambda)$ to the matrix Riccati equation in \eqref{ODEKdiscount} satisfying $K$ $\geq$ $0$, $\Lambda$ $\geq$ $0$ (hence $S(t)^{-1}$ is well-defined). Given $(K,\Lambda)$, the equations for 
$(Y,R)$  are simply linear ODEs. 
\end{Remark}

\vspace{1mm}

\noindent {\bf Proof of Theorem \ref{theoLQentropy}.}  We adapt the arguments in  \cite{baspha19} to our case with randomised controls and entropy regulariser.  

\noindent {\it Step 1.} Let us consider the function defined on $[0,T]\times\R^d\times\Pc_2(\R^d)$ by $w(t,x,\mu)$ $=$ $\bar w(t,x,\bar\mu)$, where $\bar w$ is defined on $[0,T]\times\R^d\times\R^d$ by 
\beqs
\bar w(t,x,\bar x) &=& (x-\bar{x})\trans K(t)(x-\bar{x})+\bar{x}\trans \Lambda(t)\bar{x}+2Y(t)\trans x+R(t),
\enqs
for some functions (to be determined later)  $K$, $\Lambda$, $Y$ and $R$ on $[0,T]$, and valued on $\S_+^d$, $\S_+^d$, $\R^d$, and $\R$.  Fix $(t_0,x_0,\mu_0)$ $\in$ $[0,T]\times\R^d\times\Pc_2(\R^d)$, and $\xi_0$ $\in$ $L^2(\Fc_{t_0};\R^d)$ $\sim$ $\mu_0$.   Given $\pi$ $\in$ $\Pi$ with density $p$,  and a randomised control $\alpha$ $\sim$ $\pi$, we consider the process 
\beqs
\Sc_t^\alpha &:=&  e^{-\beta(t-t_0)} \bar w(t,X_t^{t_0,x_0,\mu_0},\bar X_t^{t_0,\mu_0}) + \int_{t_0}^t e^{-\beta(s-t_0)} \big[  \bar f(X_s^{t_0,x_0,\mu_0},\bar X_s^{t_0,\mu_0},\alpha_s)   \\
& &  \hspace{5.5cm} + \;  \lambda \int_{\R^m}  \big(\log 
\mrp_t(a) \big) \mrp_t(a) \d a 
\big]  \d s,    
\enqs
for $t_0\leq t \leq T$, where we set $\mrp_t(a)$ $=$ $p(t,X_t^{t_0,x_0,\mu_0},\P_{X_t^{t_0,\xi_0}},a)$, and 
$\bar X_t^{t_0,\mu_0}$ $:=$ $\E_{\alpha\sim\pi}[X_t^{t_0,\xi_0}]$ which follows the dynamics:
\beqs
d \bar X_t &=& \big( \hat B \bar X_t  + C \bar \alpha_t) \d t, 
\enqs
with $\bar \alpha_t$ $:=$ $\E_{\alpha\sim\pi}[\alpha_t]$. 

\vspace{1mm}

\noindent {\it Step 2.}  We apply It\^o's formula to $\Sc_t^\alpha$ for $\alpha$ $\sim$ $\pi$, and take the expectation to get
\begin{align} \label{dynSc} 
\d \E_{\alpha\sim\pi}[\Sc_t^\alpha] &= \;  e^{-\beta(t-t_0)} \E_{\alpha\sim\pi}[\Dc_t^\alpha] \d t,
\end{align} 
with 
\beqs
\Dc_t^\alpha &=&  - \beta \bar w(t,X_t,\bar X_t) + \frac{\d}{\d t} \E_{\alpha\sim\pi}[\bar w(t,X_t,\bar X_t)] + \bar f(X_t,\bar X_t,\alpha_t)   + \lambda \int_{\R^m} (\log \mrp_t(a)) \mrp_t(a) \d a,
\enqs
where we omit the dependence on $t_0,x_0,\mu_0$ of $X$ and $\bar X$ to alleviate notations.  By applying It\^o's formula to $\bar w(t,X_t,\bar X_t)$, recalling the quadratic forms of $\bar w$, $\bar f$, and using the linear dynamics of $X$ and $\bar X$, we obtain similarly as in \cite{baspha19} (after careful but straightforward computations): 
\begin {align} 
\E_{\alpha\sim\pi}[\Dc_t^\alpha]  \; = &  \;   \E_{\alpha \sim\pi} \Big[  
(X_t-\bar{X}_t)\trans \big(\dot{K}(t) - \beta K(t) + Q  +K(t)B+B\trans K(t)+D\trans K(t) D\big)(X_t-\bar{X}_t)  \nonumber \\
&  \; + \; \bar X_t\trans\big( \dot{\Lambda}(t) - \beta \Lambda(t) + \hat Q +  \Lambda(t) \hat B + \hat B\trans \Lambda(t)+ \hat D \trans K(t)\hat D \big)\bar{X}_t  \nonumber \\
&   \; + \; 2\big(\dot{Y}(t) - \beta Y(t) + M  + \hat B\trans Y(t)+\hat D\trans K(t)\gamma\big)\trans X_t +  \dot{R}(t) - \beta R(t) +\gamma\trans K(t)\gamma \nonumber \\
&  + \; \alpha_t\trans S(t) \alpha_t +  2\alpha_t\trans \big(  U(t) (X_t - \bar X_t)  +   \hat U(t) \bar{X}_t + O(t) \big)  +  \lambda \int_{\R^m} (\log \mrp_t(a)) \mrp_t(a) \d a   \Big] \nonumber \\
\; = & \;    \E_{\alpha \sim\pi} \Big[  
(X_t-\bar{X}_t)\trans \big(\dot{K}(t) - \beta K(t) + Q  +K(t)B+B\trans K(t)+D\trans K(t) D\big)(X_t-\bar{X}_t)  \nonumber \\
&  \; + \; \bar X_t\trans\big( \dot{\Lambda}(t) - \beta \Lambda(t) + \hat Q +  \Lambda(t) \hat B + \hat B\trans \Lambda(t)+ \hat D \trans K(t)\hat D \big)\bar{X}_t  \nonumber \\
&   \; + \; 2\big(\dot{Y}(t) - \beta Y(t) + M  + \hat B\trans Y(t)+\hat D\trans K(t)\gamma\big)\trans X_t +  \dot{R}(t) - \beta R(t) +\gamma\trans K(t)\gamma \nonumber \\
&  
\; + \; \int_{\R^m} [ \phi_t(a) +  \lambda \log \mrp_t(a) ] \mrp_t(a) \d a \Big],  \label{expressD} 
\end{align} 
where we used in the last equality the fact that $\alpha$ $\sim$ $\pi$, and set  $\phi_t(a)$ $:=$ $a\trans S(t) a  +  2 a \trans \chi_t$ with 
$\chi_t$ $:=$ $U(t) (X_t - \bar X_t)  +   \hat U(t) \bar{X}_t + O(t)$.

\vspace{1mm}

\noindent {\it Step 3.}  Let $\phi$ be a quadratic function on $\R^m$: $\phi(a)$ $=$ $a\trans S a + 2 a\trans\chi$  for some positive-definite matrix $S$ $\in$ $\S_+^m$, and $\chi$ $\in$ $\R^m$, and denote by  
$\Dc_2(\R^m)$ the set of square integrable density functions on $\R^m$, i.e., the set of nonnegative measurable functions $\mrp$ on $\R^m$ s.t. $\int_{\R^m} \mrp(a) \d a$ $=$ $1$, and $\int_{\R^m} |a|^2 \mrp(a) \d a $ $<$ $\infty$.  Let us consider the cost functional on $\Dc_2(\R^m)$ defined by
\beqs
C_{\phi}(\mrp) & := & \int_{\R^m} [ \phi(a) + \lambda \log \mrp(a) ] \mrp(a) \d a. 
\enqs
Then, the minimizer of $C_\phi$ is achieved with $\mrp^*$ $\in$ $\Dc_2(\R^m)$ given by 
\begin{align} \label{pstar} 
\mrp^*(a) &= \;  \frac{\exp\big(-\frac{1}{\lambda} \phi(a) \big)}{\int_{\R^m} \exp\big(-\frac{1}{\lambda} \phi(a) \big) \d a }, \quad a \in \R^m.   
\end{align} 
Indeed, by considering the Lagrangian function associated to  this  minimization problem 
\beqs
L_\phi(\mrp,\nu) &=& C_\phi(\mrp) - \nu\big( \int_{\R^m} \mrp(a)\d a - 1 \big) \; = \; \int_{\R^m} \big[ \phi(a) + \lambda \log \mrp(a) - \nu \big] \mrp(a) \d a \;  +  \; \nu, 
\enqs
for $(\mrp,\nu)$ $\in$ $\Dc_2(\R^m)\times\R$, we see that the minimization over $\mrp$ is obtained pointwisely, i.e. inside the integral over $a$ $\in$ $\R^m$, hence leading to the first-order equations: 
\begin{equation}
\begin{cases}
\phi(a) + \lambda \log \mrp^*(a) - \nu^* +  \lambda  \; = \;  0, \quad a \in \R^m,  \\
\int_{a \in \R^m} \mrp^*(a) \d a \; =\;  1. 
\end{cases}
\end{equation} 
This yields the expression of $\mrp^*$ in \eqref{pstar}, which is  actually the density of  a Gaussian distribution 
\begin{align} \label{pistarinter} 
\pi^* &= \;  \Nc\Big( - S^{-1} \chi; \frac{\lambda}{2} S^{-1} \Big). 
\end{align} 
The infimum of $C_\phi$ is then equal to 
\begin{align} \label{minC} 
\inf_{\mrp \in \Dc_2(\R^m)} C_\phi(\mrp) &= \;  C_\phi(\mrp^*) \; = \;  - \chi\trans S^{-1} \chi  - \frac{\lambda m}{2}\log(2\pi)-\frac{\lambda}{2}\log\lvert\frac{\lambda}{2\text{det}(S)}\rvert. 
\end{align}

\vspace{1mm}

\noindent {\it Step 4.}  Notice that under {\bf (H1)}, the matrix  $S(t)$ $=$ $N+F\trans K(t) F$ is positive-definite for $K$ $\geq$ $0$,  and $\mrp_t(.)$ $\in$ $\Dc_2(\R^m)$ a.s. for $t$ $\in$ $[t_0,T]$. 
From \eqref{expressD} and \eqref{minC}, we then have for all $\pi$ $\in$ $\Pi$, 
\begin{align}
& \; \E_{\alpha\sim\pi}[\Dc_t^\alpha]  \\
\; = & \;    \E_{\alpha \sim\pi} \Big[  
(X_t-\bar{X}_t)\trans \big(\dot{K}(t) - \beta K(t) + Q  +K(t)B+B\trans K(t)+D\trans K(t) D\big)(X_t-\bar{X}_t)  \nonumber \\
&  \; + \; \bar X_t\trans\big( \dot{\Lambda}(t) - \beta \Lambda(t) + \hat Q +  \Lambda(t) \hat B + \hat B\trans \Lambda(t)+ \hat D \trans K(t)\hat D \big)\bar{X}_t  \nonumber \\
&   \; + \; 2\big(\dot{Y}(t) - \beta Y(t) + M  + \hat B\trans Y(t)+\hat D\trans K(t)\gamma\big)\trans X_t +  \dot{R}(t) - \beta R(t) +\gamma\trans K(t)\gamma \nonumber \\
&   \; + \; C_{\phi_t}(\mrp_t) \Big] \\
\; \geq & \;   \E_{\alpha \sim\pi} \Big[  
(X_t-\bar{X}_t)\trans \big(\dot{K}(t) - \beta K(t) + Q  +K(t)B+B\trans K(t)+D\trans K(t) D   - U(t)\trans S(t)^{-1} U(t) \big)(X_t-\bar{X}_t)  \nonumber \\
& \; + \; \bar X_t\trans\big( \dot{\Lambda}(t) - \beta \Lambda(t) + \hat Q +  \Lambda(t) \hat B + \hat B\trans \Lambda(t)+ \hat D \trans K(t)\hat D  - \hat U(t)\trans S(t)^{-1} \hat U(t)  \big)\bar{X}_t  \nonumber \\ 
&   \; + \; 2\big(\dot{Y}(t) - \beta Y(t) + M  + \hat B\trans Y(t)+\hat D\trans K(t)\gamma  - O(t)\trans S(t)^{-1} \hat U(t) \big)\trans X_t \nonumber \\
& \; + \;   \dot{R}(t) - \beta R(t) +\gamma\trans K(t)\gamma - O(t)\trans S(t)^{-1}O(t) - \frac{\lambda m}{2}\log(2\pi)-\frac{\lambda}{2}\log\lvert\frac{\lambda}{2\text{det}(S(t))}\rvert.  \label{boundD}
\end{align} 
Therefore, by taking $(K,\Lambda,Y,R)$ solution to \eqref{ODEKdiscount}, we  see that the r.h.s. of \eqref{boundD} vanishes, which means that for all $\pi$ $\in$ $\Pi$, $\E_{\alpha\sim\pi}[\Dc_t^\alpha]$ $\geq$ $0$.  Moreover, from \eqref{pistarinter}, the equality in \eqref{boundD} holds true for the choice of $\pi^*$ $\in$ $\Pi$ as defined in 
\eqref{optpolicy}, and thus
\beqs
\inf_{\pi\in\Pi} \E_{\alpha\sim\pi}[\Dc_t^\alpha] &=&  \E_{\alpha\sim\pi^*}[\Dc_t^\alpha] \; = \; 0, \quad  t \in [t_0,T]. 
\enqs 
From \eqref{dynSc}, this means that the function $t$ $\mapsto$ $\E_{\alpha\sim\pi}[\Sc_t^\alpha]$ is nondecreasing on $[t_0,T]$ for any $\pi$ $\in$ $\Pi$, and constant on $[t_0,T]$ for $\pi$ $=$ $\pi^*$. By definition of $\Sc^\alpha$, $V^\pi$,  and noting that $\bar w(T,x,\bar x)$ $=$ $\bar g(x,\bar x)$ from the terminal condition on $(K,\Lambda,Y,R)$,  
it follows  that 
\begin{align} \label{interVpi} 
w(t_0,x_0,\mu_0) \; = \; \bar w(t_0,x_0,\bar\mu_0) \; = \;  \E_{\alpha\sim\pi}[S_{t_0}^\alpha]  & \leq \;   \E_{\alpha\sim\pi}[S_{T}^\alpha] \; = \; V^{\pi}(t_0,x_0,\mu_0),  
\end{align} 
for any $\pi$ $\in$ $\Pi$, with equality   in \eqref{interVpi} for $\pi$ $=$ $\pi^*$.  We conclude that 
\beqs
\inf_{\pi \in \Pi} V^\pi(t_0,x_0,\mu_0) & = & V^{\pi^*} (t_0,x_0,\mu_0) \; = \;  w(t_0,x_0,\mu_0) \\
&=& (x_0-\bar{\mu_0})\trans K(t_0)(x_0-\bar{\mu_0})+\bar{\mu_0}\trans \Lambda(t_0)\bar{\mu_0}+2Y(t_0)\trans x_0+R(t_0). 
\enqs
\ep

\bibliographystyle{plain}

\bibliography{biblioMFRL}

\end{document}